\documentclass{sage}
\usepackage{times}

\usepackage{amsthm,amssymb,natbib,url,amsmath}
\usepackage{multicol}
\usepackage[bookmarks=true]{hyperref}
\usepackage[usenames, dvipsnames]{color}

\usepackage{amsfonts,mathrsfs}
\usepackage{verbatim}
\let\chapter\section
\usepackage[vlined,ruled,linesnumbered]{algorithm2e}
\usepackage{wrapfig}
\usepackage{graphicx} 
\usepackage{xspace}
\usepackage{subfigure}
\usepackage{multirow}

\journal{IJRR}
\volume{000}
\issue{00}
\copyrightline{$\copyright$ Copyright}
\firstpage{1}
\lastpage{40}
\doi{doi number}
\articletype{Article type}
\pubyear{2014}

\begin{document}

\title{Asymptotically Optimal Sampling-based Kinodynamic Planning}
 \author{Yanbo Li, Zakary Littlefield, Kostas E. Bekris}
\maketitle

\definecolor{myblue}{rgb}{0.15, 0.05, 0.7}
\newcommand{\edited}[1]{#1}

\newcommand{\drt}{{\tt DoublySampling-Random-Tree}}
\newcommand{\rt}{{\tt NAIVE\_RANDOM\_TREE}}
\newcommand{\DRT}{{\tt DRT}}
\newcommand{\prm}{{\tt PRM}}
\newcommand{\EST}{{\tt Expansive-Space Tree}}
\newcommand{\est}{{\tt EST}}
\newcommand{\kprmstar}{{\tt k-PRM$^*$}}
\newcommand{\prmstar}{{\tt PRM$^*$}}
\newcommand{\rrt}{{\tt RRT}}
\newcommand{\rrtextend}{{\tt RRT-Extend}}
\newcommand{\rrtconnect}{{\tt RRT-Connect}}
\newcommand{\sst}{{\tt SST}}
\newcommand{\sststar}{{\tt SST$^\ast$}}
\newcommand{\rrtc}{{\tt RRT-Connect}}
\newcommand{\mcrrt}{{\tt MonteCarlo-RRT}}
\newcommand{\rrtbestnear}{{\tt RRT-BestNear}}
\newcommand{\rrtstarbestnear}{{\tt RRT$^\ast$-BestNear}}
\newcommand{\rrtdrain}{{\tt RRT-Drain}}
\newcommand{\sparserrt}{{\tt Sparse-RRT}}
\newcommand{\stable}{{\tt STABLE\_SPARSE\_RRT}}
\newcommand{\stablestar}{{\tt STABLE\_SPARSE-RRT$^*$}}
\newcommand{\sparserrtstar}{{\tt Sparse-RRT$^*$}}
\newcommand{\rrg}{{\tt RRG}}
\newcommand{\rrtstar}{{\tt RRT$^*$}}
\newcommand{\bvp}{{\tt BVP}}
\newcommand{\alg}{{\tt ALG}}
\newcommand{\mcprop}{{\tt MonteCarlo-Prop}}
\newcommand{\steer}{{\tt Steering}}

\newcommand{\bestnear}{{\tt BestNear}}
\newcommand{\drain}{{\tt Drain}}

\newcommand{\dx}{d_x}
\newcommand{\du}{d_u}
\newcommand{\map}{\mathbb{M}}
\newcommand{\Xspace}{\mathbb{X}}
\newcommand{\Pspace}{\mathbb{P}}
\newcommand{\Uspace}{\mathbb{U}}
\newcommand{\Um}{\mathbb{U}_m}
\newcommand{\UmT}{\mathbb{U}_m^T}
\newcommand{\Xfree}{\mathbb{X}_{f}}
\newcommand{\Xgoal}{\mathbb{X}_{G}}
\newcommand{\xinit}{x_0}
\newcommand{\trajspace}{\Pi}
\newcommand{\vertx}{\mathbb{V}}
\newcommand{\graph}{\mathbb{G}}
\newcommand{\Vactive}{\mathbb{V}_{active}}
\newcommand{\Voff}{\mathbb{V}_{inactive}}
\newcommand{\Ddrain}{\delta_{s}}
\newcommand{\Dnear}{\delta_{BN}}
\newcommand{\edges}{\mathbb{E}}
\newcommand{\xnear}{\mathbb{X}_{near}}
\newcommand{\xlocal}{\mathbb{X}_{local}}

\newcommand{\ofalg}{^{ALG}_n}

\newcommand{\Prob}{\mathbb{P}}
\newcommand{\balls}{\mathbb{B}}
\newcommand{\ball}{\mathcal{B}}
\newcommand{\A}{\mathcal{A}}
\newcommand{\reals}{\mathbb{R}}
\newcommand{\integers}{\mathbb{Z}}

 \newcommand{\telos}{\hfill \ensuremath{\blacksquare}}

\newenvironment{myitem}{\begin{list}{$\bullet$}
{\setlength{\itemsep}{-0pt}
\setlength{\topsep}{0pt}
\setlength{\labelwidth}{0pt}
\setlength{\leftmargin}{10pt}
\setlength{\parsep}{-0pt}
\setlength{\itemsep}{0pt}
\setlength{\partopsep}{0pt}}}%
{\end{list}}

\newtheorem{thm}{\bf Theorem}
\newtheorem{definition}[thm]{\bf Definition}
\newtheorem{assumption}[thm]{\bf Assumption}
\newtheorem{requirement}[thm]{\bf Requirement}
\newtheorem{lemmma}[thm]{\bf Lemma}
\newtheorem{coro}[thm]{\bf Corollary}
\newcommand{\argmin}[1]{\underset{#1}{\operatorname{arg}\,\operatorname{min}}\;}
\newtheorem{prop}[thm]{Proposition}

\begin{abstract}
Sampling-based algorithms are viewed as practical solutions for
high-dimensional motion planning. Recent progress has taken advantage
of random geometric graph theory to show how asymptotic optimality can
also be achieved with these methods. Achieving this desirable property
for systems with dynamics requires solving a \emph{two-point boundary
value problem} (\bvp) in the state space of the underlying dynamical
system.  It is difficult, however, if not impractical, to generate
a \bvp\ solver for a variety of important dynamical models of robots
or physically simulated ones. Thus, an open challenge was whether it
was even possible to achieve optimality guarantees when planning for
systems without access to a \bvp\ solver. This work resolves the above
question and describes how to achieve asymptotic optimality for
kinodynamic planning using incremental sampling-based planners by
introducing a new rigorous framework. Two new methods, \stable\ (\sst)
and \sststar, result from this analysis, which are asymptotically
near-optimal and optimal, respectively. The techniques are shown to
converge fast to high-quality paths, while they maintain only a sparse
set of samples, which makes them computationally efficient. The good
performance of the planners is confirmed by experimental results using
dynamical systems benchmarks, as well as physically simulated robots.

\end{abstract}

\section{Introduction}
\label{sec:intro}

{\bf Kinodynamic Planning:} For many interesting robots it is
difficult to adapt a collision-free path into a feasible one given the
underlying dynamics. This class of robots includes ground vehicles at
high-velocities (\cite{Likhachev2009Planning-Long-D}), unmanned aerial
vehicles, such as fixed-wing airplanes
(\cite{Richter2013Polynomial-Traj}), or articulated robots with
dynamics, including balancing and locomotion systems
(\cite{Kuindersma2014An-efficiently-}). In principle, most robots
controlled by the second-order derivative of their configuration
(e.g., acceleration, torque) and which exhibit drift cannot be treated
by a decoupled approach for trajectory planning given their
controllability properties (\cite{1998Robot-Motion-Pl,
Choset2005Principles-of-R}). To solve such challenges, the idea
of \emph{kinodynamic planning} has been proposed
(\cite{Donald1993Kinodynamic-Mot}), which involves directly searching
for a collision-free and feasible trajectory in the underlying
system's state space. This is a harder problem than kinematic path
planning, as it involves searching a higher-dimensional space and
respecting the underlying flow that arises from the dynamics. Given
its importance, however, it has attracted a lot of attention in the
robotics community. The focus in this work is on the properties of the
popular sampling-based motion planners for kinodynamic challenges
(\cite{Kavraki1996Probabilistic-R, LaValle2001, Hsu2002,
Karaman2011Sampling-based-}).

\noindent {\bf Sampling-based Motion Planning:} The sampling-based
approach has been shown to be a practical solution for quickly finding
feasible paths for relatively high-dimensional motion planning
challenges (\cite{Kavraki1996Probabilistic-R, LaValle2001, Hsu2002}).
The first popular methodology, the Probabilistic Roadmap Method (\prm)
(\cite{Kavraki1996Probabilistic-R}) focused on preprocessing the
configuration space of a kinematic system so as to generate a roadmap
that can be used to quickly answer multiple queries.  Tree-based
variants, such as \rrtextend\ (\cite{LaValle2001}) and \est\
(\cite{Hsu2002}), focused on addressing kinodynamic problems. For all
these methods, the guarantee provided is relaxed to probabilistic
completeness, i.e., the probability of finding a solution if one
exists, converges to one (\cite{Kavraki1998Analsis-of-Prob,
Hsu1998On-finding-narr, Ladd2004Measure-Theoret}). This was seen as a
sufficient objective in the community given the hardness of motion
planning and the \emph{curse of dimensionality}. More recently,
however, the focus has shifted from providing feasible solutions to
achieving high-quality solutions.  A milestone has been the
identification of the conditions under which sampling-based algorithms
are asymptotically optimal. These conditions relate to the
connectivity of the underlying roadmap based on results on random
geometric graphs (\cite{Karaman2011Sampling-based-}). This line of
work provided asymptotically optimal algorithms for motion planning,
such as \prmstar\ and \rrtstar\ (\cite{Karaman2010Incremental-Sam}).

\noindent {\bf Lack of a \bvp\ Solution:} A requirement for the
generation of a motion planning roadmap is the existence of a steering
function. This function returns the optimum path between two states in
the absence of obstacles. In the case of a dynamical system, the
steering function corresponds to the solution of a two-point boundary
value problem (\bvp). Addressing this problem corresponds to solving a
differential equation, while also satisfying certain boundary
conditions. It is not easy, however, to produce a \bvp\ solution for
many interesting dynamical systems and this is the reason that roadmap
planners, including the asymptotically optimal \prmstar, cannot by
used for kinodynamic planning.

Unfortunately, \rrtstar\ also requires a steering function, as it
reasons over an underlying roadmap even though it generates a tree data
structure.  While in certain cases it is sufficient to plan for a
linearized version of the dynamics (\cite{Webb2013Kinodynamic-RRT}) or
using a numerical approximation to the \bvp\ problem, this approach
is not a general solution.  Furthermore, it does not easily address an
important class of planning challenges, where the system is simulated
using a physics engine. In this \edited{situation}, the
primitive available to the planning process is forward propagation of
the dynamics using the physics engine. Thus, an open problem for the
motion planning community was whether it was even possible to achieve
optimality given access only to a forward propagation model of the
dynamics.

\begin{wrapfigure}{r}{.5\textwidth}
\vspace{-.15in}
\includegraphics[width = .24\textwidth]{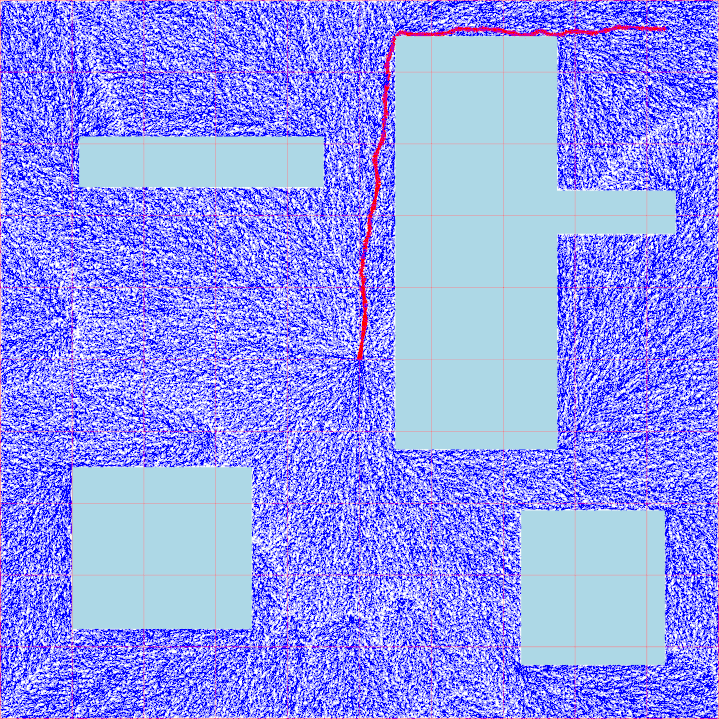}
\includegraphics[width = .24\textwidth]{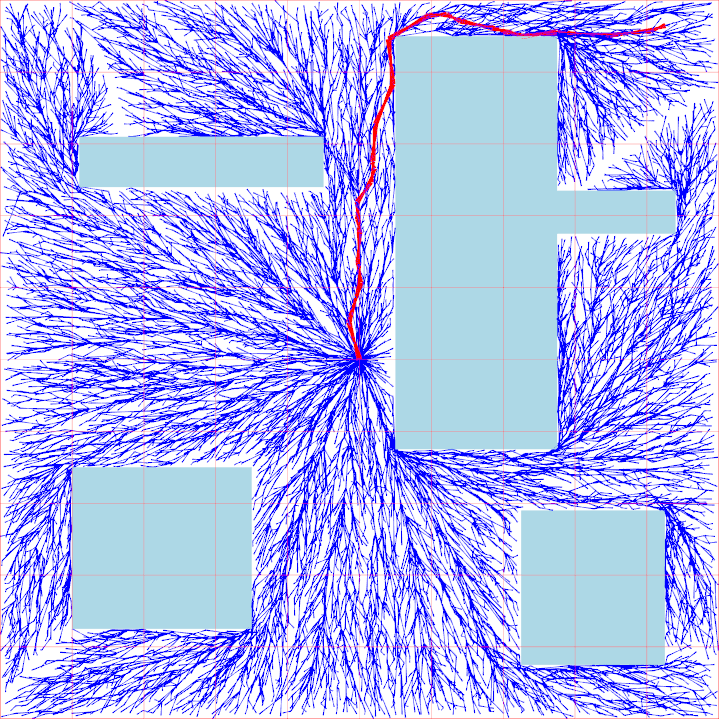}
\caption{Trees constructed by \rrtstar\ (left) and \sst\ (right)
for a 2D kinematic point system after 1 minute of computation. Solution paths
are shown in red. \sst\ does not require a steering function
as \rrtstar\ does, making \sst\ more useful in kinodynamic problems.}
\label{fig:prelim_results}
\end{wrapfigure}
\noindent {\bf Summary of Contribution:} This paper introduces a new
way to analyze the properties of incremental sampling-based algorithms
that construct a tree data structure for a wide class of kinodynamic
planning challenges. This analysis provides the conditions under which
asymptotic optimality can be achieved when a planner has access only
to a forward propagation model of the system's dynamics. The reasoning
is based on a kinodynamic system's accessibility properties and
probability theory to argue probabilistic completeness and asymptotic
optimality for non-holonomic systems where \emph{Chow's condition}
holds \citep{Chow1939Uber-Systemen-v}, eliminating the requirement for
a \bvp\ solution.  Based on these results, a series of sampling-based
planners for kinodynamic planning are considered:

\begin{enumerate}
\item[a)] A simplification of \est, which extends a tree data structure
in a random way, referred to as \rt : It is shown to be asymptotically
optimal but impractical as it does not have good convergence to high
quality paths.
\item[b)] An approach inspired by an existing variation of \rrt,
referred to as \rrtbestnear\ (\cite{Urmson2003Approaches-for-}), which
promotes the propagation of reachable states with good path cost: It
is shown to be asymptotically near-optimal and has a practical
convergence rate to high quality paths but has a per iteration cost
that is higher than that of \rrt.
\item[c)] The proposed algorithms \stable\ (\sst) and \stablestar\
(\sststar), which \edited{use the \bestnear\ selection process. They apply} a pruning operation to keep the
number of nodes stored small: they are able to achieve asymptotic
near-optimality and optimality respectively. They also have good
convergence rate to high quality paths. \sst\ has reduced per
iteration cost relative to the suboptimal \rrt\ given the pruning
operation, which accelerates searching for nearest
neighbors.
\end{enumerate}

\begin{figure*}
\includegraphics[width = .24\textwidth]{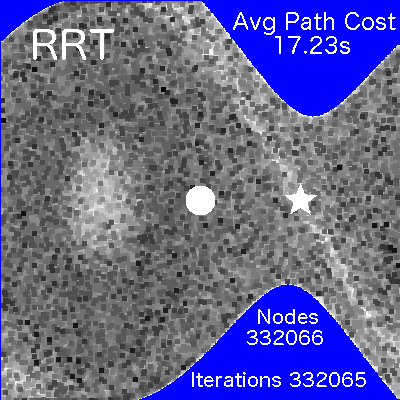}
\includegraphics[width = .24\textwidth]{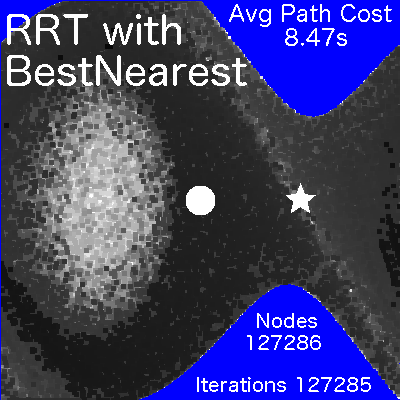}
\includegraphics[width = .24\textwidth]{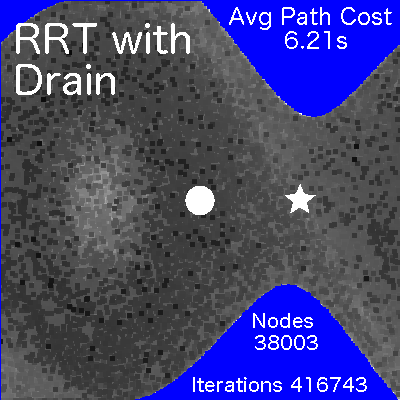}
\includegraphics[width = .24\textwidth]{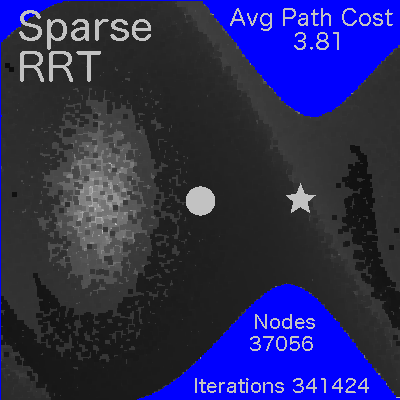}
\caption{Phase plots that show best path cost at each point in the
one-link pendulum state space for each of the proposed
modifications \edited{(the \bestnear\ primitive and the pruning
(mentioned as Drain in the third Figure above))}.  $x$-axis: pendulum
angle, $y$-axis: velocity. Blue corresponds to unexplored regions of
the state space. The circle is state $\{0,0\}$, a horizontal placement
of the pendulum, the star is state $\{\frac{\pi}{2},0\}$, an upward
configuration. Colors are computed by dividing the best path cost to a
state in a pixel by a predefined value (20.0 for \rrt\ and 10.0 for
the other methods) and then mapping the result to the range
[0,255]. All algorithms were executed for the same amount of time (5
min). For the last two methods that provide a sparse representation,
each state is coloring a 3x3 local neighborhood. The best path cost
for each pixel is displayed.}
\label{fig:pendulum_results}
\end{figure*}

\noindent An illustration of the proposed \sst's performance
for a kinematic point system is provided in
Fig. \ref{fig:prelim_results}. This is a simple challenge, where
comparison with \rrtstar\ is possible. This is a problem where \rrt\
typically does not return a path in the homotopic class of the
optimum one. \sst\ is able to do so, while also maintaining a sparse
data structure. Fig. \ref{fig:pendulum_results} describes the
performance of different components of \sst\ in searching the phase
space of a pendulum system relative to \rrt. No method is making use
of a steering function for the pendulum system. A summary of the
desirable properties of \sst\ and \sststar\ in relation to the
efficient \rrt\ and the asymptotically optimal \rrtstar\ is available
in Table \ref{tab:compare}.

\begin{table*}[b]
\begin{center}
\begin{tabular}{| p{.27\textwidth} || p{.27\textwidth} || p{.38\textwidth} |}
\hline
\rrtextend & \rrtstar & \sst/\sststar \\
\hline \hline 
{\bf Probabilistically Complete} (under conditions) & {\bf
Probabilistically Complete} & Probabilistically $\delta$-Robust
Complete / {\bf Probabilistically Complete} \\
\hline 
Provably Suboptimal & {\bf Asymptotically Optimal} & Asymptotically
$\delta$-Robust Near-Optimal / {\bf Asymptotically Optimal} \\
\hline 
{\bf Forward Propagation} & Steering Function & {\bf Forward Propagation} \\
\hline 
{\bf Single Propagation Per Iteration}  & Many Steering Calls Per
Iteration & {\bf Single Propagation Per Iteration} \\
\hline 
1 NN Query ($\mathcal{O}(\log N)$) & 1 NN + 1 K-Query
($\mathcal{O}(\log N)$) & {\bf Bounded Time Complexity Per Iteration}
/ 1 Range Query + 1 NN Query\\
\hline 
Includes All Collision-Free Samples & Includes All Collision-Free
Samples & {\bf Sparse Data Structure} / Converges to All Collision-Free
Samples\\
\hline 
\end{tabular}
\end{center}
\label{tab:compare}
\caption{Comparing \rrt, \rrtstar with the proposed
  \sst\ / \sststar, which minimize computation cost and space
  requirements while providing asymptotic (near-)optimality for
  kinodynamic planning. \edited{This table compares the following from
  top to bottom: completeness properties, optimality properties, the
  process for the \edited{extension} primitive, the number of
  extensions per iteration, the type of nearest neighbor queries
  (nearest, k-closest, and range), as well as space complexity.} The
  notion of $\delta$-robustness is introduced in this paper.}
\end{table*}

\noindent {\bf Paper Overview:} The following section provides a
more comprehensive review of the literature and the relative
contribution of this paper. Then, Section \ref{sec:setup} identifies
formally the considered problem and a set of assumptions under which
the desired properties for the proposed algorithms
hold. Section \ref{sec:algos} first outlines how sampling-based
algorithms need to be adapted so as to achieve asymptotic optimality
and efficiency in the context of kinodynamic planning. Based on this
outline, the description of \sst\ and \sststar\ is then provided, as
well as an accompanying nearest neighbor data structure, which allows
the removal of nodes to achieve a sparse tree. The description of the
algorithms is followed by the comprehensive analysis of the described
methods in Section \ref{sec:analysis}. Simulation results on a series
of systems, including kinematic ones, where comparison with \rrtstar\
is possible, as well as benchmarks with interesting dynamics are
available in Section \ref{sec:experiments}. A physically simulated
system is also considered in the same section. Finally, the paper
concludes with a discussion in Section \ref{sec:discussion}.

\section{Background}
\label{sec:background}
{\bf Planning Trajectories:} Trajectory planning for real robots
requires accounting for dynamics (e.g., friction, gravity, limits in
forces).  It can be achieved either by a decoupled
approach \citep{Bobrow1985Time-Optimal-Co, Shiller1991On-Computing-th}
or direct planning. The latter method searches the state space
of a dynamical system directly.  For underactuated, non-holonomic systems,
especially those that are not small-time locally controllable ({\tt
STLC}), the direct planning approach is preferred. The focus here is
on systems that are not {\tt STLC} but are small-time locally
accessible \citep{Chow1939Uber-Systemen-v}. The following
methodologies have been considered in the related literature for
direct planning:

\begin{myitem}

\item[-] \emph{Optimal control} can be applied 
\citep{Brockett1982Control-Theory-, lewis1995optimal-control} but
handles only simple systems.  Algebraic solutions are available
primarily for 2D point mass
systems \citep{ODunlaing1987Motion-Planning,
Canny1991An-Exact-Algori}.

\item[-] \emph{Numerical optimization} \citep{Fernandes1993Optimal-Non-hol,
betts1998survey-of-numer, Ostrowski2000Optimal-Gait-Se} can be used
but it can be expensive for global trajectories and suffers from local
minima. There has been progress along this
direction \citep{Zucker2013CHOMP:-Covarian,
Schulman2014Motion-Planning}, although highly-dynamic problems are
still challenging.

\item[-] Approaches that take advantage of \emph{differential flatness}
allow to plan for dynamical systems as if they are high-dimensional
kinematic ones \citep{Fliess1995Flatness-and-De}. While interesting
robots, such as quadrotors \citep{Sreenath2013Geometric-Contr}, can be
treated in this manner, other systems, such as fixed-wing airplanes,
are not amenable to this approach.

\item[-] \emph{Search-based methods} compute  paths over
discretizations of the state space but depend exponentially on the
resolution \citep{Sahar1985Planning-of-Min,
Shiller1988Global-Time-Opt, Barraquand1993Nonholonomic-Mu}. They also
correspond to an active area of research, including for systems with
dynamics \citep{Likhachev2009Planning-Long-D}.
 
\end{myitem}
A polynomial-time, search-based approximation framework introduced the
notion of ``kinodynamic'' planning and solved it for a dynamic point
mass \citep{Donald1993Kinodynamic-Mot}, which was then extended to
more complicated systems \citep{heinzinger1989time-optimal-tr,
donald1995provably-good-a}. This work influenced sampling-based
algorithms for kinodynamic planning.

\noindent {\bf Sampling-based Planners:} These algorithms avoid 
explicitly representing configuration space obstacles, which is
computationally hard. They instead sample vertices and connect them
with local paths in the collision-free state space resulting in a
graph data structure. The first popular sampling-based algorithm,
the Probabilistic Roadmap Method
(\prm) \citep{Kavraki1996Probabilistic-R}, precomputes a roadmap using
random sampling, which is then used to answer multiple
queries.  \rrtconnect\ \edited{returns} a tree and focuses on quickly answering
individual queries \citep{lavalle_kuffner_rrtconnect}. Bidirectional
tree variants achieve improved
performance \citep{Sanchez2001A-single-query-}. All these solutions
require a steering function, which connects two states with a local
path ignoring obstacles. For systems with symmetries it is possible to
connect bidirectional trees by using numerical methods for bridging
the gap between two states \citep{cheng_gap, lamiraux_gap}.

Two sampling-based methods that do not require a steering function
are \rrtextend\ \citep{LaValle2001} and {\tt Expansive Space Trees}
(\est) \citep{Hsu2002}. They only propagate dynamics forward in time
and aim to evenly and quickly explore the state space regardless of
obstacle placement. For all of the above methods, probabilistic
completeness can be argued under certain conditions
(\cite{Kavraki1998Analsis-of-Prob, Hsu1998On-finding-narr,
Ladd2004Measure-Theoret}). Variants of these approaches aim to
decrease the metric dependence by reducing the rate of failed node
expansions \citep{Cheng2001Reducing-Metric}, or applying adaptive
state-space subdivision \citep{Ladd2005Motion-Planning}.  Others guide
the tree using heuristics \citep{Bekris2008Informed}, local
reachability information \citep{Shkolnik2009Reachability-gu},
linearizing locally the dynamics to compute a
metric \citep{Glassman2010A-quadratic-reg}, learning
the \emph{cost-to-go} to balance or bias
exploration \citep{Li2009Balancing-State, Li2011Learning-Approx}, or
by taking advantage of grid-based discretizations \citep{Plaku:2010fk,
Sucan2012A-Sampling-Base}.  Such tree-based methods have been applied
to various interesting domains \citep{Frazzoli2002Real-Time-Motio,
BranickyHybrid2006, ZuckerMultipleRRT2007}. While \rrt\ is effective
in returning a solution quickly, it converges to a sub-optimal
solution \citep{Nechushtan2010SamplingDiagramsAutomata}.


\noindent {\bf From Probabilistic Completeness to Asymptotic
Optimality:} Some \rrt\ variants have employed heuristics to improve
path quality but are not provably
optimal \citep{Urmson2003Approaches-for-}, including anytime
variants \citep{Ferguson2006AnytimeRRT}. Important progress was achieved through the
utilization of random graph theory to rigorously show that
roadmap-based approaches, such as \prmstar\ and \rrtstar, can achieve
asymptotic optimality \citep{Karaman2011Sampling-based-}.  The
requirement is that each new sample must be tested for connection with
at least a logarithmic number of neighbors as a function of the total
number of nodes using a steering function.
Anytime \citep{Karaman2011AnytimeRRT*} and
lazy \citep{Alterovitz2011Rapidly-Explori} variants of \rrtstar\ have
also been proposed.  There are also techniques that provide asymptotic
near-optimality using sparse roadmaps, which inspire the current
work \citep{Marble2011Asymptotically-, Marble2013Asymptotically-,
Dobson2012Sparse-Roadmap-, Dobson2013Sparse-Roadmap-,
Wang2013A-fast-streamin,Shaharabani2013Sparsification-}. Sparse trees
appear in the context of feedback-based motion
planning \citep{Tedrake2009LQR-trees:-Feed}. Another line of work
follows a Lazy \prmstar\ approach to improve
performance \citep{Janson2013Fast}. A conservative estimate of the
reachable region of a system can be
constructed \citep{Karaman2013Sampling-Based-}. This reachable region
helps to define appropriate metrics under dynamics, and can be used in
conjunction with the algorithms described here. All of the above
methods, which are focused on returning high-quality paths, require
a \bvp\ solver.

\begin{wrapfigure}{r}{.4\textwidth}
\vspace{-.3in}
\includegraphics[width = .39\textwidth]{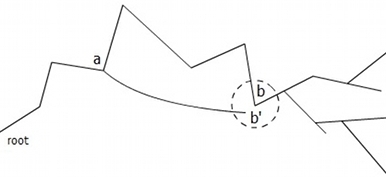}
\vspace{-.05in}
\caption{If  $b'$ is  close to 
$b$ and $cost(b') < cost(b)$, the shooting variant will prune the edge
to $b$ and replace it with $b'$. The subset of $b$ is repropagated
from $b'$.}
\vspace{-.05in}
\label{fig:shooting}
\end{wrapfigure}
\noindent {\bf Towards Asymptotic Optimality for Dynamical Systems:} A
variation of \rrtstar\ utilizes a ``shooting'' approach, shown in
Figure \ref{fig:shooting}, to improve solutions without a steering
function \citep{Jeon2011Anytime-Computa}. When propagating from node
$a$ to state $b'$ within a small distance of node $b$ and the cost to
$b'$ is smaller, $b$ is pruned and an edge from $a$ to $b'$ is
added. The subtree of $b$ is repropagated from $b'$, which may result
in node pruning if collisions occur. This method does not provably
achieve asymptotic optimality. It can be integrated with numerical
methods for decreasing the gap between $b$ and $b'$. The methods
presented here achieve formal guarantees. Improved computational
performance relative to the ``shooting'' variant is shown in the
experimental results.  Recent work provides local planners for systems
with linear or linearizable dynamics \citep{Webb2013Kinodynamic-RRT,
Goretkin:2013uq}.  There are also recent efforts on avoiding the use
of an exact steering function \citep{Jeon2013Optimal-Motion-}. The
algorithms in the current paper are applicable beyond systems with
linear dynamics but could also be combined with the above methods to
provide efficient asymptotically near-optimal solvers for such
systems.


\noindent {\bf Closely Related Contributions:} Early versions of the 
work presented here have appeared before.  Initially, a simpler
version of the proposed algorithms was proposed,
called \sparserrt\ \citep{Littlefield2013Efficient-sampl}. Good
experimental performance \edited{was achieved with this method, but it was not possible to formally argue
desirable properties.} This motivated the development
of \stable\ (\sst) and \edited{\sststar\ } in follow-up
work \citep{Li2014Sparse-Methods-}. These methods formally achieve
asymptotic (near)-optimality for kinodynamic planning. The same paper
was the first to introduce the analysis that is extended in the
current manuscript. Given these earlier efforts by the authors, this
paper provides the following contributions:
\begin{itemize}
\item It describes a \emph{general framework} for  asymptotic
(near-)optimality using sampling-based planners without a steering
function in Section \ref{sec:paradigm}. The \sst\ and \sststar\
algorithms correspond to efficient implementations of this framework.

\item It describes for the first time in Section \ref{sec:nearest}
a \emph{nearest neighbor data structure} that has been specifically
designed to support the pruning operation of the proposed
algorithms. \emph{Implementation guidelines} are introduced in the
description of \sst\ and \sststar\ that improve performance
(Sections \ref{sec:sst} and \ref{sec:sststar}).

\item Section \ref{sec:analysis} \emph{extends the analysis} by
arguing properties for a general cost function instead of trajectory
duration. It also provides all the necessary proofs that were missing
from previous work.

\item \emph{Additional experiments} are provided in
Section \ref{sec:experiments}, including simulations for a dynamical
model of a fixed-wing airplane.  There is also evaluation of the
effects the nearest neighbor data structure has on the motion planners.
\end{itemize}

There is also concurrent work \citep{Papadopoulos2014}, which presents
similar algorithms and argues experimentally that they return
high-quality trajectories for kinodynamic planning. It provides a
different way to support the argument that a simplification of \est,
i.e., the \rt\ approach, is asymptotically optimal. It doesn't argue,
however, the asymptotic near-optimality properties of the efficient
and practical methods that achieve a sparse representation, neither
studies the convergence rate of the corresponding algorithms nor
provides efficient tools for their implementation, such as the nearest
neighbor data structure described here.

\section{Problem Setup}
\label{sec:setup}
\noindent This paper considers dynamic systems that respect
time-invariant differential equations of the following form:
\vspace{-0.05in}
\begin{equation}
\dot{x}(t) = f(x(t), u(t)),\ \  x(t) \in \Xspace,\ \  u(t) \in \Uspace
\vspace{-0.05in}
\label{eq:dynamics}
\end{equation}

where $x(t)\in \Xspace \subseteq R^d$ and $u(t)\in \Uspace \subseteq
R^l$. The collision-free subset of $\Xspace$ is $\Xfree$. Let
$\mu(\Xspace)$ denote the Lebesgue measure of $\Xspace$. This work
focuses on state space manifolds that are subsets of $d$-dimensional
Euclidean spaces, which allow the definition of the $\mathbb{L}_2$
Euclidean norm $||.||$. The corresponding $r$-radius closed ball in
$\Xspace$ centered at $x$ will be $\ball_r(x)$. \edited{In other
words, the underlying state space needs to exhibit some smoothness
properties and behave locally as a Euclidean space.}

\vspace{-0.05in}
\begin{definition} (Trajectory)
A trajectory $\pi$ is a function $\pi(t):
[0,t_\pi] \rightarrow \Xfree$, where $t_\pi$ is its duration. A
trajectory $\pi$ is generated by starting at a given state $\pi(0)$
and applying a control function $\Upsilon:
[0,t_\pi] \rightarrow \Uspace$ by forward integrating
Eq. \ref{eq:dynamics}.
\end{definition}
\vspace{-0.05in}
\edited{Typically, sampling-based planners are implemented so that the applied
control function $\Upsilon$ corresponds to a piecewise constant one.
Such an underlying discretization is often unavoidable given the
presence of a digital controller. This is why the analysis provided in
this paper considers piecewise constant control functions, which are
otherwise arbitrary in nature.}
\begin{definition}
\vspace{-.05in}
{(Piecewise Constant Control Function)} A piecewise constant control
function $\bar\Upsilon$ with resolution $\Delta t$ is the
concatenation of constant control functions of the form $\Upsilon_i:
[0,k_i \cdot \Delta t] \rightarrow u_i$, where $u_i \in \Uspace$ and
$k_i \in \mathbb{Z}^+$.
\vspace{-.05in}
\end{definition}

\edited{The proposed methods and the accompanying analysis do not
critically depend on the piecewise constant nature of the input
control function. They could potentially be extended to also allow for
continuous control functions, such as those generated by splines or
using basis functions:}

\begin{wrapfigure}{l}{.30\textwidth}
\vspace{-.15in}
\includegraphics[width = .29\textwidth]{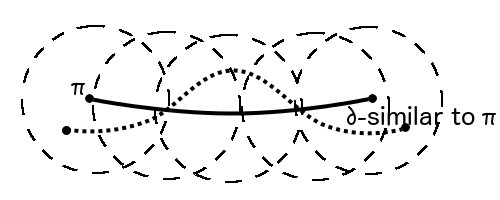}
\vspace{-.05in}
\caption{Two $\delta$-similar trajectories. }
\vspace{-.25in}
\label{fig:similar}
\end{wrapfigure}
\noindent A key notion for this work is illustrated in
Figure \ref{fig:similar} and explained below:
\vspace{-.05in}
\begin{definition} ($\delta$-Similar Trajectories)
Trajectories $\pi$, $\pi^\prime$ are $\delta$-similar if for a
continuous, nondecreasing scaling function $\sigma:
[0,t_{\pi}]\rightarrow[0,t_{\pi^\prime}]$, it is true that
$\pi^\prime(\sigma(t)) \in \ball_\delta(\pi(t))$. 
\vspace{-.05in}
\label{def:simliar}
 \end{definition}

\noindent The focus in this paper will be initially on optimal trajectories with
a certain clearance from obstacles.

\vspace{-.05in}
\begin{definition} (Obstacle Clearance) The obstacle clearance
$\epsilon$ of a trajectory $\pi$ is the minimum distance from
obstacles over all states in $\pi$, i.e., $\epsilon =
\inf_{t\in[0,t_{\pi}],x_o \in \Xspace_o}\ ||\pi(t) - x_o||$, where
$\Xspace_o = \Xspace \setminus \Xfree$.
\vspace{-.05in}
\end{definition}

\begin{wrapfigure}{r}{.28\textwidth}
\vspace{-0.35in}
\includegraphics[width = .27\textwidth]{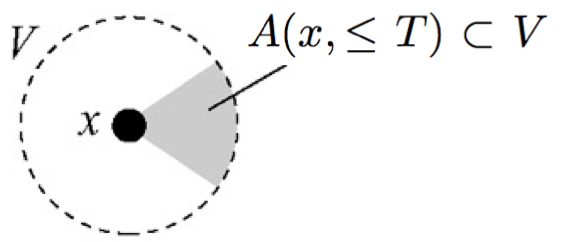}
\caption{The {\tt STLA} property. }
\vspace{-.3in}
\label{fig:stla}
\end{wrapfigure}
\noindent Then, the following assumption is helpful for the methods and the analysis.
\noindent \begin{assumption}
\vspace{-.05in}
The system described by Equation \ref{eq:dynamics} satisfies the
properties:

\begin{myitem}

\item \emph{Chow's}
condition \citep{Chow1939Uber-Systemen-v} of Small-time Locally
Accessible ({\tt STLA}) systems \citep{Choset2005Principles-of-R}:
For {\tt STLA} systems, it is true that the reachable set of states
$A(x,\leq T) \subset V $ from any state $x$ in time less than or equal
to $T$ without exiting a neighborhood $V \subset \Xspace$ of $x$, and
for any such $V$, has the same dimensionality as $\Xspace$.

\item It has bounded second derivative: $|\ddot{x}(t)| \leq M_2 \in R^+$.


\item It is \emph{Lipschitz continuous} for both of
its arguments, i.e., $\exists\ K_u > 0$ and $\exists\ K_x>0$:
\vspace{-.05in}
$$|| f(x_0, u_0) - f(x_0, u_1) || \leq K_u || u_0 - u_1||,\ \ \ \ \ \ \ \
\ \ \ || f(x_0, u_0) - f(x_1, u_0) || \leq K_x || x_0 -  x_1||.$$
\end{myitem}
\vspace{-.1in}
\label{ass:system}
\end{assumption}

\noindent The assumption that $f$ satisfies \emph{Chow's} condition implies
there always exist $\delta$-similar trajectories for any trajectory $\pi$.

\begin{lemmma}
\vspace{-.05in}
 Let there be a trajectory $\pi$ for a system satisfying
 Eq. \ref{eq:dynamics} and \emph{Chow's condition}. Then there exists
 a positive value $\delta_0$ called the {\bf dynamic clearance}, such that: $\forall\ \delta \in
 (0,\delta_0]$, $\forall\ x_{0}^\prime\in\ball_\delta(\pi(0))$, and
 $\forall\ x_{1}^\prime\in\ball_\delta(\pi(t_{\pi}))$, there exists a
 trajectory $\pi^\prime$, so that: (i) $\pi^\prime(0)=x_{0}^\prime$
 and $\pi^\prime(t_{\pi^\prime})=x_1^\prime$; (ii) $\pi$ and $\pi^{\prime}$
 are $\delta$-similar trajectories.
\vspace{-.05in}
\label{lem:closeexist}
\end{lemmma}

Lemma \ref{lem:closeexist} on the existence of ``dynamic clearance'' is
a necessary condition for all systems where sampling-based methods
work, such as \est, \rrt, and \rrtstar, are able to find a solution. A
proof sketch of Lemma \ref{lem:closeexist} can be found in Appendix A.
The interest is on trajectories with both good obstacle and dynamic
clearance, called $\delta$-robust trajectories.



\begin{definition} 
\vspace{-.05in}
($\delta$-Robust Trajectories) A trajectory $\pi$ for a dynamical
system following Eq. \ref{eq:dynamics} is called $\delta$-robust if both its
obstacle clearance $\epsilon$ and its dynamic clearance $\delta_0$ are
greater than $\delta$.
\label{def:delta_traj}
\vspace{-.05in}
\end{definition}


This paper aims to solve a variation of the motion planning problem
with dynamics for such optimal trajectories.

\begin{definition}
\vspace{-.05in}
($\delta$-Robust Feasible Motion Planning)
Given a dynamical system following Eq. \ref{eq:dynamics}, the collision-free
subset $\Xfree \subset \Xspace$, an initial state $\xinit \in \Xfree$,
a goal region $\Xgoal \subset \Xfree$, and that a $\delta$-robust trajectory
that connects $\xinit$ with a state in $\Xgoal$ exists, find
a solution trajectory $\pi$ for which $\pi(0) = \xinit$ and
$\pi(t_{\pi}) \in \Xgoal$.
\label{def:problem}
\vspace{-.05in}
\end{definition}

It will be necessary to assume that the problem can be solved using
trajectories generated by piecewise constant control functions. This
is a reasonable way to generate a trajectory using a computational
approach.

\begin{assumption}
\vspace{-.05in}
For a $\delta$-robust feasible motion planning problem, there exists a
$\delta$-robust trajectory $\pi$ generated by a piecewise constant
control function $\bar\Upsilon$.
\label{ass:piecewise}
\vspace{-.05in}
\end{assumption}


An incremental sampling-based algorithm, abbreviated here as $ALG$,
typically extends a graph data structure of feasible trajectories
over multiple iterations.  This paper considers the following
properties of such sampling-based planners.

\begin{definition}
\vspace{-.05in}
(Probabilistic $\delta$-Robust Completeness) Let $\Pi\ofalg$ denote
the set of trajectories discovered by an algorithm
$ALG$ at iteration $n$. Algorithm $ALG$ is probabilistically
$\delta$-robustly complete, if for any $\delta$-robustly feasible
motion planning problem ($\Xfree$, $\xinit$, $\Xgoal$, $\delta$) the
following holds:
\vspace{-0.1in}
\begin{align*}
&\liminf_{n\to\infty}\Prob(\ \exists\ \pi \in \Pi\ofalg: \pi \textrm{
solution to } (\Xfree, \xinit, \Xgoal, \delta) ) = 1.
\end{align*}
\vspace{-.3in}
\label{def:clearcomplete}
\vspace{-.05in}
\end{definition}

\noindent Definition \ref{def:clearcomplete} relaxes the concept
of \emph{probabilistic completeness} for algorithms with properties
that depend on the \emph{robust clearance} $\delta$ of trajectories
they can discover.  An algorithm that is \emph{probabilistically}
$\delta$\emph{-robustly complete} only demands it will eventually find
solution trajectories if one with robust clearance of $\delta$ exists.
The following discussion relates to the cost function of a trajectory
$\pi$.

\begin{assumption}
\vspace{-.05in}
 The cost function $cost(\pi)$ of a trajectory is assumed to be
\emph{Lipschitz continuous}. Specifically, $\exists\ K_c > 0$:
\vspace{-0.05in}
$$|{\tt cost}(\pi_0) - {\tt cost}(\pi_1)|\leq K_c\cdot sup_{\forall
t} \{ || \pi_0(t) - \pi_1(t) || \}, \vspace{-0.05in}$$
for all $\pi_1$, $\pi_2$ with the same start state.
Consider two trajectories $\pi_1,\pi_2$ such that their concatenation is $\pi_1|\pi_2$ (i.e., following trajectory $\pi_2$ after
trajectory $\pi_1$), the cost function satisfies:
\begin{myitem}
\item $cost(\pi_1|\pi_2) = cost(\pi_1) + cost(\pi_2)$ (additivity)
\item $cost(\pi_1) \leq cost(\pi_1|\pi_2)$ (monotonicity)
\item $\forall\ t_2 > t_1 \geq 0$, $\exists M_c > 0$,  $t_2-t_1\leq M_c \cdot |cost(\pi(t_2)) - cost(\pi(t_1))|$ (non-degeneracy)
\end{myitem}
\label{ass:cost}
\vspace{-.05in}
\end{assumption}

\noindent Then, it is possible to relax the property
of \emph{asymptotic optimality} and allow some tolerance depending on
the clearance.

\begin{definition}
\vspace{-.05in}
(Asymptotic $\delta$-robust Near-Optimality) Let $c^*$ denote the
minimum cost over all solution trajectories for a $\delta$-robust
feasible motion planning problem ($\Xfree$, $\xinit$, $\Xgoal$,
$\delta$). Let $Y\ofalg$ denote a random variable that represents the
minimum cost value among all trajectories returned by algorithm
$ALG$ at iteration $n$ for the same problem.  $ALG$ is asymptotically
$\delta$-robust near-optimal if for all independent runs:
\vspace{-0.05in}
$$\Prob(\Big\{\limsup_{n\to\infty}Y_n^{ALG}\leq
h(c^\ast,\delta)\Big\})=1 \vspace{-0.05in}$$ where
$h: \reals \times \reals \rightarrow \reals$ is a function of the
optimum cost and the $\delta$ clearance, where $h(c^\ast,\delta) \geq
c^\ast$.
\label{def:nearoptimal}
\vspace{-.05in}
\end{definition}

\noindent
The analysis will show that the proposed algorithms exhibit the above
property where $h$ has the form: $h(c^\ast,\delta) = (1
+ \alpha \cdot \delta) \cdot c^\ast$ for some constant $\alpha \geq
0$. In this case, $ALG$ is asymptotically $\delta$-robust
near-optimal with a multiplicative error.  Note that for this form of
the $h$ function, the absolute error relative to the optimum cost increases as
the optimum cost increases. This property guarantees that the
cost of the returned solution is upper bounded relative to the optimal
cost.  Recall that \rrtconnect\ returns solutions of random cost and
the error is unbounded \citep{Karaman2011Sampling-based-}.

If it is possible to argue that an algorithm satisfies the last two
properties for all decreasing values of the robust clearance
$\delta$, then this algorithm satisfies the traditional properties of
probabilistic completeness and asymptotic optimality.

{\bf Regarding Distances:} The true cost of moving between two states
corresponds to the ``cost-to-go'', which typically does not satisfy
symmetry, is not the Euclidean distance, and is not easy to
compute. Based on the ``cost-to-go'', it is possible to define an
$\epsilon$-radius sub-riemannian ball centered at $x$, which is the
set of all states where the ``cost-to-go'' from $x$ to that set is less
than or equal to $\epsilon$.  The analysis \edited{presented}, which reasons primarily
over Euclidean hyper-balls, will show that there always exists a
certain size Euclidean hyper-ball inside the sub-riemannian ball under
the above conditions. Therefore, it will be sufficient to reason about
Euclidean norms. In practice, distances may be taken with respect to a
different space, which reflect the application, and may actually be
closer to the true ``cost-to-go'' for the moving system.

\section{Algorithms}
\label{sec:algos}
This section provides sampling-based tree motion planners that achieve
the properties of Definitions \ref{def:clearcomplete}
and \ref{def:nearoptimal} for kinodynamic planning when there is no
access to a \bvp\ solver. First a general framework is described for
this purpose, and then an instantiation of this framework is given
(\sst), which is extended to an asymptotically optimal algorithm
(\sststar). 

\vspace{-.1in}
\subsection{Change in Algorithmic Paradigm}
\label{sec:paradigm}

\noindent {\bf Traditional Approach:} Given the difficulty of
kinodynamic planning (\cite{Donald1993Kinodynamic-Mot}), the early but
practical tree-based planners \citep{lavalle_kuffner_rrt, Hsu2002}
aimed for even and fast exploration of $\Xspace$ even in challenging
high-dimensional cases where greedy, heuristic expansion towards the
goal would fail. Given that computing optimal trajectories corresponds
to an even harder challenge, the focus was not on the quality of the
returned trajectory in these early methods.

\begin{algorithm}
\caption{{\tt EXPLORATION\_TREE}($\Xspace$, $\Uspace$, $\xinit$,
$T_{prop}$, $N$)}
\label{alg:exploration_tree}
$G = \{ \vertx \leftarrow \{\xinit\}, \edges \leftarrow \emptyset \}$\;
\For{$N$ iterations}
{
	$x_{selected} \leftarrow$ {\tt Exploration\_First\_Selection}($\vertx, \Xspace$)\;
	$x_{new} \leftarrow$ {\tt Fixed\_Duration\_Prop}($x_{selected}$, $\Uspace$, $T_{prop}$)\;
	\If{ {\tt CollisionFree}$(\overline{x_{selected}\to x_{new}})$}
	{
		$\vertx \leftarrow \vertx \cup \{x_{new} \}$\;
		$\edges \leftarrow \edges \cup \{ \overline{x_{selected} \to x_{new}} \}$\;
	}
}
{\bf return} $G( \vertx, \edges )$\;
\end{algorithm}

Algorithm \ref{alg:exploration_tree} summarizes the high-level
selection/propagation operation of these planners. They constructed a
graph data structure $G(V,E)$ in the form of a tree rooted at an
initial state $x_0$ in the following two-step process:

\begin{itemize} 

\item {\it Selection:} A reachable state along the tree, such as a node
 $x_{selected} \in V$, is selected. In some variants a state along an
edge of the tree can also be
selected \citep{Ladd2005Fast-Tree-Based}. The selection process is
designed so as to increase the probability of searching underexplored
parts of $\Xspace$. For instance, the \rrtextend\ algorithm samples a
random state $x_{rand}$ and then selects the closest node on the tree
as $x_{selected}$.  The objective is to achieve a ``Voronoi-bias''
that promotes exploration, i.e., nodes on the tree that correspond to
the largest Voronoi regions of $\Xspace$, given tree nodes as sites,
have a higher probability of being selected \footnote{A tree-based
planner without access to a \bvp\ solver cannot guarantee a
``Voronoi-bias'' in general. If the distance function can correctly
estimate the \emph{cost-to-go} and if the propagation behaves
similarly to the steering function, then the ``Voronoi-bias'' is
achieved.}. In \est\ implementations, nodes store the local density of
samples and those with low density are selected with higher
probability to promote exploration \citep{phillips2004}.

\item {\it Propagation:} The procedure for extending the tree 
has varied in the related literature but the scheme followed
in \rrtextend\ has been popular in most implementations. The approach
is to select a control that drives the system towards the randomly
sampled point, then forward propagate that control input for a fixed
time duration. If the resulting trajectory $\overline{x_{selected}\to
x_{new}}$ is collision-free, then it is added as an edge in the
tree. It was recently shown that this propagation scheme actually
makes \rrtextend\ lose its probabilistic completeness
guarantees \citep{Kunz2014Kinodynamic-RRT}.  In \est, a randomized
approach is employed where random controls are used. The analysis of
the proposed methods shows that a randomized approach has benefits in
terms of solution quality.

\end{itemize}

\noindent {\bf Challenge:} Optimality has only recently become the
focus of sampling-based motion planning, given the development of the
asymptotically optimal \rrtstar\
and \prmstar\ \citep{Karaman2011Sampling-based-}. This great progress,
however, does not address kinodynamic planning instances.  Both
planners are roadmap-based methods in the sense that they reason over
(in the case of \rrtstar) or explicitly construct (in the case
of \prmstar) a graph that makes use of a steering function to connect
states.  This raised the following research challenge in the
community:

\begin{center}
\emph{Is it even possible to achieve asymptotic optimality guarantees in
sampling-based kinodynamic planning?} 
\end{center}

This has been an open question in the algorithmic robotics community
and resulted in many methods that aim to provide asymptotic optimality
for systems with dynamics \citep{Karaman2013Sampling-Based-,
Webb2013Kinodynamic-RRT, Goretkin:2013uq,
Jeon2013Optimal-Motion-}. The majority of these techniques, however,
can address only specific classes of problems (e.g., systems with
linear dynamics) and do not possess the generality of the original
sampling-based tree planners. 

\noindent {\bf Progress:} The current work provides an answer to the
above open question through a comprehensive, novel analysis of
sampling-based processes for motion planning without access to a
steering function, which departs from previous analysis efforts in
this domain. In particular, the following are shown:

\begin{enumerate} 

\item \emph{It is possible to achieve asymptotic optimality in the
rather general setting of this paper's problem setup with a
sampling-based process that makes proper use of random forward
propagation and a na\"{\i}ve selection strategy.}

\item  \emph{This method, however, is computationally impractical and does not
have a good convergence rate to optimal solutions. Thus, the important
question is whether there are planners with practical convergence to
high-quality solutions.}

\item  \emph{Given this realization, this work describes a framework for
computationally efficient sampling-based planners that achieve asymptotic
near-optimality, which are then also extended to provide asymptotic
optimality.}

\end{enumerate}

\noindent {\bf Asymptotic Optimality from Random Primitives:} To
achieve these desirable properties it is necessary to clearly define
the framework which sampling-based algorithms should adopt. In
particular, it is possible to argue asymptotic optimality for the \rt\
process described in Algorithm \ref{alg:RT}. This algorithm follows
the same selection/propagation scheme of sampling-based tree planners
but applies uniform selection and calls the \mcprop\ procedure to
extend the tree.

\begin{algorithm}[h]
\caption{\rt($\Xfree$, $\Uspace$, $\xinit$, $T_{prop}$, $N$)}
\label{alg:RT}
$G = \{ \vertx \leftarrow \{\xinit\}, \edges \leftarrow \emptyset \}$\;
\For{$N$ iterations}
{
$x_{selected} \leftarrow$ {\tt Uniform\_Sampling}($\vertx$)\;
$x_{new} \leftarrow $\mcprop( $x_{selected}$, $\Uspace$, $T_{prop}$ )\;
\If{ {\tt CollisionFree}$(\overline{x_{selected}\to x_{new}})$}
{
$\vertx \leftarrow \vertx \cup \{x_{new} \}$\;
$\edges \leftarrow \edges \cup \{ \overline{x_{selected} \to x_{new}} \}$\;
}
}
{\bf return} $G( \vertx, \edges )$\;
\end{algorithm}

The \mcprop\ procedure described in Algorithm \ref{alg:mc-propagate}
is different than the {\tt Fixed\_Duration\_Prop} method that is
frequently followed in implementations of sampling-based tree
planners. The difference is that the duration of the propagation is
randomly sampled between 0 and a maximum duration $T_{prop}$ instead
of being fixed. The accompanying analysis
(Section \ref{sec:analysismcprop}) shows that this random process
provides asymptotic optimality when the only primitive to access the
dynamics is forward propagation.

\begin{algorithm}[h]
\caption{\mcprop($x_{prop}$, $\Uspace$, $T_{prop}$)}
\label{alg:mc-propagate}
$t \leftarrow$ Sample$(0, T_{prop})$; $\Upsilon \leftarrow$
Sample($\Uspace,t)$\;
{\bf return} $x_{new}\leftarrow$ $\int_0^t f(x(t), \Upsilon(t))\,dt + x_{prop} $\;
\end{algorithm}

\begin{wrapfigure}{r}{0.34\textwidth}
\vspace{-.1in}
\centering
    \includegraphics[width=0.33\textwidth]{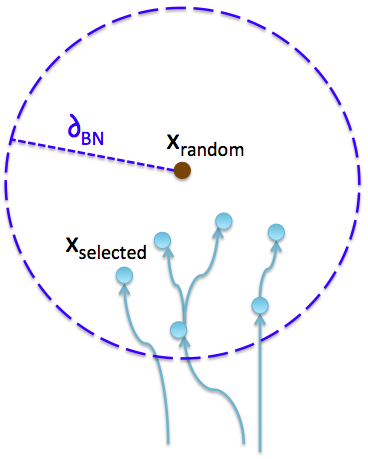} 
\vspace{-0.1in}
\caption{The
  selection of the best neighbor in \bestnear. The best path cost node
  in $\ball(x_{random},\Dnear)$ is selected.}
\label{fig:best_nearest}
\vspace{-.1in}
\end{wrapfigure}

Nevertheless, the \rt\ approach employs a na\"{\i}ve selection
strategy, where a node $x_{selected}$ is selected uniformly at
random. This has the effect that the resulting method does not have a
good convergence rate in finding high-quality solutions as a function
of iterations. It is not clear to the authors if a version of the \rt\
algorithm using an {\tt Exploration\_First\_Selection} strategy is
asymptotically optimal and most importantly \emph{whether it has
better convergence rate} properties, i.e., whether a method like \est\
or a version of \rrtextend\ that employs \mcprop\ are asymptotically
optimal with good convergence rate. The experimental indications
for \rrtextend\ with \mcprop\ are that it does not improve path
quality quickly.

\noindent {\bf Improving Convergence Rate:} A solution, however, has
been identified to this issue. In particular, the authors propose the
use of a {\tt Best\_First\_Selection} strategy as a desirable
alternative for node selection so as to achieve good convergence to
high-quality paths. In this context, best-first means that the node
$x_{selected}$ should be chosen so that the method prioritizes nodes
that correspond to good quality paths, while also balancing
exploration objectives. For instance, one way to achieve this in
an \rrt-like fashion (described in detail in the consecutive section)
is shown in Figure \ref{fig:best_nearest}, i.e., first sample a random
state $x_{random}$ and then among all the nodes on the tree within a
certain radius $\Dnear$, select the one that has the best path cost
from the root. A similar selection strategy has actually been proposed
in the past as a variant of \rrt\ that experimentally exhibited good
behavior \citep{Urmson2003Approaches-for-}. This previous work,
however, did not integrate this selection strategy with the \mcprop\ procedure 
and did not show any desirable properties for the resulting algorithm.

%
%

The analysis shows that the consideration of a best first strategy
together with the random propagation procedure leads to an
asymptotically $\delta$-robust near-optimal solution with good
convergence rate per iteration. This allows to observe improvement in
solution paths over time in practice. Nevertheless, there are additional
considerations to take into account when implementing a sampling-based
planner. In particular, the asymptotically dominant operation
computationally for these methods corresponds to nearest neighbor
queries. The implementation of {\tt Best\_First\_Selection} described
above and in Figure \ref{fig:best_nearest} requires the use of a range
query that is more expensive than the traditional closest neighbor
query in \rrt\ making the individual iteration cost of the proposed
solution more expensive. Consequently, the challenge becomes whether
this good convergence rate per iteration can be achieved, while also
reducing the running time for each iteration.

\noindent {\bf Balancing Computation Cost with Optimality:} The
property achieved with the {\tt Best\_First\_Selection} strategy is
that of asymptotic $\delta$-robust near-optimality. This means
that there should be an optimum trajectory $\pi^*$ in $\Xspace$ which has
$\delta$-robust clearance, as indicated in the problem setup. This
property also implies that it is not necessary to keep all samples as
nodes in the data structure so as to get arbitrarily close to
$\pi^*$. It is sufficient to have nodes that are in the vicinity of the path that is defined by its robust clearance
$\delta$. Thus, it is possible for a sparse data structure with a
finite set of states to sufficiently represent $\Xspace$ as long as it
can return $\delta$-similar solutions to all possible optimal
trajectories in $\Xspace$.

\begin{figure}[t]
\centering
    \includegraphics[width=\textwidth]{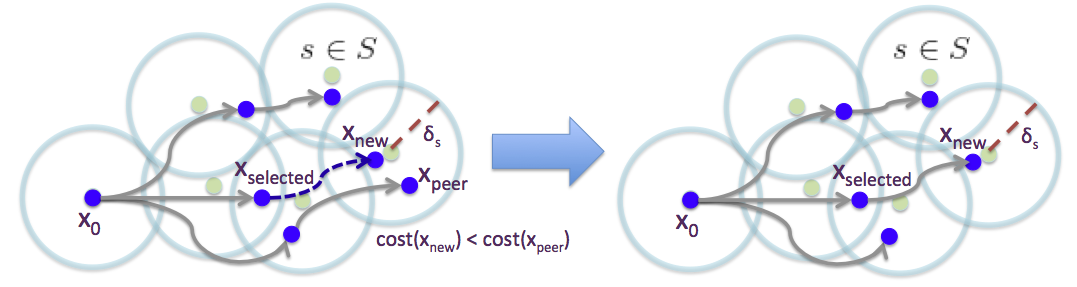}
\caption{The pruning operation to achieve a sparse data structure that
stores asymptotically near-optimal trajectories. Propagation from
$x_{selected}$ results to node $x_{new}$, which has a better path cost
than a node $x_{peer}$ in its local vicinity. Node $x_{peer}$ is
pruned and the newly propagated edge is added to the tree. If
$x_{peer}$ had children with the lowest path cost in their neighborhoods,
$x_{peer}$ would have remained in the tree but not considered for
propagation again. If $x_{new}$ had worse path cost than $x_{peer}$,
the old node would have remained in the tree and the last propagation
$\overline{x_{selected}\rightarrow x_{new}}$ would have been
ignored.}
\label{fig:pruning_1}
\vspace{-.1in}
\end{figure}

\begin{figure}[t]
\centering
    \includegraphics[width=\textwidth]{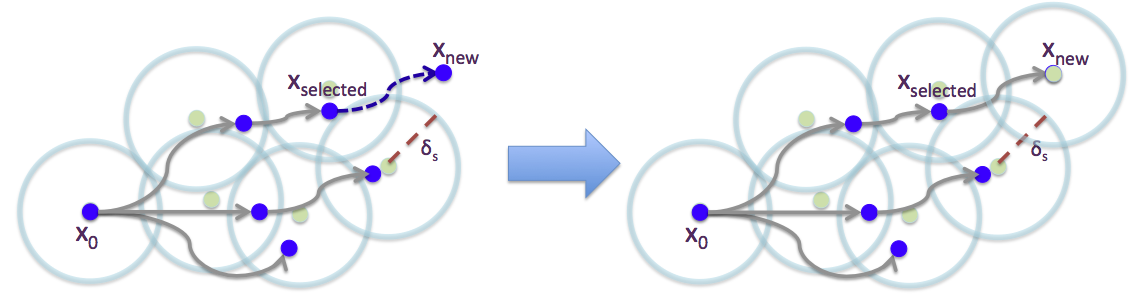}
\caption{Neighborhoods for pruning are defined based on a set of
static witness points $s \in S$, which are generated
incrementally. The indicated radii above and in
Figure \ref{fig:pruning_1} are centered in such witness points.  In
this figure, the propagation from $x_{selected}$ results in a node
$x_{new}$, which is not in the vicinity of an existing witness. In
this case, $x_{new}$ is not compared in terms of its path cost with
any existing tree node. The edge $\overline{x_{selected}\rightarrow
x_{new}}$ is added to the tree and a witness at the location of
$x_{new}$ is added to the set of witnesses $S$.}
\label{fig:pruning_2}
\vspace{-.1in}
\end{figure}

This allows for a pruning operation, where certain nodes can be
forgotten. Which trajectories should a sampling-based planner
maintain during its incremental operation and which ones should it
prune? The idea is motivated by the same objectives as that of the
{\tt Best\_First\_Selection} strategy and is illustrated in
Figures \ref{fig:pruning_1} and \ref{fig:pruning_2}. The pruning
operation should maintain nodes that correspond locally to good paths.
For instance, it is possible to evaluate whether a node has the best
cost in a local vicinity and prune neighbors with worse cost as long
as they do not have children with good path costs in their local
neighborhood. Nodes with high path cost in a local neighborhood do not
need to be considered again for propagation. There are many different
ways to define local neighborhoods. For instance, a grid-based
discretization of the space could be defined. In the accompanying
implementation and analysis, this work follows an incremental approach
of defining visited regions of the state space space as described in
Figure \ref{fig:pruning_2}.

Note that, with high probability, the pruned high-cost nodes would not
have been selected for propagation by the best first strategy
anyway. In this manner, the pruning operation reinforces the
properties of the {\tt Best\_First\_Selection} procedure in terms of
path quality. The accompanying analysis shows that the specific
pruning operation is actually maintaining the convergence properties
of the selection strategy. But it also provides significant
computational benefits.  Since the complexity of all the nearest
neighbor queries depends on the number of points in the data
structure, having a finite number of nodes, results in queries that
have bounded time complexity per iteration.  The benefits of sparsity
in motion planning have been studied over the last few years by some
of the authors \citep{Littlefield2013Efficient-sampl,
Dobson2013Sparse-Roadmap-} and others \citep{Wang2013A-fast-streamin,
Shaharabani2013Sparsification-}. The discussion section of this paper
describes the trade-offs that arise between computational efficiency
and the type of guarantee achieved in relation to the requirement for
the existence of $\delta$-robust trajectories.

\noindent {\bf A New Framework:} It is now possible to bring together
the recommended changes to the original sampling-based tree planners
and achieve a new framework for asymptotic near-optimality without a
steering function in a computationally efficient way, both in terms of
running time and memory requirements.  Table \ref{tab:framework} is
summarizing the differences between the original methods
(corresponding to the {\tt EXPLORATION\_TREE} procedure) and the
proposed framework for kinodynamic sampling-based planning. The new
framework is referred to as {\tt SPARSE\_BEST\_FIRST\_TREE} in
Algorithm \ref{alg:sparse_framework}.

\begin{table*}[h]
\begin{center}
\begin{tabular}{|| p{.09\textwidth} || p{.29\textwidth} || p{.22\textwidth} || p{.28\textwidth} ||}
\hline
 & {\tt EXPLORATION\_TREE} & \rt & {\tt
 SPARSE\_BEST\_FIRST\_TREE}\\ \hline
\hline
{\bf Selection} & {\tt Exploration\_First\_Selection} & {\tt Uniform\_Sampling} & {\tt Best\_First\_Selection}\\
 \hline
{\bf Propagation }& {\tt Fixed\_Duration\_Prop} & \mcprop & \mcprop \\
 \hline
{\bf Pruning }& N/A & N/A & {\tt Prune\_Dominated\_Nodes} \\
\hline
\hline
{\bf Properties} & Probabilistically Complete (under conditions),
Suboptimal but Computationally Efficient, Dense Data Structure &
Asymptotically Optimal but Bad Convergence Rate and Impractical, Dense
Data Structure & Asymptotically Near-Optimal with Good Convergence
Rate and Computationally Efficient with a Sparse Data Structure\\
\hline
\end{tabular}
\end{center}
\vspace{-.1in}
\caption{Outline of differences between the different
frameworks in terms of the modules they employ and their properties.}
\vspace{-.1in}
\label{tab:framework}
\end{table*}

\noindent In summary, the three modules of the new framework operate
as follows:

\begin{itemize} 
\item \emph{Selection:} The new framework still promotes the selection
of nodes in under-explored parts of $\Xspace$, as in the original
approaches, but within each local region only the nodes that
correspond to the best path from the root are selected.
\item \emph{Propagation:} The analysis accompanying this work emphasizes the
need to employ a fully random propagation process both in terms of the
selected control and duration of propagation, i.e., the \mcprop\
method, as in \est.
\item  \emph{Pruning:} Nodes that are locally dominated in terms of path cost
can be removed under certain conditions resulting in a sparse data
structure instead of storing infinitely many points.
\end{itemize}
\vspace{-.1in}

\begin{algorithm}[h]
\caption{{\tt SPARSE\_BEST\_FIRST\_TREE}($\Xfree$, $\Uspace$,
$\xinit$, $T_{prop}$, $N$)}
\label{alg:sparse_framework}
$G = \{ \vertx \leftarrow \{\xinit\}, \edges \leftarrow \emptyset \}$\;
\For{$N$ iterations}
{
	$x_{selected} \leftarrow$ {\tt Best\_First\_Selection}( $\vertx, \Xspace$)\;
	$x_{new} \leftarrow$ \mcprop( $x_{selected}$, $\Uspace$, $T_{prop}$ )\;

	\If{ {\tt CollisionFree}$(\overline{x_{selected}\to x_{new}})$}
	{
            \If{ {\tt Is\_Node\_Locally\_the\_Best}( $x_{new}$, $\vertx$ ) }
	    {
		$\vertx \leftarrow \vertx \cup \{x_{new} \}$\;
		$\edges \leftarrow \edges \cup \{ \overline{x_{selected} \to x_{new}} \}$\;

		{\tt Prune\_Dominated\_Nodes}( $x_{new}$, $G$ )\;
	    }
	}
}
{\bf return} $G( \vertx, \edges )$\;
\end{algorithm}

The following section provides an efficient instantiation of the {\tt
SPARSE\_BEST\_FIRST\_TREE} framework, which has been used both in the
theoretical analysis and the experimental evaluation of this
paper. This algorithm, called \stable\ (\sst), provides concrete
implementations of the {\tt Best\_First\_Selection}, {\tt
Is\_Node\_Locally\_the\_Best} and {\tt Prune\_Dominated\_Nodes}
procedures. The analysis shows that it is asymptotically near-optimal
with a good convergence rate and computationally efficient.

The near-optimality property stems from the consideration of
$\delta$-robust optimal trajectories. The existence of at least weak
$\delta$-robust clearance for optimal trajectories has been considered
in the related literature that achieves asymptotic optimality in the
kinematic case. To show asymptotic optimality for \rrtstar, one can
show that the requirement for the $\delta$ value reduces as the
algorithm progresses. The true value $\delta$ depends on the specific
problem to be solved and is typically not known beforehand. The way to
address this issue is to first assume an arbitrary value for $\delta$
and then repeatedly shrink the value for answering motion planning
queries. This is the approach considered here for extending \sst\ into
an asymptotically optimal approach \sststar.

\subsection{\stable\ (\sst) }
\label{sec:sst}
Algorithm \ref{alg:SST} provides a concrete implementation of the
abstract framework of {\tt SPARSE\_BEST\_FIRST\_TREE} outlined in the
previous section and corresponds to one of the proposed algorithms,
\stable\ (\sst), which is analyzed in the next section. 

At a high-level, \sst\ follows the abstract framework.  For $N$
iterations, a selection/propagation/pruning procedure is followed. The
selection follows the principle of the best first strategy to return
an existing node on the tree $x_{selected}$ (line 5). Its concrete
implementation is described in detail here. Then \mcprop\ is called
(line 6), which samples a random control and a random duration and
then integrates forward the system dynamics according to
Eq. \ref{eq:dynamics}. If the path $\overline{x_{selected}\rightarrow
x_{new}}$ is collision-free (line 7), the new node $x_{new}$ is
evaluated on whether is the best node in terms of path cost in a local
neighborhood (line 8). If $x_{new}$ is indeed better, it is added to
the tree (lines 9-10) and any previous node in the same local vicinity
that is dominated, is pruned (line 11).

\begin{algorithm}[t]
\caption{\stable( $\Xspace$, $\Uspace$, $\xinit$, $T_{prop}$, $N$,
$\Dnear$, $\Ddrain$)}
\label{alg:SST}
$\Vactive \leftarrow \{\xinit\},\Voff \leftarrow \emptyset$\;
$G = \{ V \leftarrow (\Vactive \cup \Voff), \edges \leftarrow \emptyset\}$\;
$s_0 \leftarrow \xinit$, $s_0.rep = \xinit$, $S\leftarrow \{s_0\}$\;
\For{$N$ iterations}
{
	$x_{selected}\leftarrow${\tt Best\_First\_Selection\_SST}( $\Xspace$, $\Vactive$, $\Dnear$)\;
	$x_{new}\leftarrow$ \mcprop($x_{selected}$, $\Uspace$, $T_{prop}$)\;
	
	\If{{\tt CollisionFree}$(\overline{x_{selected}\to x_{new}})$}
	{
		\If{{\tt Is\_Node\_Locally\_the\_Best\_SST}($x_{new}$, $S$, $\Ddrain$)}
		{
			$\Vactive\leftarrow \Vactive\cup\{x_{new}\}$\;
			$\edges\leftarrow\edges\cup\{ \overline{x_{selected}\to x_{new}} \}$\;
                        {\tt Prune\_Dominated\_Nodes\_SST}($x_{new}$, $\Vactive$, $\Voff$, $\edges$ )\;
		}
	}
}
{\bf return} $G$\;
\end{algorithm}

The new aspects of the approach introduced by the concrete
implementation are the following:

i) \sst\ requires an additional input parameter $\Dnear$, used in the
selection process of the {\tt Best\_First\_Selection\_SST} procedure
shown in Alg. \ref{alg:bestNearest}, inspired from previous
work \citep{Urmson2003Approaches-for-}.

ii) \sst\ requires an additional input parameter $\Ddrain$, used to
evaluate whether a newly generated node $x_{new}$ has locally the best
path cost in the {\tt Is\_Node\_Locally\_the\_Best\_SST} procedure of
Alg. \ref{alg:is_best}, useful for pruning.

iii) \sst\ splits the nodes of the tree $\vertx$ into two subsets:
$\Vactive$ and $\Voff$.  The nodes in $\Vactive$ correspond to nodes
that in a local neighborhood have the best path cost from the
root. The nodes $\Voff$ correspond to dominated nodes in terms of path
cost but have children with good path cost in their local
neighborhoods and for this reason are maintained on the tree for
connectivity purposes. Lines 1 and 2 of Algorithm \ref{alg:SST}
initialize the sets and the graph data structure
$G(\vertx,\edges)$, which will be returned by the algorithm. Only nodes
in $\Vactive$ are considered for propagation and participate in the
{\tt Best\_First\_Selection\_SST} procedure (line 5). These two sets
are updated when a new state $x_{new}$ is generated that dominates its
local neighborhood and pruning is performed (lines 9 and 11).

iv) In order to define local neighborhoods, \sst\ uses an auxiliary
set of states, called ``witnesses'' and denoted as $S$. The approach
maintains the following invariant with respect to $S$: for every
witness $s$ kept in $S$, a single node in the tree will represent that
witness (stored in the field $s.rep$ of the corresponding witness),
and that node will have the best path cost from the root within a
$\Ddrain$ distance of the witness $s$. All nodes generated within
distance $\Ddrain$ of the witness $s$ with a worse path cost then
$s.rep$ are removed from $\Vactive$, thereby resulting in a sparse
data structure. Line 3 of Algorithm \ref{alg:SST} initializes the set
$S$ to correspond to the root state of the tree, which becomes its own
representative. The set $S$ is used by the {\tt
Is\_Node\_Locally\_the\_Best\_SST} procedure to identify whether the
newly generated sample $x_{new}$ is dominating the
$\Ddrain$-neighborhood of its closest witness $s \in S$. The same
procedure is responsible for updating the set $S$.

There are two input parameters to \sst, $\Dnear$ and
$\Ddrain$. $\Dnear$ influences the number of nodes that are considered
when selecting nodes to extend. The larger this parameter is, the more
likely that exploration will be ignored and path quality will take
precedent. For this reason, care must be taken to not make $\Dnear$
too large. $\Ddrain$ is the parameter responsible for performing
pruning and providing a sparse data structure. As with $\Dnear$, there
is a tradeoff with $\Ddrain$. The larger this parameter is, the more
pruning will be performed, which helps computationally but then
problems may not be solved if it is not possible to sample inside
narrow passages. Given the analysis that follows, these two parameters
need to satisfy the relationship specified in the following
proposition:

\begin{prop}
\label{prop:deltas}
The parameters $\Dnear$ and $\Ddrain$ need to satisfy the following
relationship given the robust clearance $\delta$ of the
$\delta$-robust feasible motion planning problem that needs to be solved:
\vspace{-.05in}
$$\Dnear + 2 \cdot \Ddrain < \delta$$.
\vspace{-.05in}
\end{prop}

\begin{figure}[h]
\centering
\includegraphics[width = 6.5in]{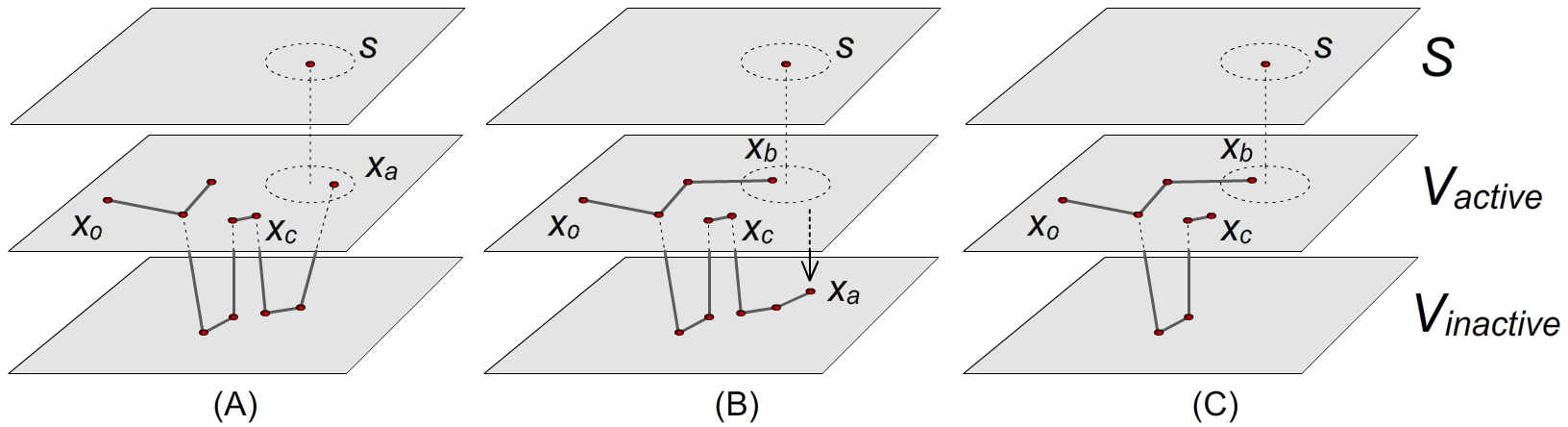}
\vspace{-0.1in}
\caption{Relation between $S$, $V_{active}$, and
  $V_{inactive}$. ($A$) A tree and a trajectory $\overline{x_{0} \to
  x_c \to x_a}$ where $x_a$ is the representative of $s$; Some of the
  nodes along this path are locally dominated in terms of path cost
  and exist in the $\Voff$ set. They remain in the tree, however,
  because $x_a$ is a representative. ($B$) The algorithm extends a new
  trajectory $\overline{x_{0} \to x_b}$ where $x_b$ has better cost
  than $x_a$. Then, $x_a$ is removed from $V_{active}$ and inserted
  into $V_{inactive}$. ($C$) The representative of $s$ is now
  $x_b$. The leaf trajectory $\overline{x_c\to x_a}$ that lies in
  $\Voff$ is recursively removed because all of these nodes are
  dominated and have no longer any children in the active set.}
\label{fig:sVV}
\end{figure}

Figure \ref{fig:sVV} summarizes the relationship between sets
$\Vactive$, $\Voff$ and $S$ in the context of the algorithm. The
following discussion outlines the implementation of the three
individual functions for the best first selection and the pruning
operation.

\noindent {\bf Best First Selection for \sst:} Algorithm \ref{alg:bestNearest}
outlines the operation. The method first samples a random point
$x_{rand}$ in the state space $\Xspace$ (line 1) and then finds a set
of states $X_{near}$ within distance $\Dnear$ of $x_{rand}$ (Line 2).
If the set $X_{near}$ is empty, then \bestnear\ defaults to using the
nearest neighbor to the random sample as in \rrt\ (line 3). Among the
states in $X_{near}$, the procedure will select the vertex that
corresponds to the lowest trajectory cost from the root of the tree
$x_0$ (Line 4).

\begin{algorithm}[h]
\caption{{\tt Best\_First\_Selection\_SST}($\Xspace$, $\vertx$, $\Dnear$)}
\label{alg:bestNearest}
$x_{rand} \leftarrow$ {\tt Sample\_State}($\Xspace$)\;
$X_{near} \leftarrow${\tt Near}$(\vertx,x_{rand},\Dnear)$\;
{\bf If} $X_{near} = \emptyset\ $ {\bf return} {\tt
Nearest}$(\vertx,x_{rand})$\;
{\bf Else} {\bf return} $\arg\min_{x \in X_{near} } cost(x)$\;
\end{algorithm}

Relative to \rrtstar, this method also uses a neighborhood and tries
to propagate a node along the best path from the root.
Nevertheless, \rrtstar\ propagates the closest node to $x_{rand}$ and
then attempts connections between all nodes in $X_{near}$ set to the
new state. These steps require multiple calls to a steering
function. Here, a near-optimal node in a neighborhood of the random
sample is directly selected for propagation, which is possible without
a steering function but only using a single forward propagation of the
dynamics. A procedure similar to \bestnear\ was presented as a
heuristic version of \rrt\ in previous
work \citep{Urmson2003Approaches-for-}. Here it is formally analyzed
to show its mathematical guarantees in terms of path quality and
convergence properties.

\noindent {\bf Pruning in \sst:} Algorithm \ref{alg:is_best} describes
the conditions under which the newly propagated node $x_{new}$ is
considered for addition to the tree. First, the closest witness
$s_{new}$ to $x_{new}$ from the set $S$ is computed (line 1). If the
closest witness is more than $\Ddrain$ away, then the sample $x_{new}$
becomes a new witness itself (lines 2-5). The representative of the
witness $s$ is stored in the variable $x_{peer}$ (line 6). Then the
new sample $x_{new}$ is considered viable for addition in the tree, if
at least one of two conditions holds (line 7): i) there is no
representative $x_{peer}$, i.e., the sample $x_{new}$ was just added
as a witness or ii) the cost of the new sample $cost(x_{new})$ is less
than the cost of the witness' representative $cost(x_{peer})$. If the
function returns true, node $x_{new}$ is added to the tree and the
active set of nodes $\Vactive$. If not, then the last propagation is
ignored.

\begin{algorithm}[h]
\caption{{\tt Is\_Node\_Locally\_the\_Best\_SST}($x_{new}$, $S$, $\Ddrain$)}
\label{alg:is_best}
$s_{new} \leftarrow$ {\tt Nearest}($S$,$x_{new}$)\;
\If{$||x_{new} - s_{new}|| > \Ddrain $}
{
	$S \leftarrow S \cup \{x_{new}\}$\;
	$s_{new} \leftarrow x_{new}$\;
	$s_{new}.rep \leftarrow NULL$\;
}
$x_{peer}\leftarrow s_{new}.rep$\;
\If{$x_{peer}==NULL$ or {\tt cost}($x_{new}$) $\ <\ ${\tt cost}($x_{peer}$)}
{
    {\bf return} true\;
}
{\bf return} false\;
\end{algorithm}

Algorithm \ref{alg:prune_dominated} describes the pruning process of
dominated nodes when \sst\ is adding node $x_{new}$. First the witness
$s_{new}$ of the new node and its previous representative $x_{peer}$
are found (lines 1-2). The previous representative, which is dominated
by $x_{new}$ in terms of path cost, is removed from the active set of
nodes $\Vactive$ and is added to the inactive one $\Voff$ (lines
4-5). Then, $x_{new}$ replaces $x_{peer}$ as the representative of its
closest witness $s$ (line 6). If $x_{peer}$ is a leaf node, then it
can also safely be removed from the tree (lines 7-11). The removal of
$x_{peer}$ may cause a cascading effect for its parents, if they were
already in the inactive set $\Voff$ and the only reason they were
maintained in the tree was because they were leading to $x_{peer}$
(lines 7-11). This cascading effect is also illustrated in
Figure \ref{fig:sVV} (C).

\begin{algorithm}[h]
\caption{{\tt Prune\_Dominated\_Nodes\_SST}($x_{new}$, $\Vactive$,
    $\Voff$, $\edges$ )}
\label{alg:prune_dominated}
$s_{new} \leftarrow$ {\tt Nearest}($S$,$x_{new}$)\;
$x_{peer} \leftarrow s_{new}.rep$\;
\If{$x_{peer}!=NULL$}
{
$\Vactive \leftarrow \Vactive \setminus\ \{x_{peer}\} $\;
$\Voff \leftarrow \Voff \cup\ \{x_{peer}\} $\;
}
$s_{new}.rep \leftarrow x_{new}$\;
\While{ $x_{peer}!=NULL$ {\bf\ and} {\tt IsLeaf} $(x_{peer})$ {\bf\ and} $x_{peer} \in \Voff$  } 
{
	$x_{parent} \leftarrow ${\tt Parent}$(x_{peer})$\;
        $\edges \leftarrow \edges \setminus \{\overline{x_{parent}\to x_{peer}}\}$\;
	$\Voff \leftarrow \Voff \setminus\ \{x_{peer}\} $\;
	$x_{peer} \leftarrow x_{parent}$\;
}
\end{algorithm}

\noindent {\bf Implementation Guidelines:} The pseudocode provided
here for \sst\ contains certain inefficiencies to simplify its
description, which should be avoided in an actual implementation. 

In particular, in line 7 of the \stable\ procedure, the trajectory
$\overline{x_{selected}\rightarrow x_{new}}$ is collision checked and
then the algorithm evaluates whether $x_{new}$ is useful to be added
to the tree. Typically, the operations for evaluating whether
$x_{new}$ is useful (nearest neighbor queries, data structure
management and mathematical comparisons) are faster than collision
checking a trajectory. Consequently, it is computationally
advantageous if the check for whether $x_{new}$ is useful, is
performed before the collision checking of
$\overline{x_{selected}\rightarrow x_{new}}$. This is possible if the
underlying moving system is modeled through a set of state update
equations of the form of \edited{Equation} \ref{eq:dynamics}. If, however, the
moving system is a physically simulated one, then it is not possible
to figure out what is the actual final state $x_{new}$ of the
propagated trajectory, without first performing collision
checking. Thus, in the case of a physically simulated system, the
description of the algorithm is closer to the implementation. 

Another issue relates to the first two lines of
\edited{Algorithm} \ref{alg:prune_dominated}, which find the closest witness to the
new node $x_{new}$ and its previous representative. These operations
have actually already taken place in \edited{Algorithm} \ref{alg:is_best} (lines 1
and 6 respectively). An efficient implementation would avoid the
second call to a nearest neighbor query and reuse the information
regarding the closest witness to node $x_{new}$ between the two
algorithms.

\subsection{\stablestar\ (\sst*)}
\sst\ is providing only asymptotic $\delta$-robust
near-optimality.  Asymptotic optimality cannot be achieved by \sst\
directly primarily due to the fixed sized pruning operation employed.
The solution to this is to slowly reduce the radii $\Dnear$ and
$\Ddrain$ employed by the algorithm eventually converging to
iterations that are similar to the \rt\ approach. The key to \sststar,
which is provided in \edited{Algorithm} \ref{alg:SSTstar}, is to make sure that the
rate of reducing the pruning is slow enough to achieve an
anytime behavior, where initial solutions are found for large radii
and then they are improved. As the radii decrease, the algorithm is
able to discover new homotopic classes that correspond to narrow
passages where solution trajectories have reduced clearance.

\begin{algorithm}[h]
\caption{{\tt SST}$^\ast$( $\Xspace$, $\Uspace$, $\xinit$, $T_{prop}$, $N_0$,
$\delta_{BN,0}$, $\delta_{s,0}$, $\xi$ )}
\label{alg:SSTstar}
$j\leftarrow 0$; $N\leftarrow N_0$\;
$\Ddrain \leftarrow \delta_{s,0}$; $\Dnear\leftarrow \delta_{BN,0}$\;
\While{true}
{
	$SST( \Xspace, \Uspace, \xinit, T_{prop}, N, \Dnear, \Ddrain)$\;
	$\Ddrain\leftarrow\xi\cdot\Ddrain$;
	$\Dnear\leftarrow\xi\cdot\Dnear$\; 
        $j\leftarrow j+1$\;
	$N \leftarrow (1+\log j) \cdot \xi^{-(d+l+1)j} \cdot N_0$;
}
\end{algorithm}

\sststar\ provides a schedule for reducing the two radii parameters to \sst,
$\Dnear$ and $\Ddrain$ over time. It receives as input an additional
parameter $\xi$, which is used to decrease the radii $\Dnear$ and
$\Ddrain$ over consecutive calls to \sst\ \edited{(note that $d$ and
$l$ are the dimensionalities of the state and control spaces
respectively)}. This, in effect, makes pruning more difficult to
occur, turns the selection procedure more towards an exploration
objective instead of a best-first strategy and increases the number of
nodes in the data structure. As the number of iterations approaches
infinity, pruning will no longer be performed, the selection process
works in a uniformly at random manner and all collision-free states
will be generated.

Alg. \ref{alg:SSTstar} is a meta-algorithm that repeatedly calls \sst\
as a building block.  In the above call, \sst\ is assumed to be
operating on the same graph data structure $G$ over repeated
calls. It is possible to take advantage of previously generated
versions of the graph data structures with some additional
considerations, e.g., instead of clearing out all states in
$V_{active}$ from previous iterations, one can carefully modify the
pruning procedure to take advantages of the existing $V_{active}$ set
given the updated radii.

\label{sec:sststar}

\subsection{Nearest Neighbor Data Structure}
\label{sec:nearest}
The implementation of \sst\ imposes certain technical requirements
from the underlying nearest neighbor data structure that are not
typical for existing sampling-based motion planners. In particular,
given the pruning operation, it is necessary to have an efficient
implementation of deletion from the nearest neighbor data structure.
In most nearest neighbor structures, a removal of a node will cause
the entire data structure to be frequently rebuilt, severely
increasing run times.

\begin{algorithm}[h]
\caption{{\tt Find\_Closest}($\graph$,$v$)}
\label{alg:findclosest}
$V_{rand} \leftarrow$ {\tt Sample\_Random\_Vertices}$(\graph.\vertx
)$\; $v_{min} \leftarrow \argmin{x \in V_{rand} } ||x - v ||$\;
\Repeat{$v_{min}$ unchanged}
{
	$Nodes \leftarrow {\tt Neighbors}(v_{min})\cup\{v_{min}\} $\;
	$v_{min} \leftarrow \argmin{x \in Nodes} ||x - v||$\;
}
{\bf return} $v_{min}$\;
\end{algorithm}

The goal here is to describe a simple idea for performing approximate
nearest neighbor search using a graph structure $\graph$ that stores
the nodes of the tree and on its edges stores distances between them
according to $d_x( \cdot ,\cdot)$. This approach builds on top of
ideas from random graph theory.  Graphs are conducive to easy removal,
but some overhead is placed in node addition to maintain this data
structure incrementally.

The key operation is finding the closest node in a graph, which is
performed by following a hill climbing approach shown in
\edited{Algorithm} \ref{alg:findclosest}. A random set of nodes is first sampled
from the existing structure, proportional to $\sqrt{\|\vertx\|}$ (line
1). From this set of nodes, the closest node to the query node $v$ is
determined by applying linear search according to $d_x( \cdot ,\cdot)$
(line 2). From the closest node, a hill climbing process is performed
by searching the local neighborhood of the closest node on the graph
to identify whether there are nodes that are closer to the query one
(line 3-6). Once no closer nodes can be found, the locally best node
is returned (line 7).

On top of this operation, it is also possible to define a way for
approximately finding the $k$-closest nodes or the nodes that are
within a certain radius $\delta$. 

\begin{algorithm}[h]
\caption{{\tt Find\_K\_Close}($\graph$,$v$,$k$)}
\label{alg:findkclose}
$v_{min} \leftarrow {\tt FindClosest}( \graph, v )$\;
$K_{near} \leftarrow \{v_{min}\}$\;
\Repeat{$K_{near}$ unchanged}
{
	$Nodes \leftarrow {\tt Neighbors}(K_{near})$\;
	$K_{near} \leftarrow K_{near} \cup Nodes$\;
	$K_{near} \leftarrow$ {\tt Keep\_K\_Closest}($K_{near}$,$v$,$k$)\;
}
{\bf return} $K_{near}$\;
\end{algorithm}
\vspace{-.05in}

The idea in both cases is to start from the closest node by calling
\edited{Algorithm} \ref{alg:findclosest}. Then, each corresponding method searches
the local neighborhoods of the discovered nodes (initially just the
closest node) for either the $k$-closest ones or those nodes that are
within $\delta$ distance. The methods iterate by searching locally
until there is no change in the list.

\begin{algorithm}[h]
\caption{{\tt AddNode}($\graph$,$v$)}
\label{alg:addnode}
$\graph.\vertx \leftarrow \graph.\vertx \cup \{v\}$\;
$K_{near} \leftarrow${\tt FindKClose}( $\graph$, $v$, $k\propto\log(|\graph.\vertx|)$ )\;
\ForEach{$x \in K_{near}$}
{
	$\graph.\edges \leftarrow \graph.\edges \cup \{(v,x)\} \cup \{(x,v)\}$\;
}
{\bf return} $\graph$\;
\end{algorithm}
\vspace{-.05in}

The process of adding nodes to the nearest neighbor data structure is
shown in \edited{Algorithm} \ref{alg:addnode}. It is achieved by first finding the
$k$ closest nodes and then adding edges to them. The number $k$ should
be at least a logarithmic number of nodes as a function of the total
number of nodes to ensure the graph is connected \edited{(similar
to \prmstar)}.

\begin{algorithm}[h*]
\caption{{\tt RemoveNode}($\graph$,$v$)}
\label{alg:removenode}
\ForEach{$\{e \in \graph.\edges\ |\ e.source = v\ \|\ e.target = v\}$}
{
	$\graph.\edges \leftarrow \graph.\edges \setminus {e}$\;
}
$\graph.\vertx \leftarrow \graph.\vertx \setminus {v}$\;
{\bf return} $\graph$\;
\end{algorithm}
\vspace{-.05in}

The reason for using a graph data structure for the nearest neighbor
operations is the ease of removal shown in \edited{Algorithm}
\ref{alg:removenode}. Most implementations of graph data structures
provide such a primitive that is typically quite fast.  This can be
sped up even more if a link to the nearest neighbor graph node is kept
with the tree node allowing for constant time removal.


\section{Analysis}
\label{sec:analysis}
In this section, arguments for the proposed framework
are provided. Sec. \ref{sec:analysismcprop} begins by discussing the
requirements of \mcprop\ and what properties this primitive provides.
Then, in Sec. \ref{sec:analysisrt}, an analysis of the \rt\ approach
is outlined, showing that this algorithm can achieve asymptotic
optimality. To address the poor convergence rate of that approach, the
properties of using the best-first selection strategy are detailed in
Sec. \ref{sec:analysisbestnear}. Finally, in order to introduce the
pruning operation, properties of \sst\ and \sststar\ are studied in
Sec. \ref{sec:analysissst} and \ref{sec:analysissststar}.

\vspace{-.1in}
\subsection{Properties of \mcprop}
\label{sec:analysismcprop}
\edited{The \mcprop\ procedure is a simple primitive for generating
random controls, but provides desirable properties in the context of
achieving asymptotic optimality properties for systems without access
to a steering function. This section aims to illustrate these
desirable properties, given the assumptions from
Section \ref{sec:setup}. Much of the following analysis will use these
results to prove the probabilistic completeness and asymptotic
near-optimality properties of \sst\ and asymptotic optimality
of \sststar. These algorithms are using \mcprop\ for generating random
controls.}

\begin{figure}[h]
\vspace{-0.1in}
\begin{center}
\includegraphics[width=.45\textwidth]{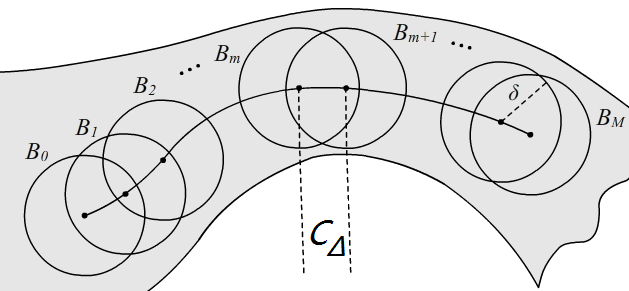}
\hspace{0.1in}
\includegraphics[width=.45\textwidth]{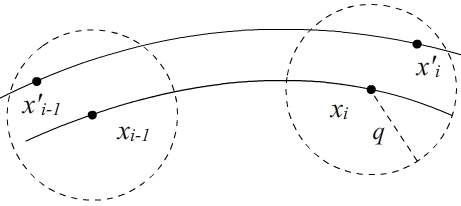}
\end{center}
\caption{(left) An example of a covering ball sequence over a given
  trajectory of radius $\delta$, where each ball is placed so that its
  center has cost $C_{\Delta}$ from the previous ball center. (right)
  The states involved in the arguments regarding the properties of
  random local propagation. }
\vspace{-0.15in}
\label{fig:B_n}
\end{figure}

\edited{The analysis first considers a $\delta$-robust optimal path for
a specific planning query, which is guaranteed to exist for the
specified problem setup. For such a path, consider a covering ball
sequence (an illustration is shown in Fig. \ref{fig:B_n}(left)):}

\begin{definition} (Covering Balls) Given a trajectory $\pi(t)$:
$[0,t_{\pi}]\rightarrow\Xfree$, robust clearance $\delta \in R^+$, and
  a cost value $C_\Delta>0$, the set of covering balls
  $\balls$($\pi(t)$, $\delta$, $C_\Delta$) is defined as a set of
  $M+1$ hyper-balls: \{$\ball_\delta(x_0)$, $\ball_\delta(x_1)$, ...,
  $\ball_\delta(x_M)$\} of radius $\delta$, where $x_i$ are defined
  such that {\tt Cost}($\overline{x_i\to x_{i+1}}$)$=C_\Delta\ $ for $i
  = 0,1,...,M-1$.

\label{def:balls}
\end{definition}

Note that Assumption \ref{ass:cost} about the Lipschitz continuity of
the cost function and Definition \ref{def:balls} imply that for any
given trajectory $\pi$, where $cost(\pi)=C$, and a given duration
$T>0$, it is possible to define a set of covering balls
$\balls$($\pi(t)$, $\delta$, $C_\Delta$) for some $C_\Delta>0$, where
the centers $x_i$ of those balls occur at time $t_i$ of the executed
trajectory.  Since for the given problem setup, the cost function is
non-decreasing along the trajectory and non-degenerate, every segment
of $\pi$ will have a positive cost value.

The covering ball sequence, in conjunction with the following theorem,
provide \edited{a basis for the remaining arguments. In particular,
much of the arguments presented in the rest of
Section \ref{sec:analysis} will consider this covering ball sequence
and the fact that the proposed algorithm can generate a path, which
exists entirely in this covering ball sequence. Once the generation of
such a path asymptotically is proven, its properties in terms of path
quality relatively to the $\delta$-robust optimal path will be
examined.}

\begin{thm}
 For two trajectories $\pi, \pi^\prime$ and any period $T \geq 0$, so
 that $\pi(0)=\pi^\prime(0)=x_0$ and $\Delta u=\sup_t(||u(t) -
 u^\prime(t)||)$:
\vspace{-.05in}
 $$||\pi^\prime(T) - \pi(T) || < K_u\cdot T\cdot e^{K_x\cdot T}\cdot \Delta
 u.$$
\vspace{-.15in}
\label{thm:boundedend}
\end{thm} 

Intuitively, this theorem guarantees that for two trajectories
starting from the same state, the distance between their end states,
in the worst case, is bounded by a function of the difference of their
control vectors. This theorem examines the worst case, and as a
result, the exact bound value is conservative. The proof can be found
in Appendix B. From this theorem, the following corollary is
immediate.

\vspace{-.05in}
\begin{coro}
 For two trajectories $\pi$ and $\pi^\prime$ such that
 $\pi(0)=\pi^\prime(0)=x_0$ and $\Delta u=\sup_t(||u(t), u^\prime(t)||)$:
 $\lim_{\Delta u\rightarrow 0^+} || \pi(T)
 - \pi^\prime(T) || =0$  for any period $T\geq 0$.
 \label{coro:converge} 
\end{coro}
\vspace{-.05in}

\edited{Corollary \ref{coro:converge} is the reason why \mcprop\ can be used
to replace a {\tt Steering} function}. By having the opportunity to
continuously sample control vectors and propagate them forward from an
individual state $x_0$, one can get arbitrarily close to the optimal
control vector, i.e., producing a $\delta$-similar trajectory, where
the $\delta$ value can get arbitrarily small.

The following theorem guarantees that the probability of generating
$\delta$-similar trajectories is nonzero when starting from a
different initial point inside a $\delta$-ball, allowing situations
similar to Figure \ref{fig:B_n} (right) to occur. This property shows
why \mcprop\ is a valid propagation primitive for use in an
asymptotically optimal motion planner.

\begin{thm}
Given a trajectory $\pi$ of duration $t_\pi$, the {\it success
probability} for \mcprop\ to generate a $\delta$-similar trajectory $\pi^\prime$ to
$\pi$ when called from an input state
$\pi^\prime(0) \in \ball_\delta(\pi(0))$ and for a propagation duration
$t_{\pi^\prime} = T_{prop} > t_\pi$ is lower bounded by a positive value $\rho_\delta>0$.
\label{thm:mccomplete}
\vspace{-0.1in}
\end{thm}

\begin{wrapfigure}{l}{.43\textwidth}
\vspace{-.3in}
  \begin{center}
  \includegraphics[width=.42\textwidth]{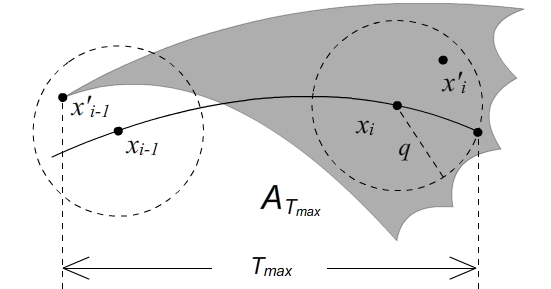}
  \end{center}
\vspace{-.1in}
\caption{An illustration of the local reachability set for
  $x_{i-1}^\prime$. Gray region $A_{T_{max}} = A_{T_{prop}}$ denotes
  the set of states that is reachable from $x_{i-1}^\prime$ within
  duration $[0, t_{\pi^\prime}]$.}
\label{fig:BinA}
\end{wrapfigure}

{\bf Proof:} As in Figure \ref{fig:BinA}, consider that the start of
trajectory $\pi$ is $\pi(0) = x_{i-1}$, while its end is $\pi(t_\pi) =
x_i$. Similarly for $\pi^\prime$: $\pi^\prime(0) = x^\prime_{i-1}$ and
$\pi^\prime(t_{\pi^\prime}) = x_i^\prime$. From Lemma
\ref{lem:closeexist} regarding the existence of dynamic clearance we
have the following: regardless of where $x^\prime_{i-1}$ is located
inside $\ball_\delta(x_{i-1})$, there must exist a $\delta$-similar
trajectory $\pi^\prime$ to $\pi$ starting at $x^\prime_{i-1}$ and
ending at $x_i^\prime$. Therefore, if the reachable set of nodes
$A_{T_{prop}}$ from $x_{i-1}^\prime$ is considered, it must be true
that $\ball_\delta(x_i)\subseteq A_{T_{prop}}$.

In other words, $A_{T_{prop}}$ has the same dimensionality $d$ as the
state space (Assumption \ref{ass:system}), as in in
Fig. \ref{fig:BinA}. The goal is to determine a probability $\rho$
that trajectory $\pi^\prime$ will have an endpoint in
$\ball_\delta(\pi(t_\pi))$.

\begin{figure}[b]
  \begin{center}
  \includegraphics[width=.40\textwidth]{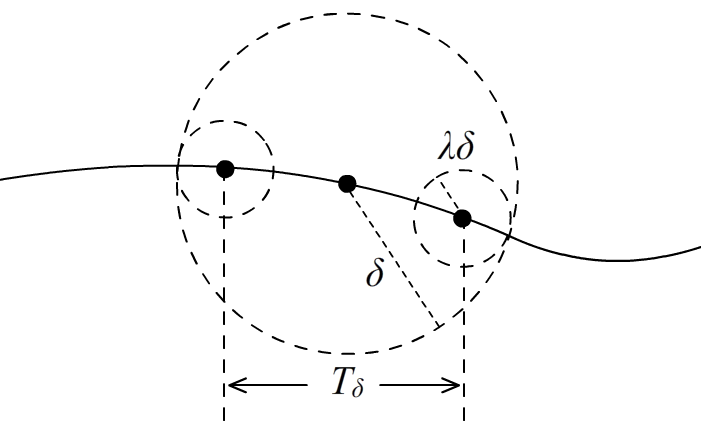}
\hspace{0.2in}
  \includegraphics[width=.50\textwidth]{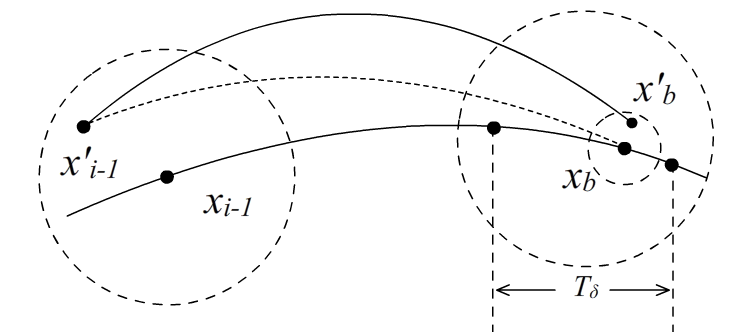}
  \end{center}
\caption{(left) A constructed segment of trajectory $\pi$ of duration
  $T_\delta$. (right) The dotted curve illustrates the existence of a
  trajectory, and the solid curve above it illustrates one possible
  edge that is created by \mcprop.}
\vspace{-.2in}
\label{fig:dSeq}
\end{figure}

Consider Fig. \ref{fig:dSeq} (left). Given a $\lambda\in(0, 1)$,
construct a ball region $b =
\ball_{\lambda\delta}(x_b)$, such that the center state $x_b \in
\pi(t)$ and $b \subset \ball_\delta(x_i)$. Let $\Lambda_\delta$ denote
the union of all such $b$ regions. Clearly, all of $x_b$ form a
segment of trajectory $\pi(t)$. Let $T_\delta$ denote the time
duration of this trajectory segment. For any state $x_b$, there must
exist a $\delta$-similar to $\pi$ trajectory $\pi_b
= \overline{x_{i-1}^\prime\to x_b}$, due to
Lemma \ref{lem:closeexist}.

Recall that \mcprop\ samples a duration for integration, and then,
samples a control vector in $\Upsilon$.  The probability to sample a
duration $t_{\pi_b}$ for $\pi_b$ so that it reaches the region $\Lambda\delta$
is $T_\delta/T_{prop}$.

Since the trajectory segment exists, it corresponds to a control
vector $u_m \in \Upsilon$. \mcprop\ only needs to sample a control
vector $u_m^\prime$, such that it is close to $u_m$ and results in a
$\delta$-similar trajectory. Then Theorem \ref{thm:boundedend}
guarantees that \mcprop\ can generate trajectory $\pi_b^\prime =
\overline{x_{i-1}^\prime\to x_b^\prime}$, which has bounded ``spatial
difference'' from $\overline{x_{i-1}^\prime\to x_b}$. And both of them
have exactly the same duration of $t_{\pi_b}$ (see Fig. \ref{fig:dSeq}
(right) for an illustration).  More formally, given the ``spatial
difference'' $\lambda\delta$, if \mcprop\ samples a control vector
$u_m^\prime$ such that:

\vspace{-.1in}
$$|| u_m^\prime - u_m||\leq\frac{\lambda\delta}{K_u\cdot T_{prop}\cdot
  e^{K_x\cdot T_{prop}}}\ \ \ \ \ \ \Rightarrow \ \ \ \ \ \ || x_b -
x_b^\prime ||<\lambda\delta.$$
\vspace{-.1in}

Therefore, starting from state $x_{i-1}^\prime$, with propagation
parameter $T_{prop}$, \mcprop\ generates a $\delta$-similar trajectory
$\overline{x_{i-1}^\prime\to x_b^\prime}$ to $\overline{x_{i-1}\to
  x_i}$ with probability at
least 

\vspace{-.05in}
$$\rho_{\delta}=\frac{T_\delta}{T_{prop}}\cdot
\frac{\zeta\cdot(\frac{\lambda\delta}{K_u\cdot T_{prop}\cdot
    e^{K_x\cdot T_{prop}}})^w}{\mu(U_m)}>0.$$
\vspace{-.05in}
\telos

This theorem guarantees that the maximum ``spatial difference''
between $\pi(t)$ and $\pi^\prime(t)$, within time $T$, can be bounded
and the bound is proportional to the maximum difference of their
control vectors. This duration bound also implies a cost bound, which
will be leveraged by the following theorems.

\vspace{-.1in}
\subsection{Naive Algorithm: Already Asymptotically Optimal}
\label{sec:analysisrt}
This section considers the impractical sampling-based tree algorithm
\edited{outlined in Algorithm} \ref{alg:RT}, which does not employ a
steering function. Instead, it selects uniformly at random a reachable
state in the existing tree and applies random propagation to extend
it. The following discussion argues that this algorithm eventually
generates trajectories $\delta$-similar to optimal ones. \edited{The
general idea is to prove by induction that a sequence of trajectories
between the covering balls of an optimal trajectory can be
generated. This proof shows probabilistic completeness. Then, from the
properties of \mcprop, the quality of the trajectory generated in this
manner is examined. Finally, if the radius of the covering-ball
sequence tends toward zero, asymptotic optimality is achieved.}

Consider an optimal trajectory $\pi^\ast$ and its covering ball
sequence $\balls$($\pi^\ast(t)$, $\delta$, $C_\Delta$).  Let
$A_k^{(n)}$ denote the event that at the $n^{th}$ iteration of $ALG$,
a $\delta$-similar trajectory $\pi$ to the $k^{th}$ segment of the
optimal sub-trajectory $\overline{x_{k-1}^\ast\to x_k^\ast}$ is
generated, such that $\pi(0) \in \ball_{\delta}(x_{k-1}^\ast)$ and
$\pi(t_{\pi}) \in \ball_{\delta}(x^\ast_k)$.  Then, let $E_k^{(n)}$
denote the event that from iteration $1$ to $n$, an algorithm
generates at least one such trajectory, thereby expressing whether an
event $A_k^{(n)}$ has occurred. The following theorems reason about
the value of $E_k^{(\infty)}$ where $k$ is the number of segments in
$\pi^\ast$ resulting from the choice of $T_{prop}$.

\vspace{-0.1in}
\begin{thm}
\rt\ will eventually generate a $\delta$-similar trajectory to an
optimal one for any robust clearance $\delta > 0$.
\label{thm:drtsimilar}
\end{thm}
\vspace{-0.1in}
\noindent The proof of Theorem \ref{thm:drtsimilar} is in Appendix
C. From this theorem, the following is true.
\vspace{-0.1in}
\begin{coro}
\rt\ is {\it probabilistically complete}.
\label{coro:drtcomplete}
\end{coro}
\vspace{-0.1in}

\vspace{-0.1in}
\begin{thm}
 \rt\ is {\it asymptotically optimal}.
 \label{thm:drtoptimal}
\end{thm}
\vspace{-0.1in}

The proof of \edited{Theorem} \ref{thm:drtoptimal} is in Appendix D
and shows it is possible to achieve asymptotic optimality in a rather
na\"{\i}ve way. This approach is impractical to use however.  Consider
the rate of convergence for the probability $\Prob(E_k^{(n)})$ where
$k$ denotes the $k^{th}$ ball and $n$ is the number of
iterations. Given \edited{Theorem} \ref{thm:drtsimilar},
$\Prob(E_k^{(n)})$ converges to 1. But the following is also true.




\vspace{-0.1in}
\begin{thm}
For the worst case, the $k^{th}$ segments of the trajectory returned
by \rt\ converges logarithmically to the near optimal solution,
i.e., $\lim_{n\to\infty}\frac{|\Prob(E_{k}^{(n+2)})
- \Prob(E_{k}^{(n+1)})|}{|\Prob(E_k^{(n+1)}) - \Prob(E_k^{(n)})|}=1.$
\label{thm:convergenRateRTlog}
\end{thm}

\vspace{-0.1in}

The significance of Theorem \ref{thm:convergenRateRTlog} (proven in
Appendix E) comes from the realization that expecting to generate a
$\delta$-similar trajectory \emph{segment} to an optimal trajectory
$\pi^\ast$ requires an exponential number of iterations with this
approach. This can also be illustrated in the following way. In
the \rt\ approach, as in \rrtconnect, each vertex in $V$ has unbounded
degree asymptotically.

\vspace{-0.1in}
\begin{thm}
 For any state $x_i\in V$, such that $x_i$ is added into $V$ at
 iteration $i$, \rt\ will select $x_i$ to be propagated {\it
 infinitely often} as the execution time goes to infinity.
\vspace{-.1in}
$$\Prob(\limsup_{n\to\infty}\{x_i\text{ is selected}\})
 = 1.$$ \label{thm:drtinfinite} \vspace{-.4in}
\end{thm}

Theorem \ref{thm:drtinfinite} (proven in Appendix F) indicates
that \rt\ will attempt an infinite number of propagations from each
node, and the duration of the propagation does not decrease, unlike
in \rrtconnect\ where the expected length of new branches converge to
0 \citep{Karaman2011Sampling-based-}. The assumption
of \emph{Lipschitz continuity} of the system is enough to guarantee
optimality. Due to this reason, \rt\ is trivially asymptotically
optimal.

Another way to reason about the speed of convergence is the
following. Let $p$ be the probability of an event to happen. The
expected number of independent trials for that event to happen is
$1/p$.  Then, the probability of such an event happening converges to
and is always greater than $1-e^{-1}\approx 63.21\%$, after $1/p$
independent trials, as $p\to 0$ \citep{Grimmett:2001fk}.  Consider
event $A_1$ from the previous discussion (the event of generating the
first $\delta$-similar trajectory segment to an optimum one at any
particular iteration) and recall that the success probability of
the \mcprop\ function is $\rho$. If $\xinit$ is selected for \mcprop,
then the probability of $\Prob(A_1|\{\xinit\text{ is selected}\})
= \rho$. Then the ``expected number'' of times we need to select
$\xinit$ for $A_1$ to happen is $1/{\rho}$. The expected number of
times that $\xinit$ is selected after $n$ iterations is
$\sum_{i=1}^n\frac{1}{i}$. This yields the following expression for
sufficiently large $n$: $
\frac{1}{\rho}=\sum_{i=1}^n\frac{1}{i} \approx ln(n) + c_\gamma
$ where $c_\gamma$ is the Euler-Mascheroni constant, which yields: $
n\approx e^{(\rho^{-1} - c_\gamma)}.  $ Therefore, in order even for
event $E_1$ (event of $A_1$ happening at least once ) to happen with
approximately $1-e^{-1}$ probability for small $\rho$ values, the
expected number of iterations is exponential to the reciprocal of the
success probability $\rho$ of the \mcprop\ function. This
implies \emph{intractability}. For efficiency purposes it is necessary
to have methods where $n$ does not depend exponentially to
$\frac{1}{\rho}$.

\vspace{-.1in}
\subsection{Using \bestnear: Improving Convergence Rate}
\label{sec:analysisbestnear}
A computationally efficient alternative to \rt\ for finding a path, if
one exists, is referred to here as \rrtbestnear, which works like \rt\
but switches line 3 in \edited{Algorithm} \ref{alg:RT} with the procedure in
\edited{Algorithm} \ref{alg:bestNearest}. An important observation from the
complexity discussion for \rt\ is that the exponential term arises
from the use of uniform random sampling for selection among the
existing nodes. By not using any path cost information when performing
selection, the likelihood of generating good trajectories becomes very
low, even if it is still non-zero.

\begin{figure}[ht]
\vspace{-.1in}
  \begin{center}
	\includegraphics[width=.48\textwidth]{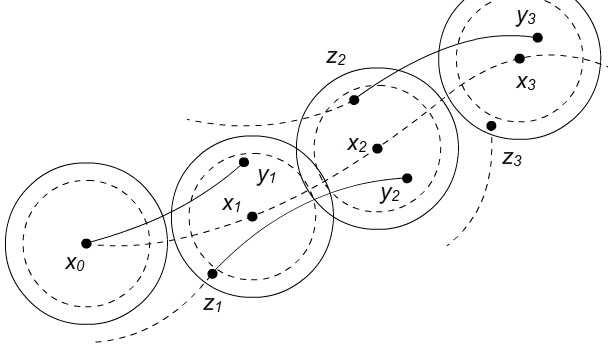}
  \hspace{.5in}
	\includegraphics[width=.28\textwidth]{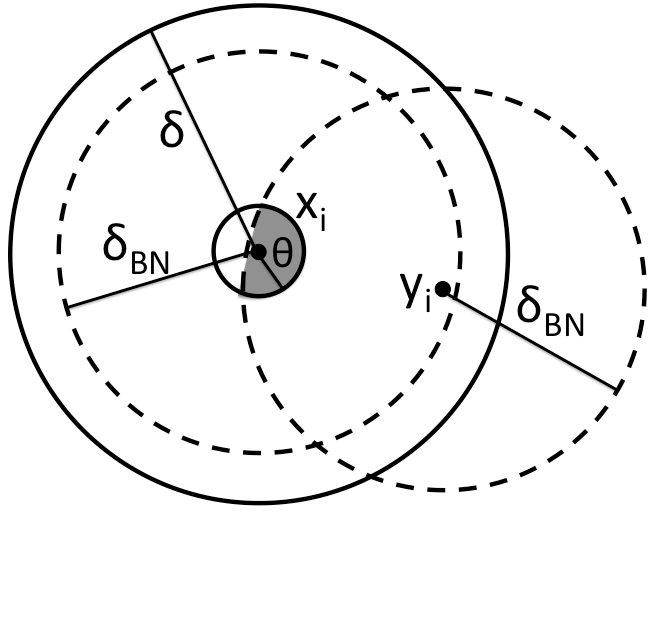}
  \end{center}
\vspace{-.2in}
  \caption{(left) Illustration of different trajectories generated
    by \mcprop\ inside the covering balls
    $\balls$($\pi^\ast$,$\delta$,$C_\Delta$). Many trajectories may
    enter these balls, but may not be $\delta$-similar to the optimal
    one. (right) Sampling $x_{rand}$ in the gray region guarantees
    that a node $z_i \in \ball_\delta(x_i)$ is selected for
    propagation so that either $z_i = y_i$ or $cost(z_i) <
    cost(y_i)$. \vspace{-.1in}}
\label{fig:sparserrt_reasoning}
\end{figure}

The analysis of \rrtbestnear\ involves similar event \edited{constructions as in the previous section}:
$A_k^{(n)}$ and $E_k^{(n)}$ are defined as in the previous section,
except the endpoint of the trajectory segment generated must be in
$\ball_{\Dnear}(x_k^\ast)$. The propagation from \mcprop\ still has
positive probability of occurring, but is different from
$\rho_\delta$. The changed probability for \mcprop\ to generate such a
trajectory is defined as $\rho_{\delta\to\Dnear}$ The probabilities of
these events will also change due to the new selection process and
more constrained propagation requirements. It must be shown that nodes
that have good quality should have a positive probability of
selection.  Consider the selection mechanism \bestnear\ in the context
of Figure \ref{fig:sparserrt_reasoning}.

\vspace{-0.05in}
\begin{lemmma} 
Assuming uniform sampling in the {\tt Sample} function of \bestnear,
if $\exists\ x$ s.t. $x\in \ball_{\delta_{BN}}(x_i^\ast)$ at iteration
$n$, then the probability that \bestnear\ selects for propagation a
node $x^\prime\in \ball_{\delta}(x_i^\ast)$ can be lower bounded by a
positive constant $\gamma$ for every $n^\prime > n$.  
 \label{lem:gammaBN}
\end{lemmma}
\vspace{-0.05in}

\noindent {\bf Proof:} Consider the case that a random sample $x_{rand}$ is
placed at the intersection of a small ball of radius $\theta = \delta
- \Dnear$ (guaranteed positive from Proposition \ref{prop:deltas}),
and of a $\Dnear$-radius ball centered at a state
$y_i \in \ball_{\Dnear}(x_i)$ that was generated during an iteration
of an algorithm. State $y_i$ exists with probability
$\Prob(E_k^{(n)})$. In other words, if $x_{rand} \in \ball_\theta(x_i)
\cap \ball_{\Dnear}(y_i)$, then $y_i$ will always be considered by \bestnear\ because $y_i$ will always be within $\Dnear$ distance of a random sample there. 
The small circle is defined so that the $\Dnear$ ball of $x_{rand}$
can only reach states in $\ball_\delta(x_i)$. It is also required that
$x_{rand}$ is in the $\Dnear$-radius ball centered at $y_i$, so that
at least one node in $\ball_\delta(x_i)$ is guaranteed to be
returned. Thus, the probability the algorithm select for propagation a
node $x^\prime\in \ball_{\delta}(x_i^\ast)$ can be lower bounded by
the following expression:
\vspace{-0.05in}
\begin{equation*}
\gamma = \frac{ \mu( \ball_\theta(x_i) \cap
  \ball_{\Dnear}(x^\prime)  )} {\mu(\Xfree)} > 0
\vspace{-0.1in}
\end{equation*}\telos

\noindent
With \edited{Theorem} \ref{thm:mccomplete}
and \edited{Lemma} \ref{lem:gammaBN}, both the selection and
propagation probabilities are positive and it is possible to argue
\edited{probabilistic completeness of \rrtbestnear. The full proof is provided in Appendix G.}:

\edited{
\begin{thm}
\rrtbestnear\ will eventually generate a $\delta$-similar trajectory to any optimal trajectory.
\label{thm:deltacompleteBN}
\end{thm}
}

The proof of asymptotic $\delta$-robust near-optimality follows
directly from Theorem. \ref{thm:deltacompleteBN}, the \emph{Lipschitz
continuity}, \emph{additivity}, and \emph{monotonicity} of the cost
function (Assumption \ref{ass:cost}). \edited{Theorem \ref{thm:deltacompleteBN}} is
already examining the generation of a $\delta$-similar trajectory to
$\pi^\ast$, but the bound on the cost needs to be calculated \edited{(as is constructed in Appendix H)}.

\edited{
\begin{thm}
\rrtbestnear\ is asymptotically $\delta$-robustly near-optimal.
\label{thm:bestnearoptimal}
\end{thm}}

\noindent
\edited{The addition of \bestnear\ was introduced to address the convergence rate issues of \rt. Theorem \ref{thm:convergenRateBNlinear} quantifies this convergence rate.}

\vspace{-.1in}
\begin{thm}
For the worst case, the $k^{th}$ segment of the trajectory returned by \rrtbestnear\ converges linearly to the near optimal solution, i.e,
$\lim_{n\to\infty}\frac{|\Prob(E_{k}^{(n+1)}) - 1|}{|\Prob(E_k^{(n)}) - 1|}=(1-\gamma\rho_{\delta\to\delta_{BN}})\in(0, 1).$
\label{thm:convergenRateBNlinear}
\end{thm} 
\vspace{-.1in}

\noindent {\bf Proof:} Applying the boundary condition of
\edited{Equation} \ref{eq:EpropagateBN}, consider the ratio of the probabilities
between iteration $n+1$ and $n$.

\vspace{-.1in}
$$\frac{|\Prob(E_{k}^{(n+1)}) - 1|}{|\Prob(E_k^{(n)}) - 1|}
=\frac{\prod_{j=1}^{n+1}(1-\Prob(E_{k-1}^{(j)})\cdot\gamma\rho_{\delta\to\delta_{BN}})}{\prod_{j=1}^{n}(1-\Prob(E_{k-1}^{(j)})\cdot\gamma\rho_{\delta\to\delta_{BN}})}
=1-\Prob(E_{k-1}^{(n+1)})\cdot\gamma\rho_{\delta\to\delta_{BN}}$$
\vspace{-.1in}

Taking $\lim_{n\to\infty}$, and given \edited{Theorem} \ref{thm:deltacompleteBN}
such that $\lim_{n\to\infty}\Prob(E_{k-1}^{(n+1)})=1$, the following holds:

\vspace{-.1in}
$$\lim_{n\to\infty}\frac{|\Prob(E_{k}^{(n+1)}) - 1|}{|\Prob(E_k^{(n)})
  - 1|}
=\lim_{n\to\infty}(1-\Prob(E_{k-1}^{(n+1)})\cdot\gamma\rho_{\delta\to\delta_{BN}})
=1-\gamma\rho_{\delta\to\delta_{BN}} \in(0, 1). \ \ \ \ \ \ \ \telos$$

\edited{Theorem} \ref{thm:convergenRateBNlinear} states that \rrtbestnear\
converges linearly to near optimal solutions. Recall that the \rt\
approach converges logarithmically (sub-linearly). This difference
indicates that \rrtbestnear\ converges significantly faster than \rt.
Now consider the expected number of iterations, i.e. the iterations
needed to return a near-optimal trajectory with a certain
probability. Specifically, the convergence rate depends on the
difficulty level of the kinodynamic planning problem, which is
measured by the probability $\rho_{\delta\to\delta_{BN}}$ of
successfully generating a $\delta$-similar trajectory segment
connecting two covering balls.

Recall that the expected number of iterations for $E_1$ to succeed
for \rt\ was $n\approx e^{c_\gamma}\cdot e^{(\rho^{-1})}$. In
the case of \rrtbestnear\ for event $E_1$, this expected number of
iterations is
$\frac{1}{1-e^{-1}}\cdot\frac{1}{\gamma\rho_{\delta\to\delta_{BN}}}$. This
is a significant improvement already for event $E_1$ (though providing
a weaker near-optimality guarantee).  For the cases of $E_k$, ($k>1$),
the expected number of iterations for \rrtbestnear\ linearly depends
on the length of the optimal trajectory. While for \rt, it is already
intractable even for the first ball.

On the other hand, in terms of ``per iteration'' computation
time, \rrtbestnear\ is worse than \rrt. The \bestnear\ procedure
requires a $\Dnear$-radius query operation which is computationally
more expensive than the nearest neighbor query in \rrt.
Therefore, \rrtbestnear\ shall be increasingly slower than \rrt.
Nevertheless, the following section shows that maintaining a sparse
data structure can help in this direction.

\vspace{-.1in}
\subsection{\stable\ Analysis}
\label{sec:analysissst}
This section argues that the introduction of the \emph{pruning
process} in \sst\ does not compromise asymptotic $\delta$-robust
optimality and improves the computational efficiency.  Consider the
selection mechanism used in \sst.

\vspace{-.1in}
\begin{lemmma}
Let $\delta_c = \delta-\Dnear-2\delta_s$. If a state $x_{new}\in
V_{active}$ is generated at iteration $n$ so that $x \in
\ball_{\delta_c}(x_i^\ast)$, then for every iteration $n^\prime \geq
n$, there is a state $x^\prime\in V_{active}$ so that $x^\prime\in
\ball_{(\delta-\Dnear)}(x_i^\ast)$ and {\tt
  cost}($x^\prime$) $\leq$ {\tt cost}($x$).  \label{lem:stabilization}
\end{lemmma}
\vspace{-.1in}

\begin{wrapfigure}{l}{3.5in}
\vspace{-.15in}
\includegraphics[width = 3.4in]{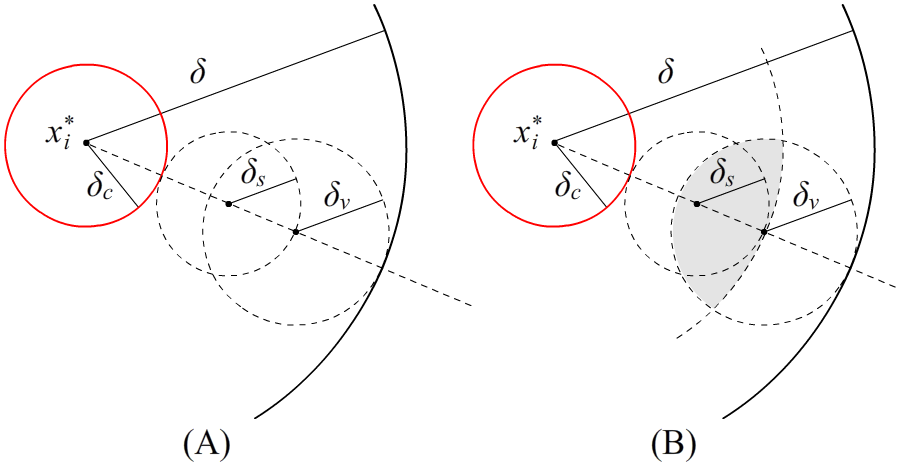}
\caption{The selection mechanism of \sst.  \vspace{-.2in}}
\label{fig:select_cover}
\end{wrapfigure}

\noindent {\bf Proof:} Given $x$, a node generated by \sst, then it is
guaranteed that a witness point $s$ is located near $x$. As in Fig.
\ref{fig:select_cover} (A), the witness point $s$ can be located, in the
worst case, at distance $\delta_s$ away from the boundary of
$\ball_{\delta_c}(x_i^\ast)$ if $x\in \ball_{\delta_c}(x_i^\ast)$.

Note that $x$ can be removed from $\Vactive$ by \sst\ in later
iterations.  In fact, $x$ almost surely will be removed if $x\not
=\xinit$. It is possible that when $x$ is removed, there could be no
state in the ball $\ball_{\delta_c}(x_i^\ast)$.  Nevertheless, the
witness sample $s$ will not be deleted. A node $x^\prime$ representing
$s$ will always exist in $\Vactive$ and $x^\prime$ will not leave the
ball $\ball_{\delta_s}(s)$.  It is guaranteed by \sst\ that the cost
of the $x^\prime$ will never increase, i.e., {\tt
  cost}($x^\prime$)$\leq${\tt cost}($x$). In addition, $x^\prime$ has
to exist inside
$\ball_{\delta-\Dnear}(x^\ast_i)=\ball_{\delta_c+2\delta_s}(x^\ast_i)$. \telos

Lemma \ref{lem:stabilization} is where \sst\ gains its {\tt Stable}
moniker. By examining what happens when a trajectory is generated that
ends in $\ball_{\delta_c}(x_i^\ast)$, a guarantee can be made that
there will always be a state in the $\ball_\delta(x_i^\ast)$, thus
becoming a stable point. The relationship between $\Dnear$,$ \Ddrain$,
and $\delta$ must satisfy the requirements of
Proposition \ref{prop:deltas} in order to provide this property.
After proving the continued existence of
$x^\prime \in \ball_{\delta-\Dnear}(x_i^\ast)$, \edited{Lemma} \ref{lem:gamma}
provides a lower bound for the probability of selecting $x^\prime$.

\vspace{-.1in}
\begin{lemmma}
Assuming uniform sampling in the {\tt Sample} function of \bestnear,
if $\exists\ x\in \Vactive$ so that $x\in \ball_{\delta_c}(x_i^\ast)$
at iteration $n$, then the probability that \bestnear\ selects for
propagation a node $x^\prime\in \ball_{\delta}(x_i^\ast)$ can be lower
bounded by a positive constant $\gamma_{sst}$ for every $n^\prime >
n$.  \label{lem:gamma}
\end{lemmma}
\vspace{-.1in}

\noindent {\bf Proof:} See Fig. \ref{fig:select_cover}(A):
\bestnear\ performs uniform random sampling in $\Xspace$ to generate
$x_{rand}$, and then examines the ball $\ball_{\Dnear}(x_{rand})$ to
find the best path node. In order for a node in
$\ball_{\delta}(x_i^\ast)$ to be returned, the sample needs to be in
$\ball_{\delta-\Dnear}(x_i^\ast)$.  If the sample is outside this
ball, then a node not in $\ball_{\delta}(x_i^\ast)$ can be considered,
and therefore may be selected.

Next, consider the size of the intersection of
$\ball_{\delta-\Dnear}(x_i^\ast)$ and a ball of radius $\Dnear$ that
is entirely enclosed in $\ball_{\delta}(x_i^\ast)$. Let $x_v$ denote
the center of this ball. This intersection, highlighted in
Fig. \ref{fig:select_cover}(B), represents the area that a sample can be
generated so as to return a state from ball
$\ball_{\delta-\Dnear}(x_i^\ast)$.  In the worst case, the center of this
ball $\ball_{\Dnear}(x_v)$ could be on the border of
$\ball_{\delta-\Dnear}(x_i^\ast)$, as seen in
Fig. \ref{fig:select_cover} B.  Then, the probability of sampling a
state in this region can be computed as: $\gamma_{sst} = \inf
\mathbb{P}\Big(\big\{x^\prime\text{ returned by {\tt BestNear}}:
x^\prime\in \ball_{\delta}(x_i^\ast)\big\}\Big)
=\frac{\mu(\ball_{\delta-\Dnear}(x_i^\ast)\cap
  \ball_{\Dnear}(x_v))}{\mu(\Xfree)}$.  This is the smallest region
that will guarantee selection of a node in $\ball_{\delta}(x_i)$.

\telos

Lemma \ref{lem:gamma} shows that the probability to select a near
optimal state within the covering ball sequence with a non-decreasing
cost can be lower bounded. It is almost identical to the selection
mechanism of \rrtbestnear. Similarly to the analysis of \rrtbestnear,
the probability that \mcprop\ is now again different. The trajectories
considered here must enter balls of radius $\delta_c$, so the changed
probability for \mcprop\ to generate such a trajectory is
$\rho_{\delta\to\delta_c}$. With $\gamma_{sst}$ and
$\rho_{\delta\to\delta_c}$ defined, the completeness of \sst\ can be
argued.

\begin{thm}
{\tt STABLE} \sparserrt\ is probabilistically 
$\delta$-robustly complete. e.g.,
\vspace{-.05in}
\[\liminf_{n\to\infty}\Prob(\ \exists\ \pi \in \Pi^{SST}_n: \pi \text{
solution to } (\Xfree, \xinit, \Xgoal, \delta) ) = 1.\]
\label{thm:deltacomplete}
\end{thm}

\vspace{-.25in}

\begin{thm}
{\tt STABLE} \sparserrt\ is asymptotically $\delta$-robustly
near-optimal. e.g.
\vspace{-.05in}
\[\Prob\Big(\Big\{\limsup_{n\to\infty}Y_n^{SST}\leq(1+\frac{K_x\delta}{C_\Delta})\cdot C^\ast\Big\}\Big)=1.\]
\label{thm:nearoptimal}
\end{thm}
\vspace{-.15in}

The proofs for Theorem \ref{thm:deltacomplete} and
Theorem \ref{thm:nearoptimal} are almost identical to the proofs of
Theorem \ref{thm:deltacompleteBN} and
Theorem \ref{thm:bestnearoptimal} respectively. The only differences
are the different probabilities $\gamma_{sst}$ and
$\rho_{\delta\to\delta_c}$. By changing the radii in the proofs of
Theorem \ref{thm:deltacomplete} and Theorem \ref{thm:nearoptimal} to
their correct values in \sst, the proofs hold.

\vspace{-.05in}
\begin{thm}
In the worst case, the $k^{th}$ segment of the trajectory returned
by \sst\ converges linearly to the near optimal one, i.e.,
\vspace{-0.05in}
\[\lim_{n\to\infty}\frac{|\Prob(E_{k}^{(n+1)}) - 1|}{|\Prob(E_k^{(n)}) - 1|}=(1-\gamma\rho_{\delta\to\delta_c})\in(0, 1).\]
\vspace{-0.05in}
\label{thm:convergenRateSSTlinear}
\end{thm} 
\vspace{-.05in}

The convergence rate and expected iterations for \sst\ are again almost
identical to that of \rrtbestnear, since both the selection
mechanism and the propagation probability of \sst\ can be bounded by
constants.

The benefit of \sst\ is that the per iteration complexity ends up
being smaller than \rrtbestnear. The most expensive operation for the
family of algorithms discussed in this paper asymptotically is the
near neighbor query. \sst\ delivers noticeable computational
improvement over \rrtbestnear\ due to the reduced size of the tree
data structure. The rest of this section examines the influence of the
sparse data structure, which is brought by the pruning process in \sst.

Among a set of size $n$ points, the average time complexity for a
nearest neighbor query is $\mathcal{O}(\log n)$. The average time
complexity of the range query for near neighbors is $\mathcal{O}(n)$,
since the result is a fixed proportional subset of the whole set.
Using this information, it is possible to estimate the overall
asymptotic time complexities for \rrtbestnear\ and \sst\ to return
near-optimal solutions with probability at least $1-e^{-1}\approx
63.21\%$.

\vspace{-0.1in}
\begin{lemmma}
For a $k$ segment optimal trajectory with $\delta$ clearance, the expected running time for \rrtbestnear\ to return a near-optimal solution with $1-e^{-1}$ probability can be evaluated as:
$\mathcal{O}\Big( (\frac{k}{\gamma\rho_{\delta\to\delta_c}})^2 \Big)$
  \label{lem:complexityBN}
\end{lemmma}
\vspace{-0.1in}

\noindent {\bf Proof:}
Let $N_p$ denote $\frac{k}{(1-e^{-1})\gamma\rho}$. The total time
computation after $N_p$ iterations can be evaluated as, $
\mathcal{O}\Big( \sum_{i=1}^{N_p} c\cdot i\Big) = \mathcal{O}\Big( c\cdot\frac{N_p(N_p+1)}{2} \Big)
	= \mathcal{O}\Big( (\frac{k}{\gamma\rho_{\delta\to\delta_c}})^2 \Big).
$\telos

For \rrtextend\ the expected number iterations needed to generate a
trajectory can be bounded by
$\frac{k}{\rho\gamma_{rrt}}$ \citep{LaValle2001}.  For the $k^{th}$
segment of a trajectory with $\delta$ clearance, the expected running
time for \rrtextend\ to return a solution with $1-e^{-1}$ probability
can be evaluated as:
$\mathcal{O}\Big( \frac{k}{\rho\gamma_{rrt}}\cdot \log(\frac{k}{\rho\gamma_{rrt}}) \Big)$. 

Now consider \sst. Since each $s\in S$ has claimed a $\delta_s$ radius
hyper-ball in the state space, then the following is true:

\vspace{-0.1in}
\begin{lemmma}
For any two distinct witnesses of \sst: $s_1,s_2\in S$, where
$s_1\not= s_2$, the distance between them is at least $\delta_s$,
e.g., $\forall s_1,s_2\in S: ||s_1 -
s_2||>\delta_s$. \label{lem:hardcore}
\end{lemmma}
\vspace{-0.1in}

Lemma \ref{lem:hardcore} implies that the size of the set $S$ can be
bounded, if the free space $\Xfree$ is bounded.

\vspace{-0.1in}
\begin{coro}
If $\Xfree$ is bounded, the number of points of the set $S$ and nodes in $\Vactive$ is always finite, i.e.,
$\exists M\in\mathcal{O}(\delta^{-d}) : |S| = |V_{active}| \leq M$.
  \label{coro:finiteS}
\end{coro}
\vspace{-0.1in}

Corollary \ref{coro:finiteS} indicates that the total number of points
in set $S$ can be bounded. Then, the complexity of any near neighbors
query can be bounded. Now the improved time complexity of \sst\
relative to \rrt\ can be formulated.

\vspace{-0.1in}
\begin{lemmma}
For a $k$ segment optimal trajectory with $\delta$ clearance, the expected running time for \sst\ to return a near-optimal solution with $1-e^{-1}$ probability can be evaluated as,
$\mathcal{O}\Big( \delta^{-d} \cdot \frac{k}{\gamma\rho_{\delta\to\delta_c}} \Big)$.
  \label{lem:complexitySST}
\end{lemmma}
\vspace{-0.1in}

\noindent {\bf Proof:}
Let $N_p$ denote
$\frac{k}{(1-e^{-1})\gamma\rho_{\delta\to\delta_c}}$. Due to
\edited{Corollary} \ref{coro:finiteS}, the total computation time after $N_p$
iterations is: $
\mathcal{O}\Big( \sum_{i=1}^{N_p} c\cdot\delta^{-d} + N_p\Big)
= \mathcal{O}\Big( \delta^{-d} \cdot \frac{k}{\gamma\rho_{\delta\to\delta_c}} \Big).$
Note that the second term $N_p$ describes the worst case of deletion
of nodes in $V_{inactive}$ in \edited{Algorithm} \ref{alg:SST}. For $N_p$
iterations, in the worst case, the algorithm can delete at most $N_p$
nodes.
\telos

\vspace{-.1in}
\subsection{\sst* Analysis}
\label{sec:analysissststar}

In \sst, for given $\delta$, $\Ddrain$, and $\Dnear$ values,
$\gamma_{sst}$ and $\rho_{\delta\to\delta_{c}}$ are two constants
describing the probability of selecting a near-optimal state for
propagation and of successfully propagating to the next ball
region. Note that if $\Dnear$ and $\Ddrain$ are reduced over time, the
related $\delta$ value can be smaller. This is the intuition behind
why \sststar\ provides asymptotic optimality. If after a sprint of
iterations where $\Dnear$ and $\Ddrain$ are kept static, they are
reduced slightly, this should allow for the generation of trajectories
with smaller clearance, i.e., closer to the true optimum.

\begin{lemmma}
For a $\ball_i$ of radius $\delta$ and a ball $\ball_i^\prime$ with
radius $\delta^\prime$, such that $\delta^\prime / \delta=\alpha$, where
$\alpha\in (0,1)$, there is
$\frac{\hat{\rho}_{\delta^\prime}}{\hat{\rho}_{\delta}}=\alpha^{w+1}$
\label{lem:rhoRatio}
\end{lemmma}

Lemma \ref{lem:rhoRatio} says that when the probability $\rho$ decreases over time, it is reduced by a factor $\alpha$ set to the power of the size of the piecewise constant control vector plus one. The proof of this relationship is in Appendix \edited{I}.

\begin{lemmma}
Given $\delta>0$, and $\delta_{BN}>0$, for a scale $\alpha\in (0,1)$, let $\delta^\prime=\alpha\delta$ and $\delta_{BN}^\prime=\alpha\delta_{BN}$, there is
$\frac{\gamma^\prime}{\gamma}=\alpha^d$
\label{lem:gammaRatio}
\end{lemmma}

Lemma \ref{lem:gammaRatio} says that a similar relationship exists for values of $\gamma$. This probability is defined purely geometrically in the state space, so its proof is trivial. Now that these relationships have been established, properties of \sststar\ can be shown.

\vspace{-0.1in}
\begin{thm}
$SST^\ast$ is {\it probabilistically complete}. i.e.,
  $\liminf_{j\to\infty}\Prob(\{ \exists\ x_{goal} \in
  (V_n^{SST^\ast}\cap\Xgoal) \})=1$
\label{thm:sststarcomplete}
\end{thm}
\vspace{-0.1in}

\noindent {\bf Proof:} Let $E_{k,j}^{(i)}$ ($k\geq 1$) denote the event $E_k$
(as seen from earlier proofs) at sprint $j$, after $i$ iterations within the
sub-function \sst. Then:

\vspace{-.25in}
\begin{align}
\Prob(E_{k,j})&=1-\prod_{i=1}^{K(j)}(1-\Prob(E_{k-1,j}^{(i)})\gamma^{(j)}\rho^{(j)}).
\label{eq:sststarEprop}
\end{align}
\vspace{-.25in}

where $\gamma^{(j)}$ and $\rho^{(j)}$ are the values that have been used to bound selection and trajectory generation probability, but for the $\Dnear$ and $\Ddrain$ values during the $j^{th}$ sprint.
Let $c$ be a constant $1\leq c\leq K(j)$ and $p_c$ be the value of
$\Prob(E_{k-1,j}^{(c+1)})$ for a given $j$. Note that within the same
sprint $j$, Eq. \ref{eq:sststarEprop} is equivalent to
Eq. \ref{eq:EpropagateBN}. Let $P_c$ be a constant, such that
$P_c=\prod_{i=1}^{c}(1-\Prob(E_{k-1,j}^{(i)})\gamma^{(j)}\rho^{(j)})$. Then,
Eq. \ref{eq:sststarEprop} becomes:

\vspace{-.25in}
\begin{align}
\Prob(E_{k,j})	=1-P_c \prod_{i=c+1}^{K(j)}(1-\Prob(E_{k-1,j}^{(i)})\gamma^{(j)}\rho^{(j)}).
\label{eq:Edivide}
\end{align}
\vspace{-.25in}

And clearly, any $\Prob(E_{k,j}^{(i)})$ ($k\geq 0$) is strictly positive and
non-decreasing, as $i$ increases, meaning $p_c$ can be used as a lower bound.
Then Eq. \ref{eq:Edivide} becomes:

\vspace{-.25in}
\begin{align}
\Prob(E_{k,j})	&=1-P_c
\prod_{i=c+1}^{K(j)}(1-\Prob(E_{k-1,j}^{(i)})\gamma^{(j)}\rho^{(j)}) \geq 1-P_c \prod_{i=c+1}^{K(j)}(1-p_c\cdot\gamma^{(j)}\rho^{(j)})\notag\\
				&=1-P_c
(1-p_c\cdot\gamma^{(j)}\rho^{(j)})^{K(j)-c} = 1-P_c \Big[(1-p_c\cdot\gamma^{(j)}\rho^{(j)})^{\frac{1}{\gamma^{(j)}\rho^{(j)}}}\Big]^{\cdot\gamma^{(j)}\rho^{(j)}\cdot (K(j)-c)}.
\label{eq:Eshrink}
\end{align}
\vspace{-.25in}

Since the inequality $(1-\frac{\alpha}{x})^x<e^{-\alpha}$ for all
$x>1$ and $\alpha>0$. Then Eq. \ref{eq:Eshrink} becomes, borrowing from Algorithm \ref{alg:SSTstar} the expression for the number of iterations $K(j)$, expression (\ref{eq:Eshrink}) becomes:

\vspace{-.25in}
\begin{align}
\Prob(E_{k,j})&\geq 1-P_c \Big[(1-p_c\cdot\gamma^{(j)}\rho^{(j)})^{\frac{1}{\gamma^{(j)}\rho^{(j)}}}\Big]^{\gamma^{(j)}\rho^{(j)}\cdot (K(j)-c)}\notag\\
					&>1-P_c (e^{-p_c})^{\gamma^{(j)}\rho^{(j)}\cdot (K(j)-c)}\notag\\
					&=1-P_c (e^{-p_c})^{\gamma^{(j)}\rho^{(j)}\cdot\xi^{-(d+w+1)j}\cdot k_0\cdot(1+\log j)-\gamma^{(j)}\rho^{(j)}c}.
					\label{eq:EshrinkBeta}
\end{align}
\vspace{-.25in}

Let $\beta=k_0\cdot\gamma^{(0)}\rho^{(0)}$, Eq. \ref{eq:EshrinkBeta}
becomes:

\vspace{-.25in}
\begin{align}
\Prob(E_{k,j})	&> 1-P_c (e^{-p_c})^{\gamma^{(j)}\rho^{(j)}\cdot\xi^{-(d+w+1)j}\cdot k_0\cdot(1+\log j)-\gamma^{(j)}\rho^{(j)}c}\notag\\
						&=1-P_c (e^{-p_c})^{\gamma^{(0)}\rho^{(0)}\cdot k_0\cdot(1+\log j)-\gamma^{(j)}\rho^{(j)}c}\notag\\
						&=1-P_c (e^{-p_c})^{\beta\cdot(1+\log j)-\gamma^{(j)}\rho^{(j)}c}
						\label{eq:Esimplified}
\end{align}
\vspace{-.25in}

because of Lemma \ref{lem:rhoRatio}-\ref{lem:gammaRatio}. As $j$ increases to infinity, the following holds:

\vspace{-.25in}
\begin{align*}
\lim_{j\to\infty}\Prob(E_{k,j}) > \lim_{j\to\infty} 1-P_c (e^{-p_c})^{\beta\cdot(1+\log j)-\gamma^{(j)}\rho^{(j)}c}
								= \lim_{j\to\infty} 1-\frac{P_c\cdot(e^{-p_c})^{\beta\cdot(1+\log j)}}{(e^{-p_c})^{\gamma^{(j)}\rho^{(j)}c}}
								= 1-\frac{0}{1}=1.
\end{align*}
\vspace{-.25in}

Since the limit exists, therefore it is true that
$\liminf_{j\to\infty}\Prob(E_{k,j})=1$. \telos

\noindent Next, the argument regarding asymptotic optimality.

\vspace{-.1in}
\begin{thm}
{\tt SST$^\ast$} is {\it asymptotically optimal}. i.e., $\Prob(\{\limsup_{j\to\infty}Y_j^{SST^\ast}=c^\ast\})=1$.
\label{thm:sststaroptimality}
\end{thm}
\vspace{-.1in}

\noindent {\bf Proof:} Since event $E_{k,j}$ implies
event \{$Y_n^{SST^\ast}\leq(1+c_\alpha\delta)\cdot c^\ast$\},
therefore at the end of the $j^{th}$ sprint, from Eq. \ref{eq:Esimplified}:

\vspace{-.25in}
\begin{align*}
\Prob(\{Y_j^{SST}\leq(1+c_\alpha\delta^{(j)})\cdot c^\ast\})=\Prob(E_{k,j})
			>1-\frac{P_c\cdot(e^{-p_c})^{\beta\cdot(1+\log j)}}{e^{\gamma^{(j)}\rho^{(j)}c}}
\end{align*}
\vspace{-.25in}

\noindent As $j\to\infty$, clearly $\lim_{j\to\infty}\delta^{(j)}=0$,
$\lim_{j\to\infty}\gamma^{(j)}=0$, and
$\lim_{j\to\infty}\rho^{(j)}=0$. Then, it is true that:

\vspace{-.25in}
\begin{align*}
\Prob(\{\lim_{j\to\infty}Y_j^{SST}\leq\lim_{j\to\infty}(1+c_\alpha\delta^{(j)})\cdot c^\ast\})
=\Prob(\{\lim_{j\to\infty}Y_j^{SST}\leq c^\ast\})
>\lim_{j\to\infty}\Big[1-\frac{P_c\cdot(e^{-p_c})^{\beta\cdot(1+\log j)}}{e^{\gamma^{(j)}\rho^{(j)}c}}\Big]=1-\frac{0}{1}=1.
\end{align*}
\vspace{-.25in}

\telos


\edited{Algorithm} \ref{alg:SSTstar} describes a process that gradually relaxes the
``sparsification'', which increasingly allows adding active states. At
a high level perspective, \rrtstar\ employs the same idea
implicitly. Recall that \rrtstar\ also allows adding states as the
algorithm proceeds. The difference is that \rrtstar\ adds one state
per iteration, while \sststar\ adds a set of states per batch of
iterations. Generally speaking, all sampling-based algorithms need to
increasingly add states to cover the space. With this approach,
sampling-based algorithms avoid knowing the minimum clearance
parameter $\delta$.


In \sststar, the data structure is always a tree, meaning that, at any
moment there are $n$ edges and $n+1$ vertices (\rrtstar\ trims edges from the underlying {\tt RRG} graph). The system
accessibility property (\emph{Ball-Box} theorem) guarantees that it is
possible to extend edges from one ball region to the next. It is also
possible to argue that this will happen almost
surely. The \emph{Lipschitz continuity} assumption of the cost
function allows a near-optimal bound on the trajectories.  The
best-first selection strategy and the pruning process make the above
guarantees practical and computationally efficient.

\section{Experimental Evaluation}
\label{sec:experiments}

\begin{figure}[h]
\centering
  \parbox{.24\textwidth}{\includegraphics[width=.24\textwidth]{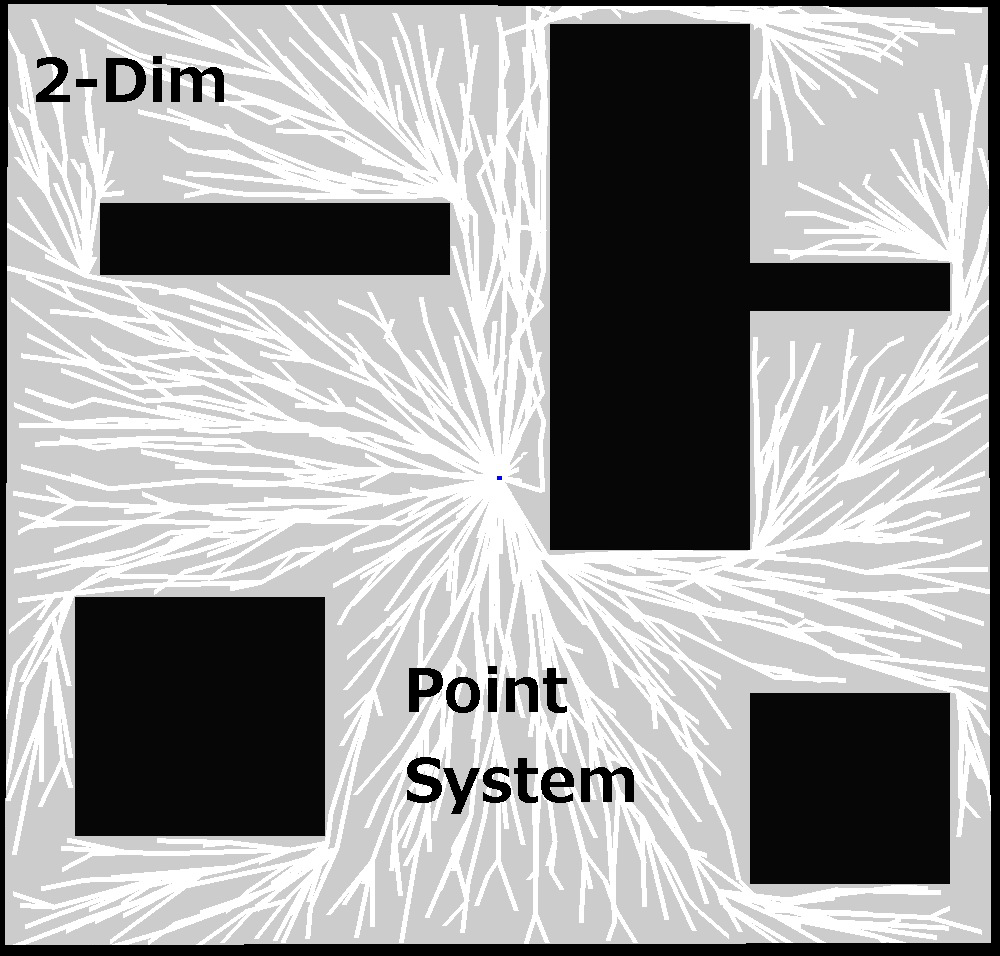}}
  \parbox{.24\textwidth}{\includegraphics[width=.23\textwidth]{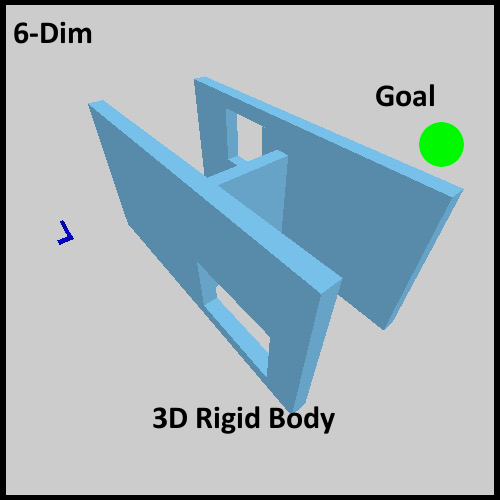}}
  \parbox{.24\textwidth}{\includegraphics[width=.24\textwidth]{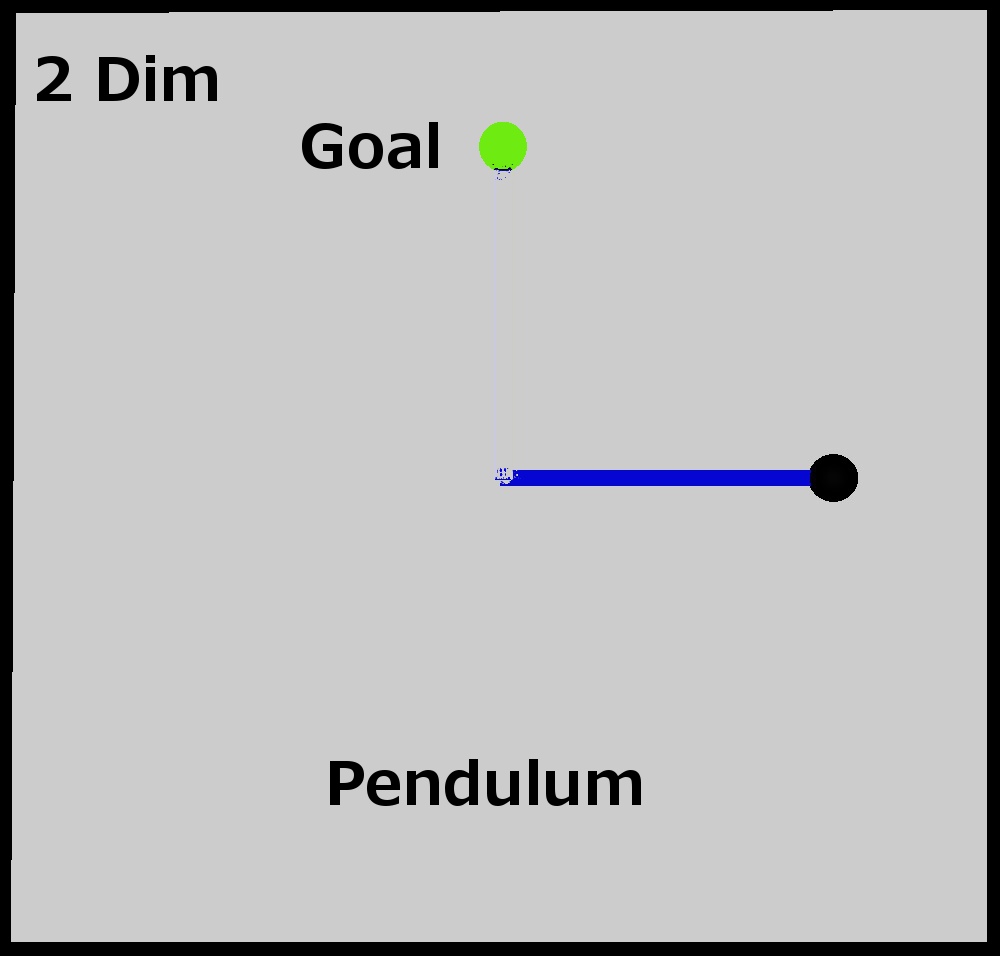}}
  \parbox{.24\textwidth}{\includegraphics[width=.24\textwidth]{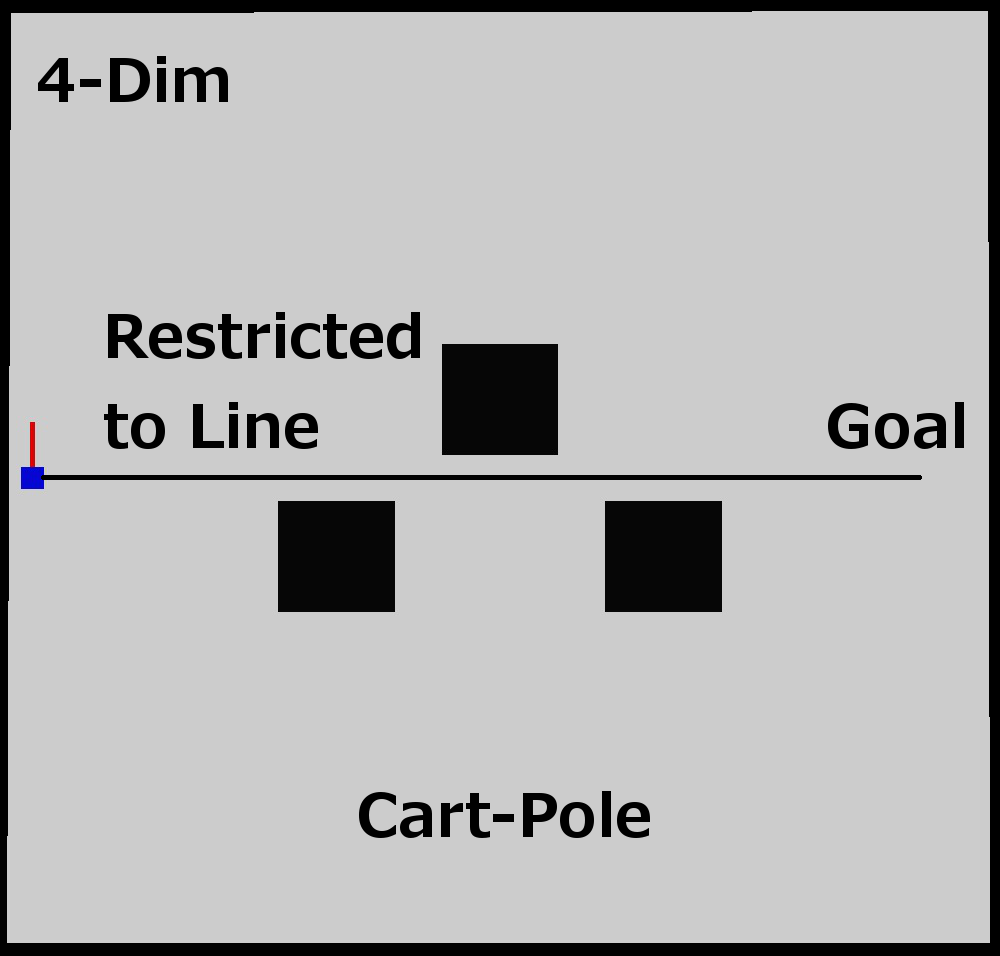}}\\
  \parbox{.24\textwidth}{\includegraphics[width=.24\textwidth]{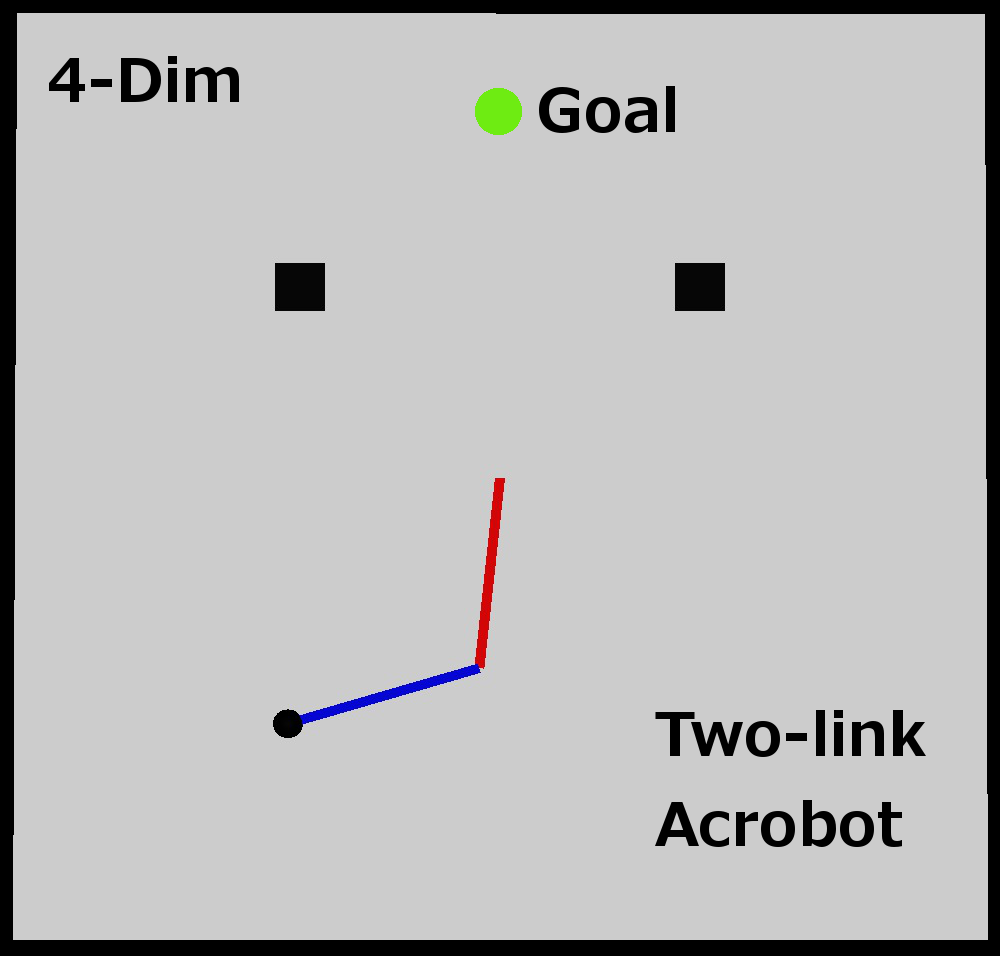}}
  \parbox{.24\textwidth}{\includegraphics[width=.24\textwidth]{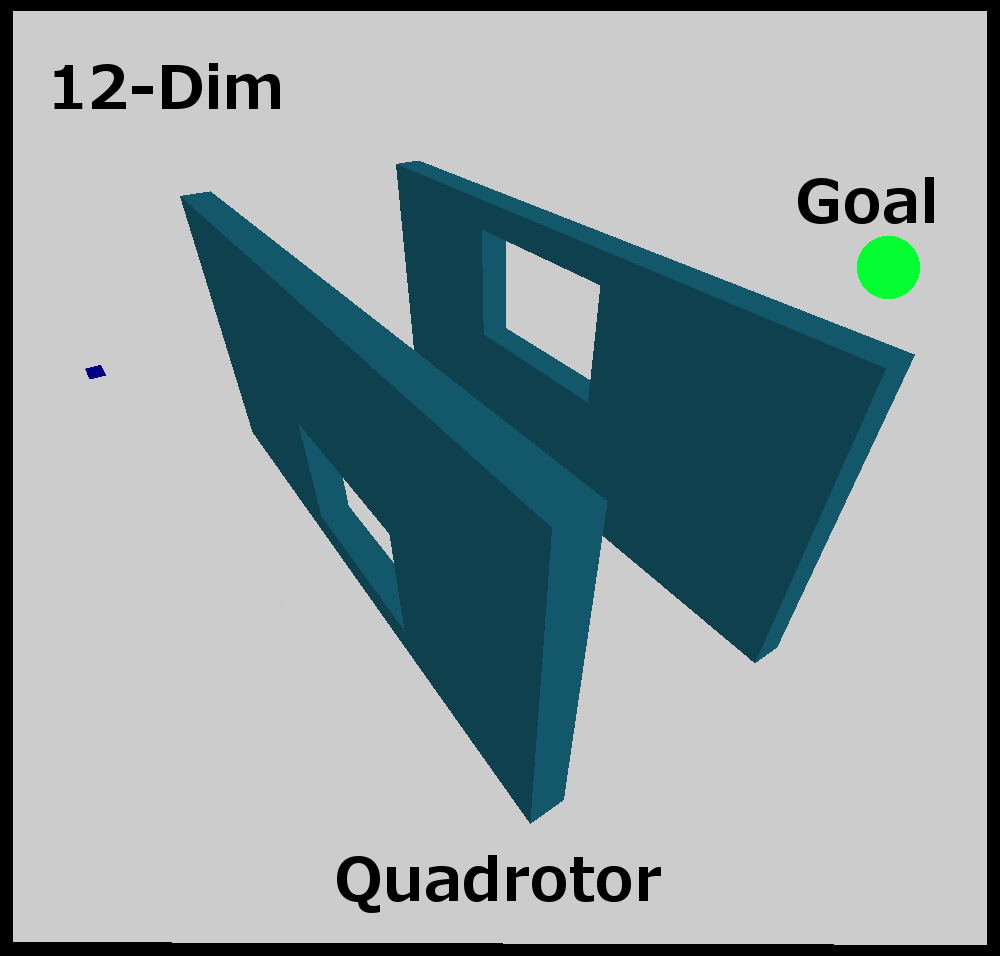}}
  \parbox{.24\textwidth}{\includegraphics[width=.24\textwidth]{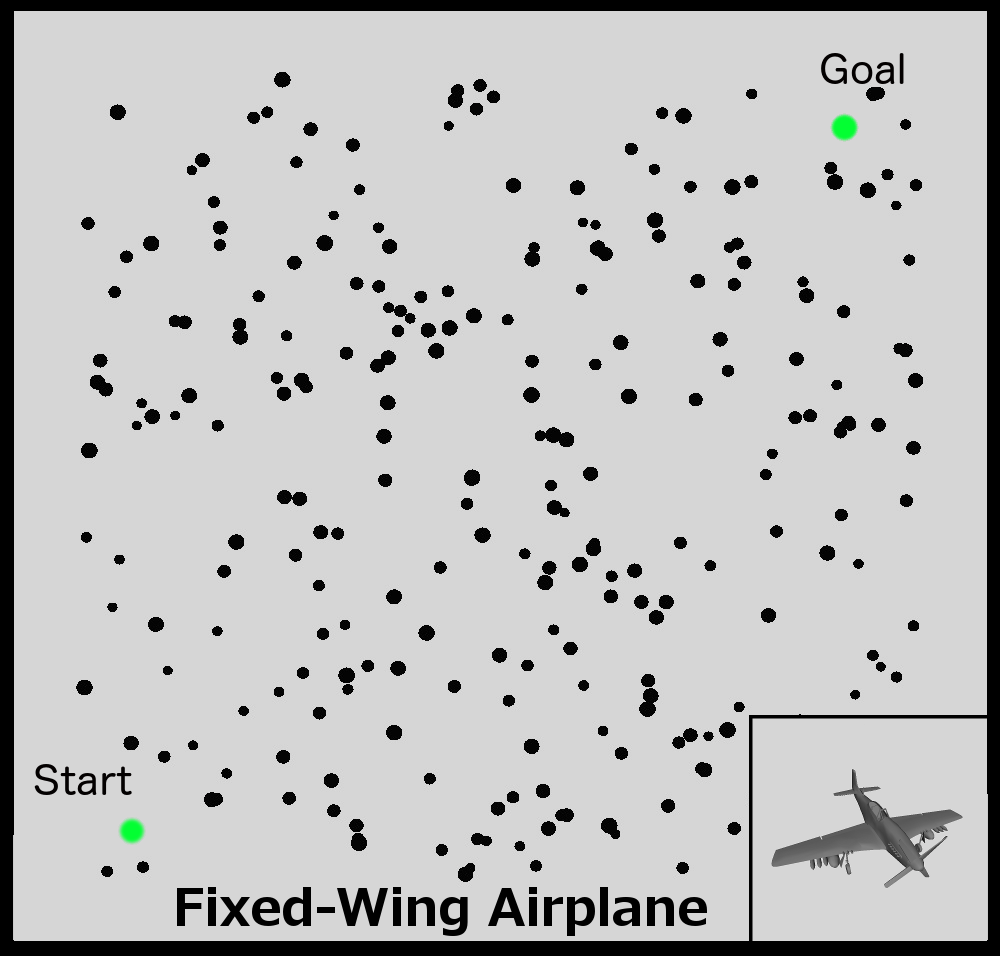}}
  \caption{The different benchmarks. From left to right and top to
    bottom, a kinematic point, \edited{3D rigid body}, a pendulum, a
    cart-pole among obstacles, a passive-active acrobot, a 12-dim
    quadrotor, fixed-wing aircraft \edited{(with much more restricted
    movement compared to the quadrotor)}. Each experiment is averaged
    over 50 runs of each algorithm.}
\vspace{-.1in}
\label{fig:environ}
\end{figure}

\begin{table*}[h!]
\begin{center}
\begin{tabular}{|p{.2\textwidth}||p{.3\textwidth}|p{.2\textwidth}|p{.02\textwidth}|p{.02\textwidth}|}
\hline
System & Parameters & Distance Function & $\Ddrain$ & $\Dnear$ \\
\hline
Kinematic Point & 2 Dim. State, 2 Dim. Control & Euclidean Distance & .5 & 1 \\
\hline
\edited{3D Rigid Body} & \edited{6 Dim. State, 6 Dim. Control} & \edited{Euclidean Distance} & \edited{2} & \edited{4 }\\
\hline
Simple Pendulum & 2 Dim. State, 1 Dim. Control, No Damping & Euclidean Distance& .2 & .3\\
\hline
Two-Link Acrobot  \citep{spong_acrobot} & 4 Dim. State, 1 Dim. Control,& Euclidean Distance & .5 & 1\\
\hline
Cart-Pole \citep{Papadopoulos2014} & 4 Dim. State, 1 Dim. Control, & Euclidean Distance & 1 & 2 \\
\hline
Quadrotor \citep{Ai-Omari2013Integrated-simu}& 12 Dim. State, 4 Dim. Control, & Distance in ${\mathbb SE}3$& 3 & 5\\
\hline
Fixed-Wing Airplane \citep{Paranjape2013Motion-primitiv} & 9 Dim. State, 3 Dim. Control,  & Euclidean Distance in ${\mathbb R}^3$ & 2 & 6\\
\hline
\end{tabular}
\end{center}
\caption{The experimental setup used to evaluate \sst. Parameters are
available in the corresponding references. \edited{Values for
$\Ddrain$ and $\Dnear$ have been selected based on the features of
each planning challenge.} }
\label{tab:expsetups}
\end{table*}

In order to evaluate the proposed method, a set of experiments
involving several different systems have been conducted. The proposed
algorithm \sst\ is compared against \rrt\ \citep{LaValle2001} as a
baseline and also with another algorithm: (a) if a steering function
is available, a comparison with
\rrtstar\ \citep{Karaman2011Sampling-based-} is conducted, (b) if
\rrtstar\ cannot be used, a comparison with an alternative based on a
``shooting'' function is
utilized \citep{Jeon2011Anytime-Computa}. Different versions of \rrt\
were evaluated depending on the benchmark. In the case where a
steering function is available, \rrt\ corresponds to \rrtconnect. When
a steering function is not available, a version of \rrt\
using \mcprop\ is used, which is similar to \rrtextend.


The overall results show that \sst\ can provide consistently improving
path quality given more iterations as \rrtstar\ does for kinematic
systems, achieving running times equivalent (if not better \edited{than)} \rrt,
and maintaining a small number of nodes, all while using a very simple
random propagation primitive.

\begin{figure}[h!]
\centering
  \parbox{.49\textwidth}{\includegraphics[width=.49\textwidth]{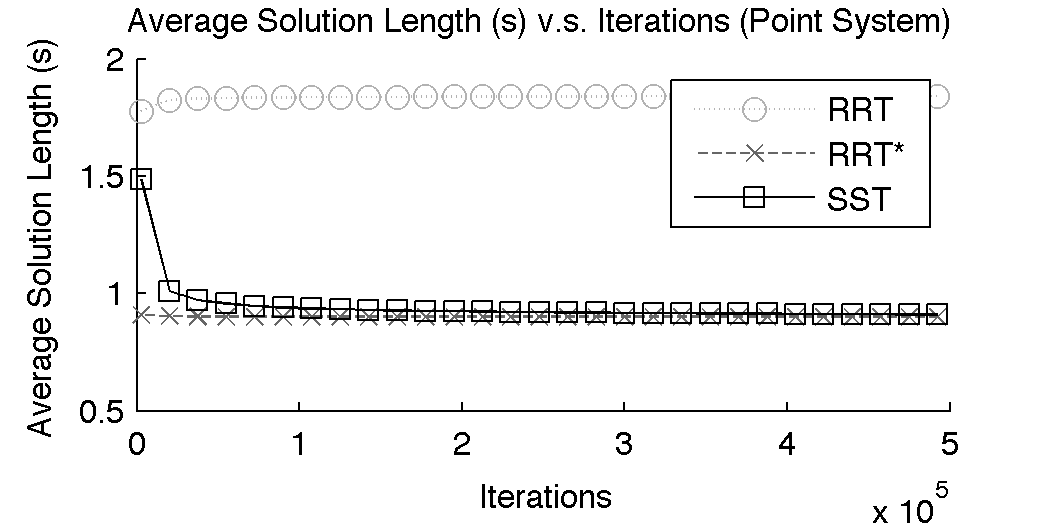}}
  \parbox{.49\textwidth}{\includegraphics[width=.45\textwidth]{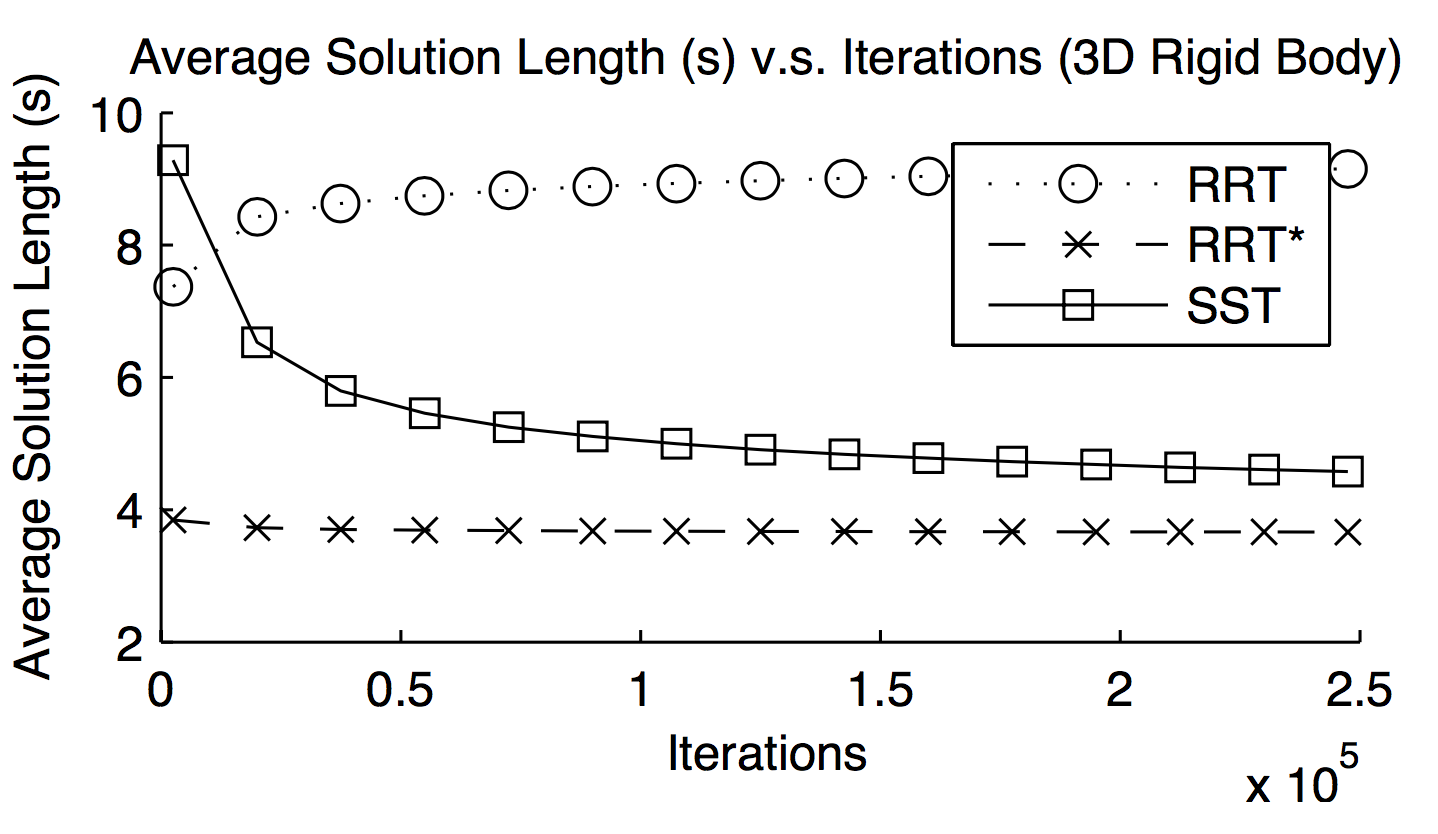}}
  \parbox{.49\textwidth}{\includegraphics[width=.49\textwidth]{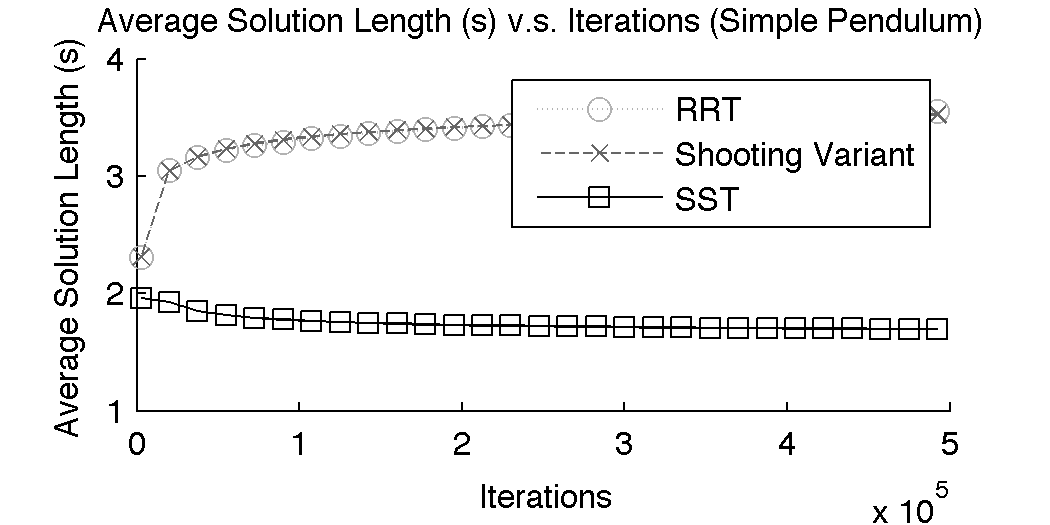}}
  \parbox{.49\textwidth}{\includegraphics[width=.49\textwidth]{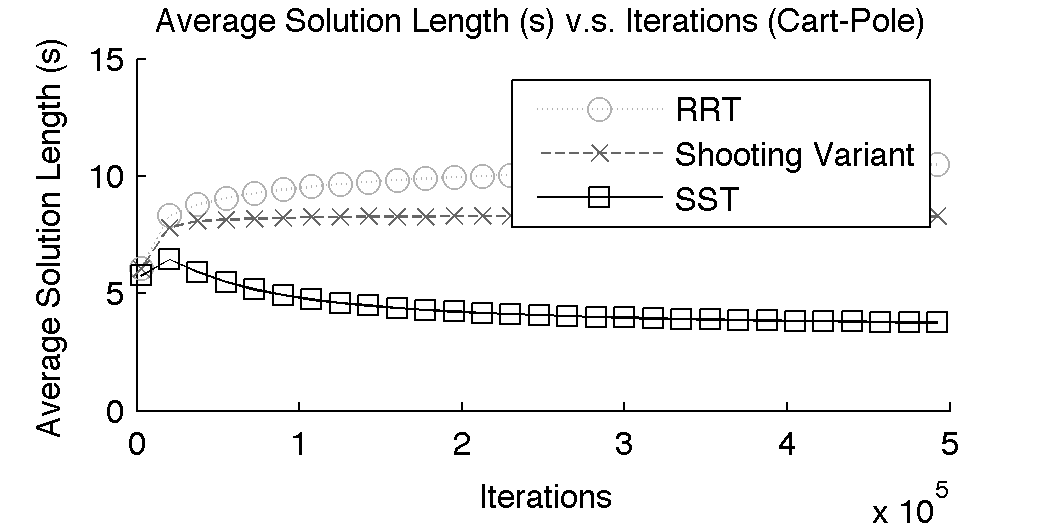}}
  \parbox{.49\textwidth}{\includegraphics[width=.49\textwidth]{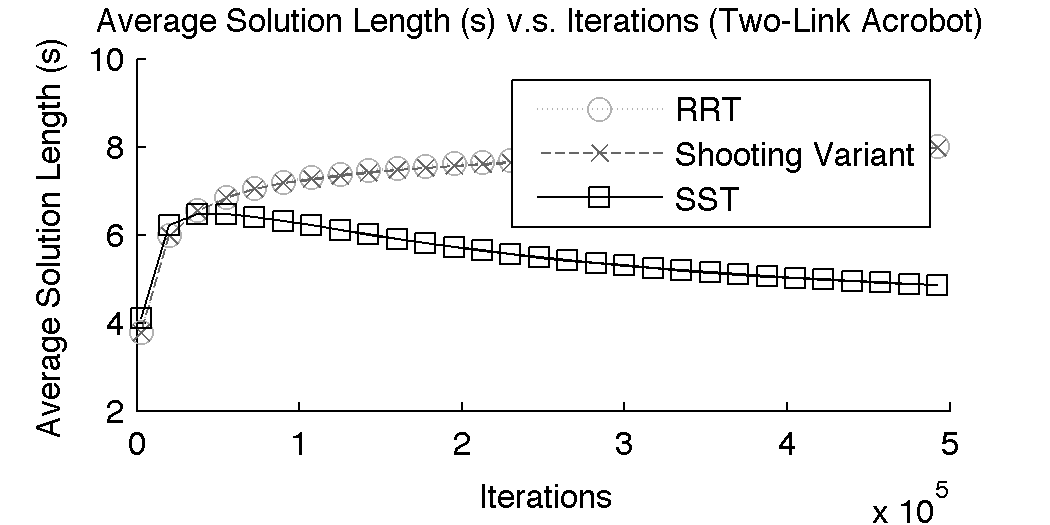}}
  \parbox{.49\textwidth}{\includegraphics[width=.49\textwidth]{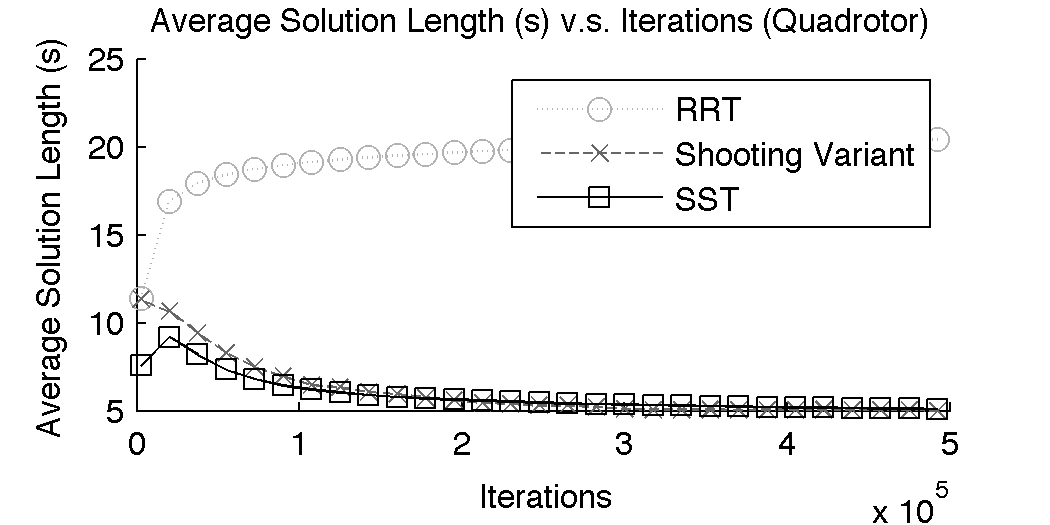}}
  \parbox{.49\textwidth}{\includegraphics[width=.49\textwidth]{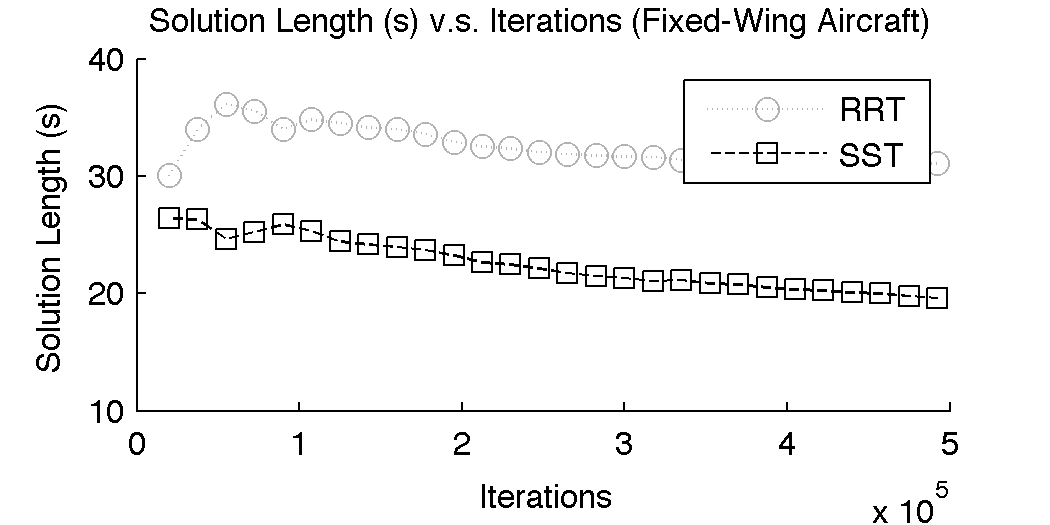}}
  \caption{The average cost to each node in the tree for each algorithm (\rrt, \rrtstar\ or the shooting approach, and \sst).}
\label{fig:qual}
\end{figure}

Figure \ref{fig:environ} illustrates the various setups that the
algorithms have been evaluated on and Table \ref{tab:expsetups} shows
details about the experimental setups. \edited{The parameters of \sst\ are chosen by hand 
from an expert user, but could be determined by examining performance of previous attempts. }

\noindent
\edited{ 
\textbf{Kinematic Point.} A simple system for a baseline comparison. The state space is 2D
  $(x,y)$, the control space is 2D
  $(v,\theta)$, and the dynamics are:
\vspace{-.1in}
\begin{align*}
\dot{x} = v \cos(\theta)\ \ \ \ \ \dot{y} &= v \sin(\theta).\\
\end{align*}
}

\noindent
\edited{
\textbf{3D Rigid Body.} A free-flying rigid body. The state space is 6D
  $(x,y,z,\alpha,\beta,\gamma)$ signifying the space of SE(3) and the control space is 6D
  $(\dot{x},\dot{y},\dot{z},\dot{\alpha},\dot{\beta},\dot{\gamma})$ representing the velocities of these degrees of freedom.
}

\noindent
\edited{
\textbf{Simple Pendulum.} A pendulum system typical in control literature. The state space is 2D
  $(\theta,\dot{\theta})$, the control space is 1D
  $(\tau)$, and the dynamics are:
\vspace{-.1in}
\begin{align*}
\ddot{\theta} = \frac{(\tau - mgl*\cos(\theta)*0.5)* 3}{ml^2}.
\end{align*}
where $m=1$ and $l=1$.
}

\noindent
\edited{
\textbf{Cart-Pole.} Another typical control system where a block mass on a track has to balance a pendulum. The state space is 4D
  $(x,\theta,\dot{x},\dot{\theta})$ and the control space is 1D
  $(f)$ which is the force on the block mass. The dynamics are from \citep{Papadopoulos2014}.
}

\noindent
\edited{
\textbf{Two-link Acrobot.} The two-link acrobot model with a passive root joint. The state space is 4D
  $(\theta_1,\theta_2,\dot{\theta_1},\dot{\theta_2})$ and the control space is 1D
  $(\tau)$ which is the torque on the active joint. The dynamics are from \citep{spong_acrobot}.
}

\noindent
\edited{
\textbf{Fixed-wing airplane.} An airplane flying among
  cylinders. The state space is 9D
  $(x,y,z,v,\alpha,\beta,\theta,\omega,\tau)$, the control space is 3D
  $(\tau_{des},\alpha_{des},\beta_{des})$, and the dynamics are from
  \citep{Paranjape2013Motion-primitiv}.
}

\noindent
\edited{
\textbf{Quadrotor.} A quadrotor flying through windows. The state space is 12D
  $(x,y,z,\alpha,\beta,\gamma,\dot{x},\dot{y},\dot{z},\dot{\alpha},\dot{\beta},\dot{\gamma})$, the control space is 4D
  $(w_1,w_2,w_3,w_4)$ corresponding to the rotor torques, and the dynamics are from
  \citep{Ai-Omari2013Integrated-simu}.
}


\begin{figure}[h!]
\centering
  \parbox{.49\textwidth}{\includegraphics[width=.49\textwidth]{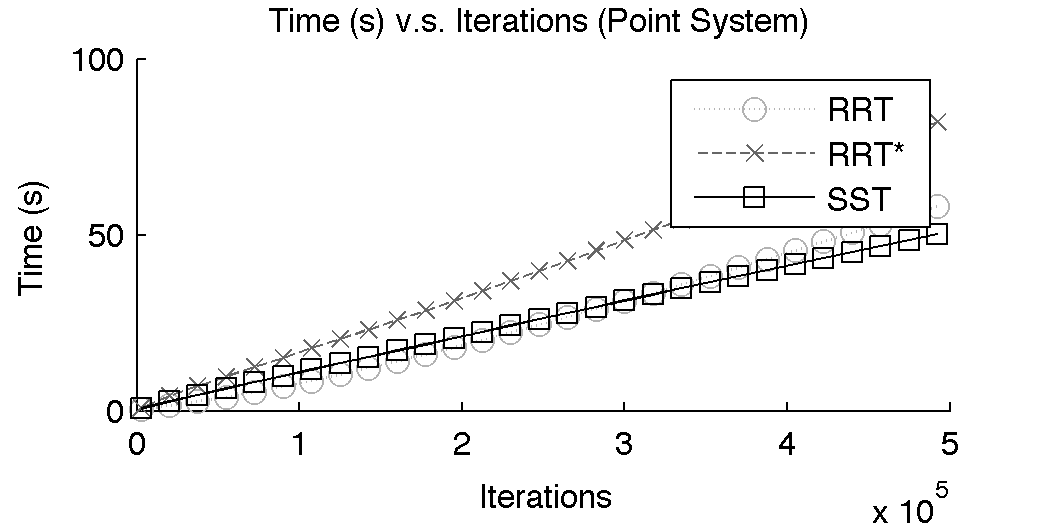}}
  \parbox{.49\textwidth}{\includegraphics[width=.45\textwidth]{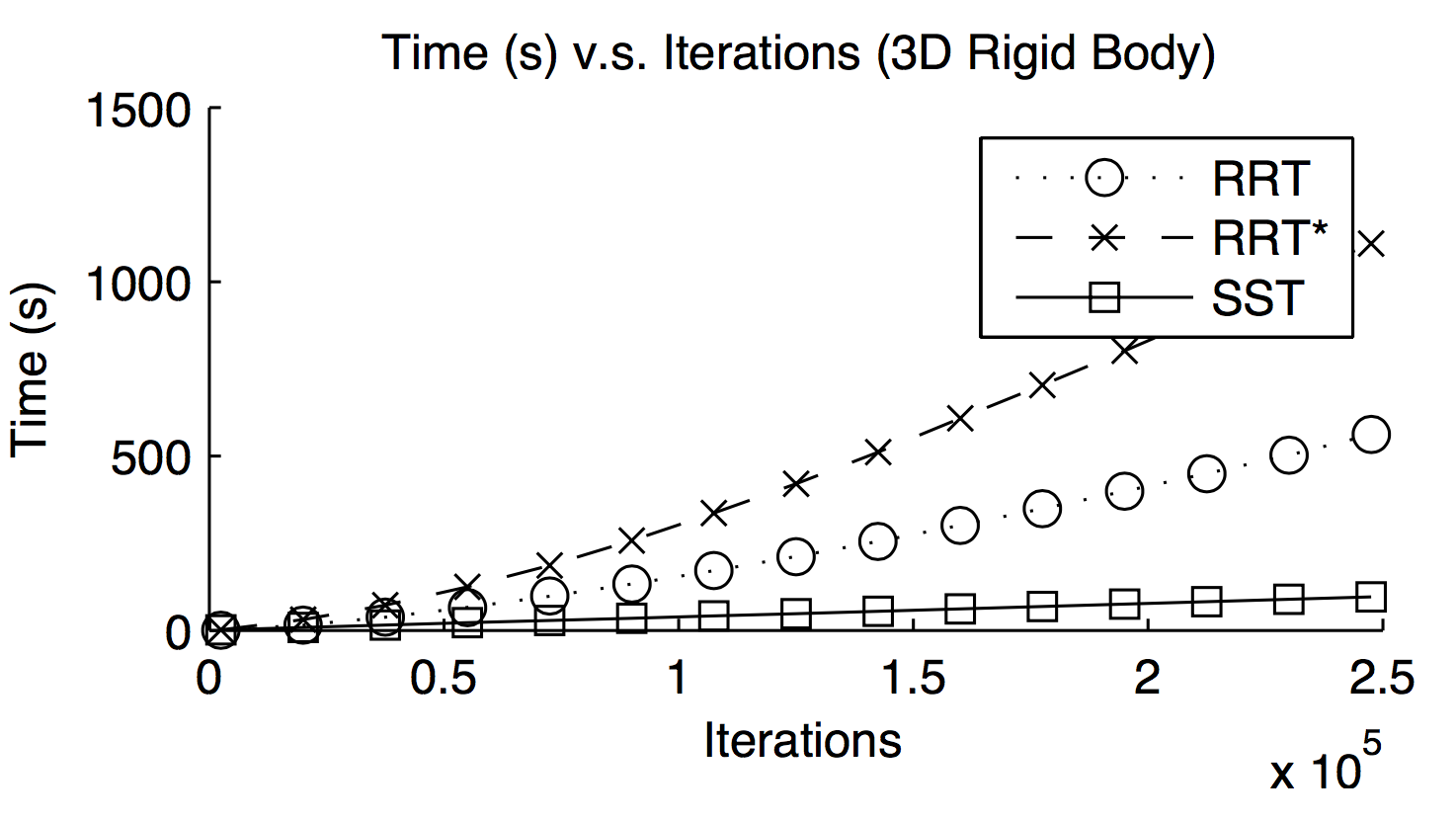}}
  \parbox{.49\textwidth}{\includegraphics[width=.49\textwidth]{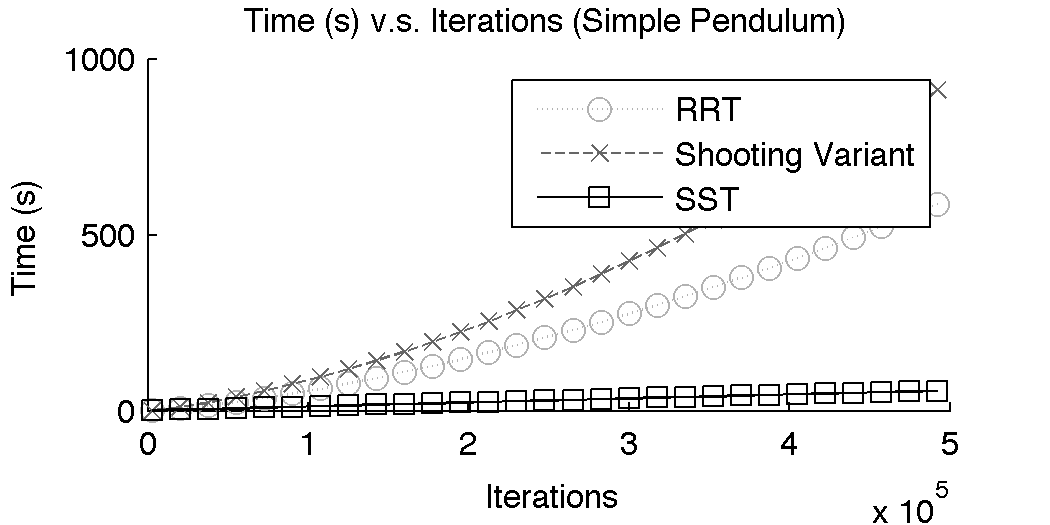}}
  \parbox{.49\textwidth}{\includegraphics[width=.49\textwidth]{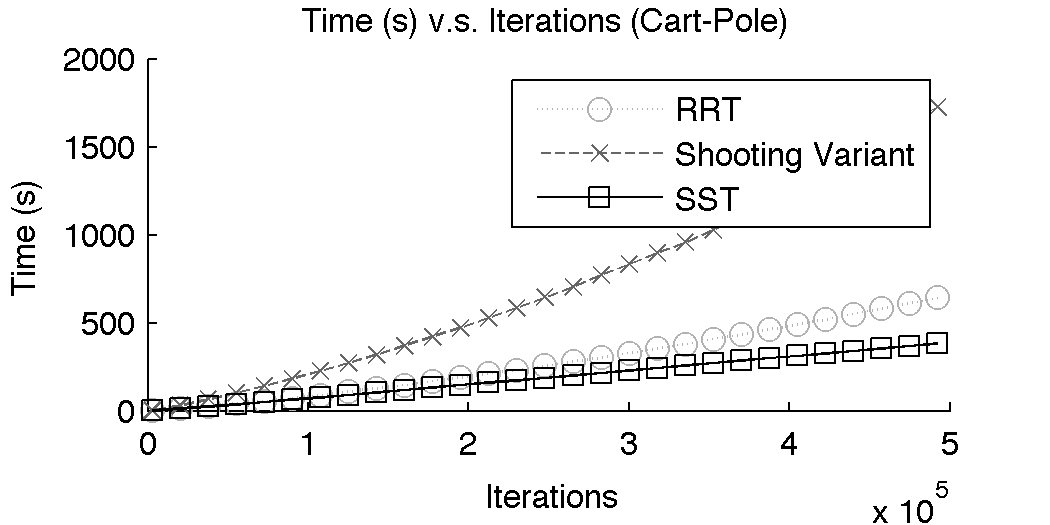}}
  \parbox{.49\textwidth}{\includegraphics[width=.49\textwidth]{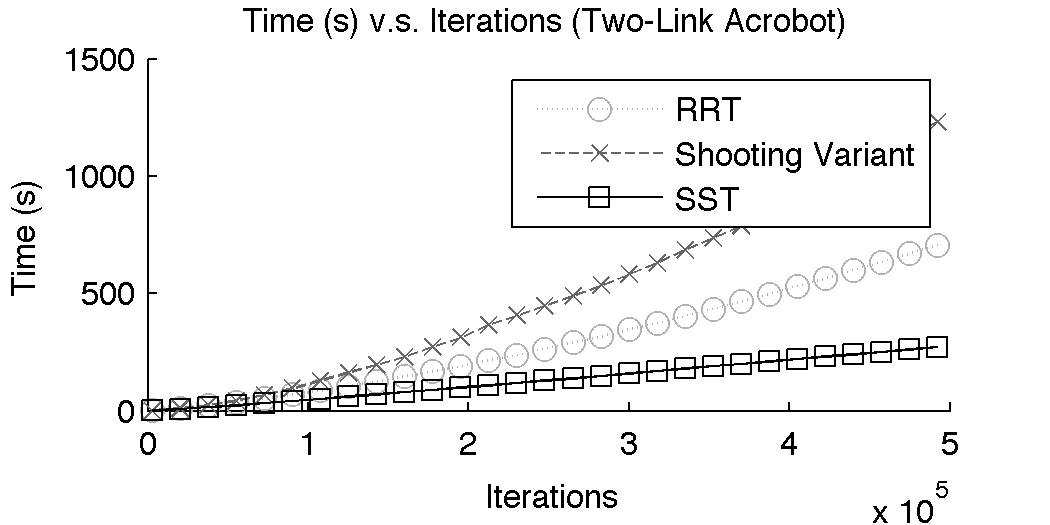}}
  \parbox{.49\textwidth}{\includegraphics[width=.49\textwidth]{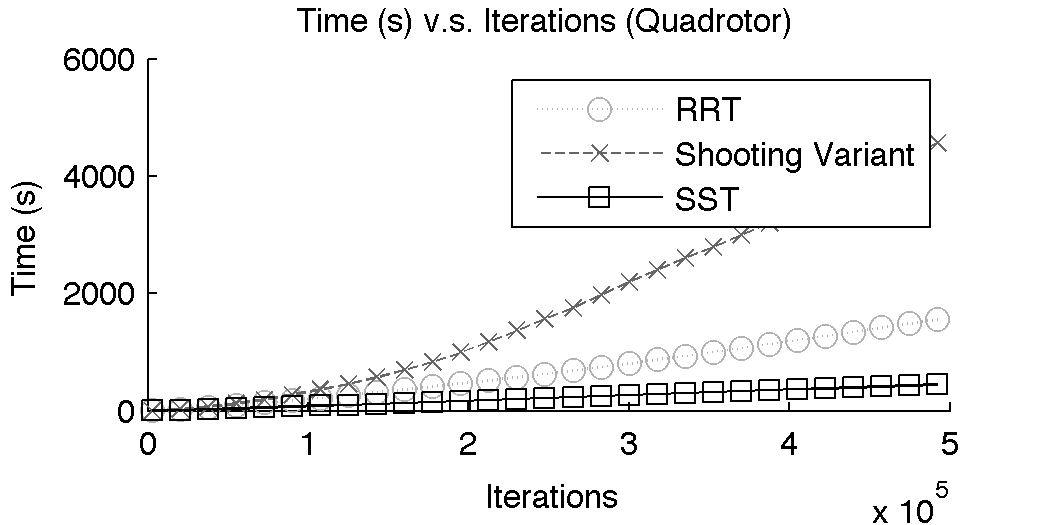}}
  \parbox{.49\textwidth}{\includegraphics[width=.49\textwidth]{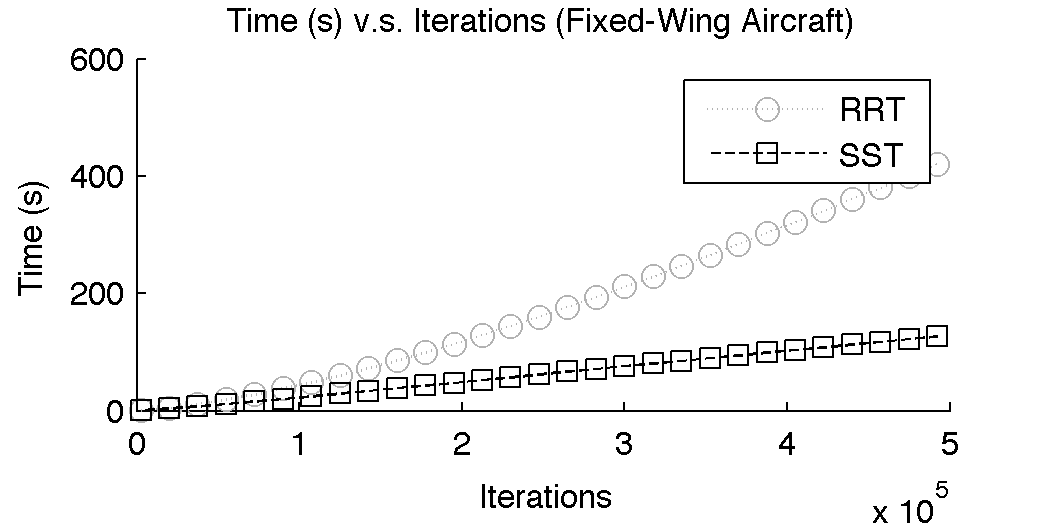}}
  \caption{The time for execution for each algorithm (\rrt, \rrtstar\ or the shooting approach, and \edited{\sst)}.}
\label{fig:time}
\end{figure}

\subsection{Quality of Solution Trajectories}

In Figure \ref{fig:qual} the average solution quality to nodes in each
tree is shown. This average is a measure of the quality of
trajectories generated \edited{to all reachable parts of the state space}. In every case, \sst\ is able to improve
quality over time. By looking at all of the nodes in the tree as a
whole, the global behavior of improving path costs can be
observed. \rrt\ will increase this average over time because it
chooses suboptimal nodes and further propagates them, thus making
those average values increase over time.

It is interesting to note that the approach based on the shooting
function had varying success in these scenarios. The systems with
highly nonlinear dynamics (e.g., all the systems with a pendulum-like
behavior) did not perform better than \rrt. This could result from the
choice of distance function for these scenarios or from the inaccuracy
in the shooting method. Notably, \sst\ does not have this problem for
the same distance function and with random propagations and continues
to provide good performance. The shooting method did perform well in
the quadrotor environment, but failed to return solutions for most of 
the fixed-wing airplane runs and was therefore omitted.

\subsection{Time Efficiency}

Figure \ref{fig:time} shows time vs. iterations plots for each of the
systems. The graphs show the amount of time it took to achieve a
number of iterations.  The running time of \sst\ is always comparable
or better than \rrt. \rrtstar\ has a higher time cost per iteration as
expected.  Initially \sst\ is slightly slower than \rrt\ for the
kinematic point, but becomes increasingly more efficient later
on. This is explained by Lemma \ref{lem:complexitySST}, since
\sst\ has better running time than \rrt\ given the sparse data
structure.

\sst\ has another advantage over other \rrt\ variants. Due to the
pruning operation, there is another criterion in addition to being
collision-free that newly generated states must satisfy to be added to
the tree. Any new state must both be collision-free and dominant in
the region around the witness sample in $S$. Because of this, the
collision check at Line 8 of \edited{Algorithm} \ref{alg:SST} can be shifted to
after Line 15. In the event that collision checking is more expensive
than a nearest neighbor query in $S$, this can result in improved
computational efficiency depending on the scenario. This strategy was
not used in these experiments, but can be beneficial in domains
where collision checking is the dominant computational factor.

\begin{figure}[h!]
\centering
  \parbox{.49\textwidth}{\includegraphics[width=.49\textwidth]{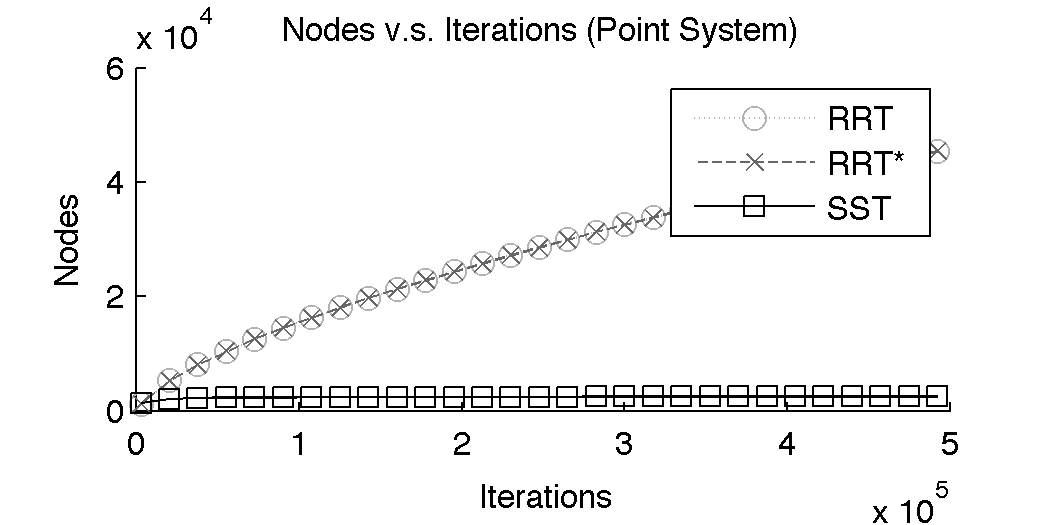}}
  \parbox{.49\textwidth}{\includegraphics[width=.45\textwidth]{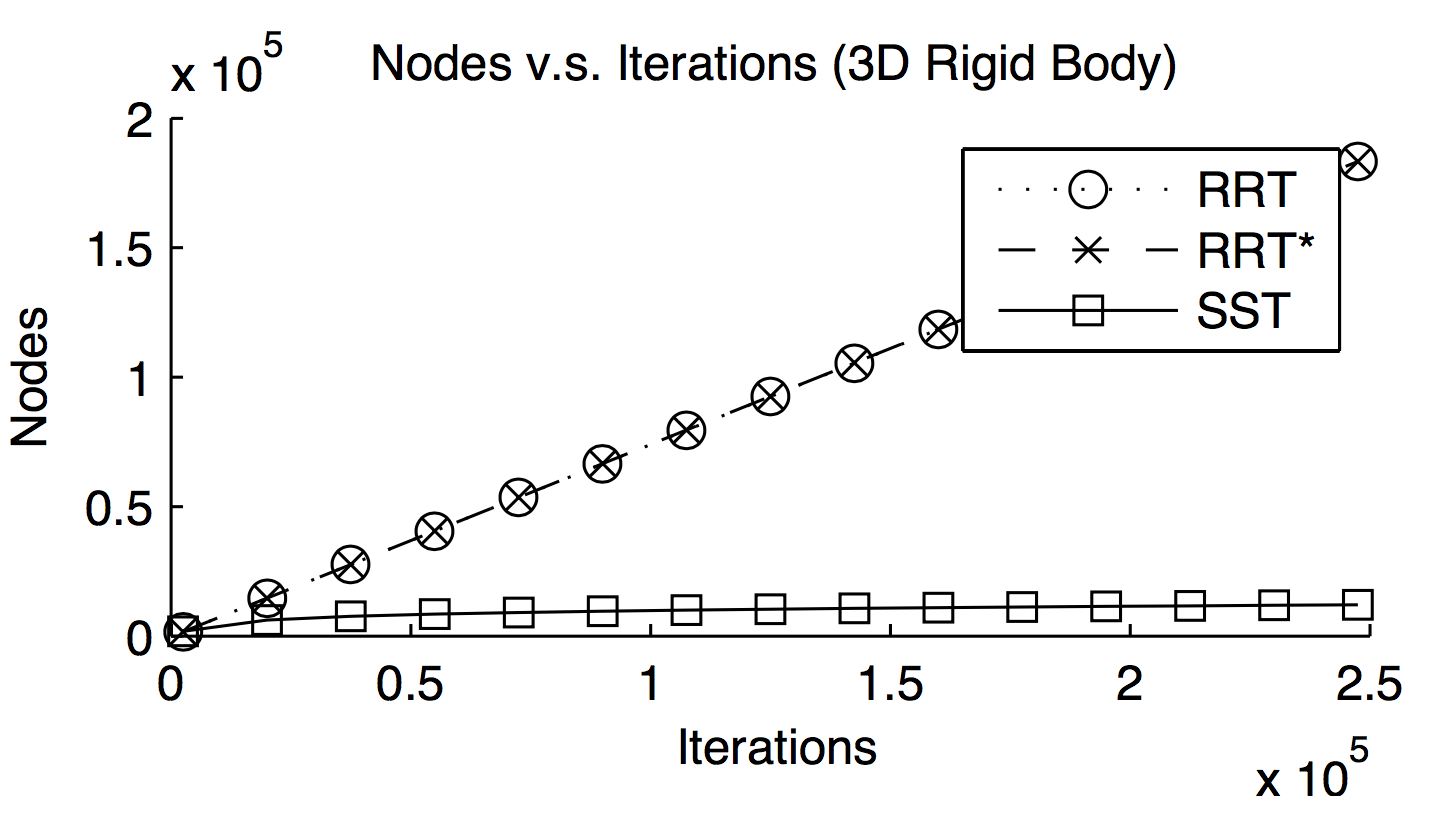}}
  \parbox{.49\textwidth}{\includegraphics[width=.49\textwidth]{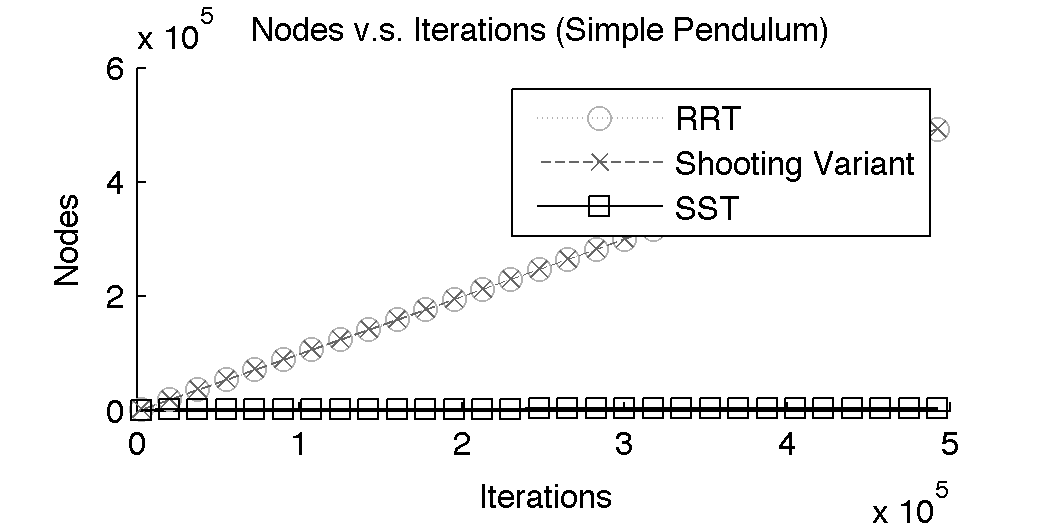}}
  \parbox{.49\textwidth}{\includegraphics[width=.49\textwidth]{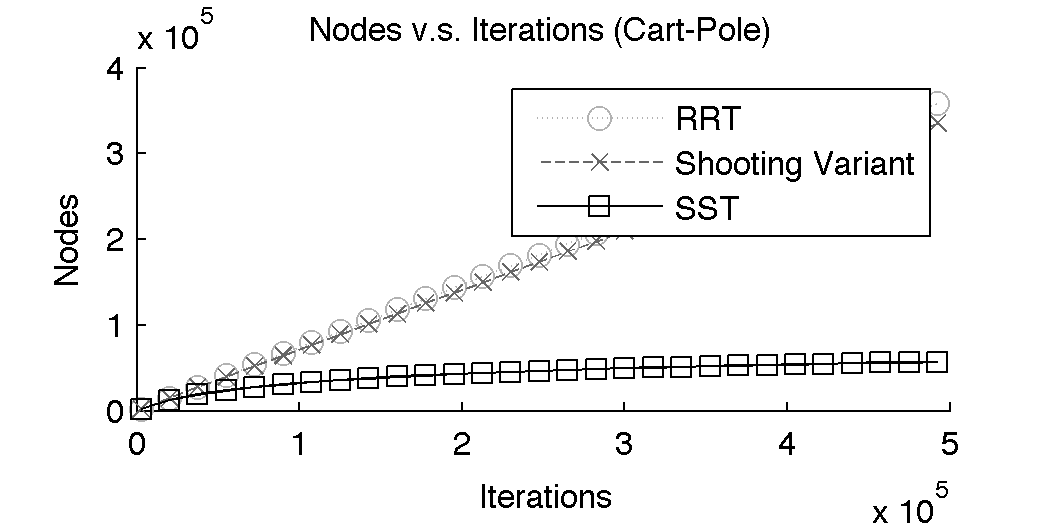}}
  \parbox{.49\textwidth}{\includegraphics[width=.49\textwidth]{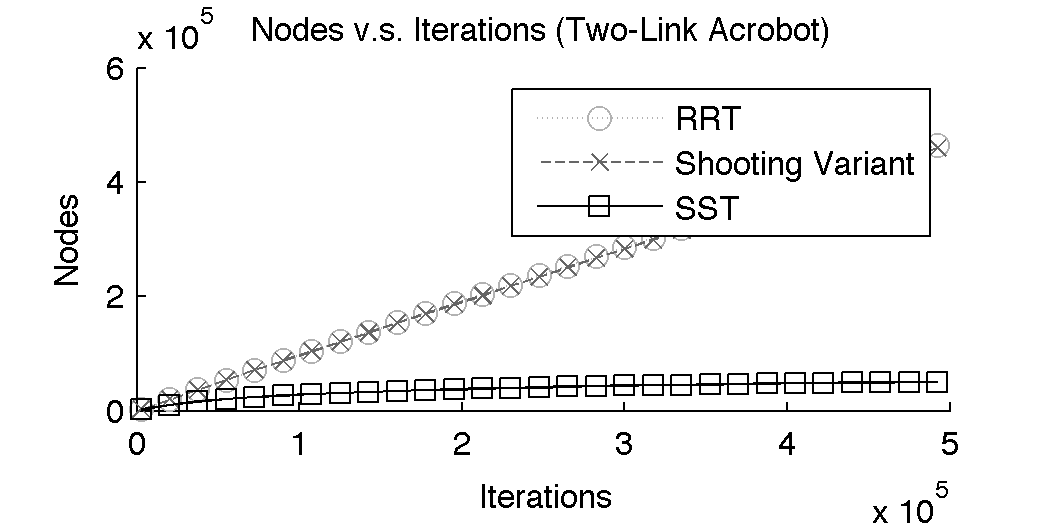}}
  \parbox{.49\textwidth}{\includegraphics[width=.49\textwidth]{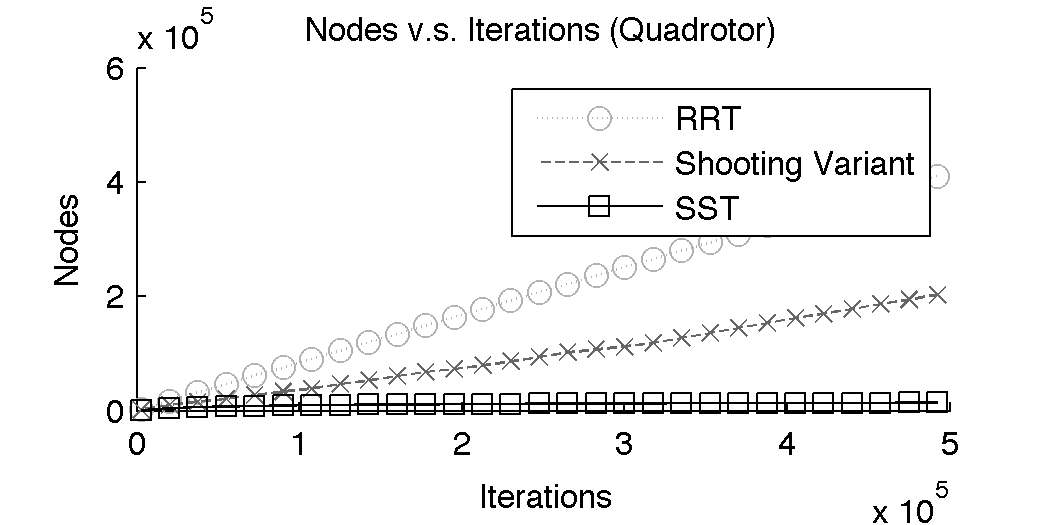}}
  \parbox{.49\textwidth}{\includegraphics[width=.49\textwidth]{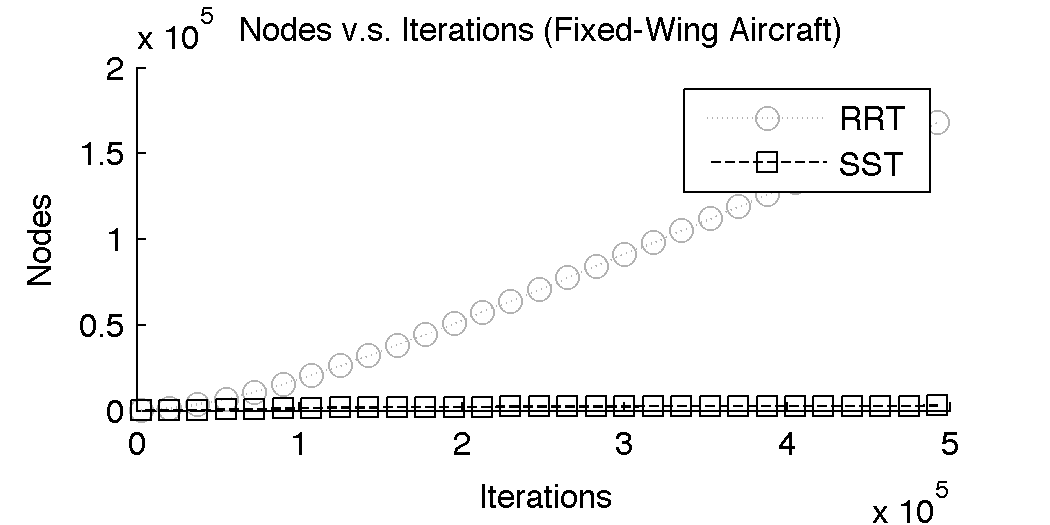}}
  \caption{The number of nodes stored in each algorithm (\rrt, \rrtstar\ or the shooting approach, and \edited{\sst)}.}
\label{fig:nodes}
\end{figure}

\subsection{Space Efficiency}

One of the major gains of using \sst\ is the smaller number of
nodes that are needed in the data structure. Figure \ref{fig:nodes}
shows the number of nodes stored by each of the algorithms.  The
number of nodes is significantly lower in \sst, even when considering
the witness set $S$. The sparse data structure of \sst\ makes the
memory requirements quite small, in contrast to \rrt\ and \rrtstar,
which do not perform any pruning operations. In the case of shooting,
sometimes the inaccuracy of the shooting primitive will cause
collisions to occur in resimulated trees, pruning them from the data
structure. This can lead to losing solution trajectories.

These results showcase the large efficiency gains when a sparse data
structure can be generated. There is a tradeoff, however,
between the sparseness of the data structure and allowing for a
diverse set of paths to be generated.  Path diversity can be helpful
for discovering the homotopic class of the optimal solution in
practice. In all of these scenarios, there is either only one
homotopic class for solutions or the pruning radius $\Ddrain$ is small
enough to allow each homotopic class to be potentially explored. Even
considering this, significant pruning can still be achieved.

\edited{One can draw  parallels between \sst\ and grid-based
methods, as both methodologies end up maintaining a discrete set of
witness states in the state space. One concern with grid-based
approaches is that they have an exponential dependency in the
dimensionality of the state space. In the worst case, \sst\ shares the
same property. At the same time, however, it has certain
advantages. Typically, the discretization followed by grid-based
methods corresponds to fixed witnesses defined before the problem is
known. In \sst\ the witnesses arise on the fly and are adaptive to the
features of the state space. A benefit of following this approach is
the capability to find solutions sooner in practice without explicitly
constructing or reasoning over the entire grid, which has an
exponential number of points. After an initial solution is found,
witness nodes can be removed, improving space complexity even further,
similar to branch-and-bound techniques.}

\subsection{\edited{Dependence on Parameters}}

\begin{table*}
\begin{center}
\noindent
\resizebox{\textwidth}{!} {
\begin{tabular}{cc|*{15}{c|}} 
\cline{3-17} & & \multicolumn{15}{ c| }{$\Dnear$} \\ 
\cline{3-17} & & 	\multicolumn{3}{ c| }{$1.0$} &
					\multicolumn{3}{ c| }{$1.2$} &
					\multicolumn{3}{ c| }{$1.4$} &
					\multicolumn{3}{ c| }{$1.6$} &
					\multicolumn{3}{ c| }{$1.8$}\\ 
\cline{3-17} & & IT & IC & FC & IT & IC & FC & IT & IC & FC & IT & IC & FC & IT & IC & FC \\ 
\cline{1-17} \multicolumn{1}{ |c }{\multirow{5}{*}{$\Ddrain$} } 
& \multicolumn{1}{ |c| }{0.2} & 0.1105 & 3.4201 & 1.7782 & 0.1296 & 3.2891 & 1.7798 & 0.1516 & 3.2248 & 1.7866 & 0.1687 & 3.0865 & 1.7890 & 
0.2095 & 3.0641 & 1.7949 \\ 
\cline{2-17} \multicolumn{1}{ |c }{} 
& \multicolumn{1}{ |c| }{0.4} & 0.1190 & 3.2445 & 1.7851 & 0.0915 & 3.2614 & 1.7797 & 0.0938 & 3.1364 & 1.7833 & 0.0961 & 3.0506 & 1.7829 & 0.0906 & 3.1027 & 1.7852 \\ 
\cline{2-17} \multicolumn{1}{ |c }{} 
& \multicolumn{1}{ |c| }{0.6} & 0.0603 & 3.2155 & 1.7916 & 0.0999 & 3.2105 & 1.7973 & 0.0670 & 2.9795 & 1.7988 & 0.0671 & 2.9523 & 1.7987 & 0.0679 & 2.8082 & 1.7971 \\ 
\cline{2-17} \multicolumn{1}{ |c }{} 
& \multicolumn{1}{ |c| }{0.8} & 0.0451 & 3.0468 & 1.8229 & 0.0593 & 2.9554 & 1.8273 & 0.0498 & 2.8908 & 1.8193 & 0.0545 & 2.8334 & 1.8232 & 0.0724 & 2.6549 & 1.8416 \\ 
\cline{2-17} \multicolumn{1}{ |c }{} 
& \multicolumn{1}{ |c| }{1.0} & 0.0548 & 2.7695 & 1.8627 & 0.0635 & 2.7371 & 1.8723 & 0.0567 & 2.7365 & 1.8621 & 0.0595 & 2.7185 & 1.8853 & 0.0601 & 2.7493 & 1.8846 \\ 
\cline{1-17} 
\end{tabular}
}
\caption{\edited{A comparison of different parameter choices in
    \sst. The problem setup is the 2D point where the distance
    function is the typical Euclidean metric. For each parameter
    selection, the time to compute an initial solution (IT), the
    initial solution cost in seconds (IC), and final solution cost in
    seconds (FC) after 60 seconds of execution time.} }
\label{tab:param_eval}
\end{center}
\end{table*}






\edited{Table \ref{tab:param_eval} shows statistics for running \sst\
with several different parameter choices. The problem setup is the
simple case of the 2D kinematic point. Larger values for the pruning
radius, $\Ddrain$, result in initial solutions being discovered
sooner. Larger values also restrict the convergence to better
solutions. Larger values for the selection radius, $\Dnear$, provide
better solution cost for initial solutions, but requires more
computational effort. These tradeoffs can be weighed for the
application area depending on the importance of finding solutions
early and the quality of those solutions.}

\subsection{Physically-simulated Car Evaluation}

\begin{figure}[h!]
\centering
  \parbox{.33\textwidth}{\includegraphics[width=.33\textwidth]{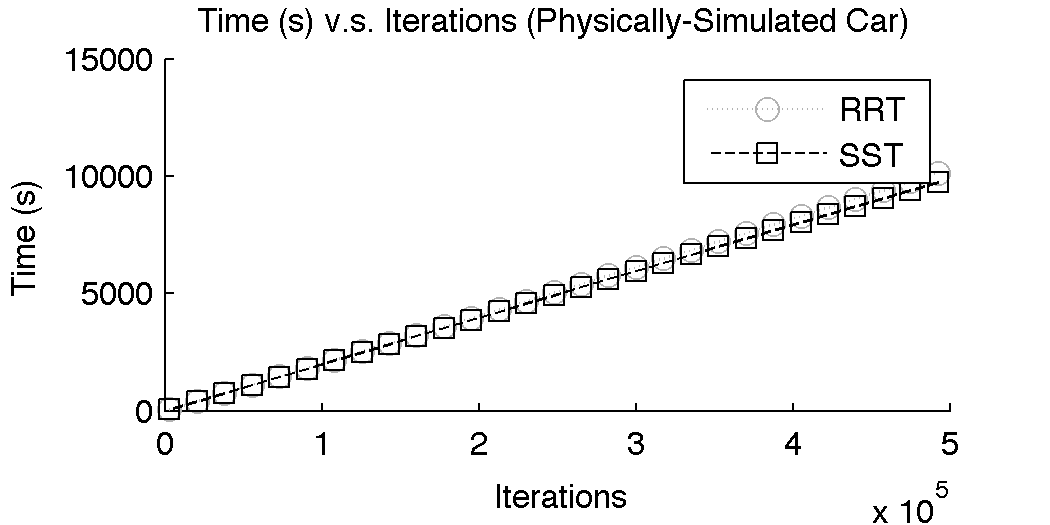}} \parbox{.33\textwidth}{\includegraphics[width=.33\textwidth]{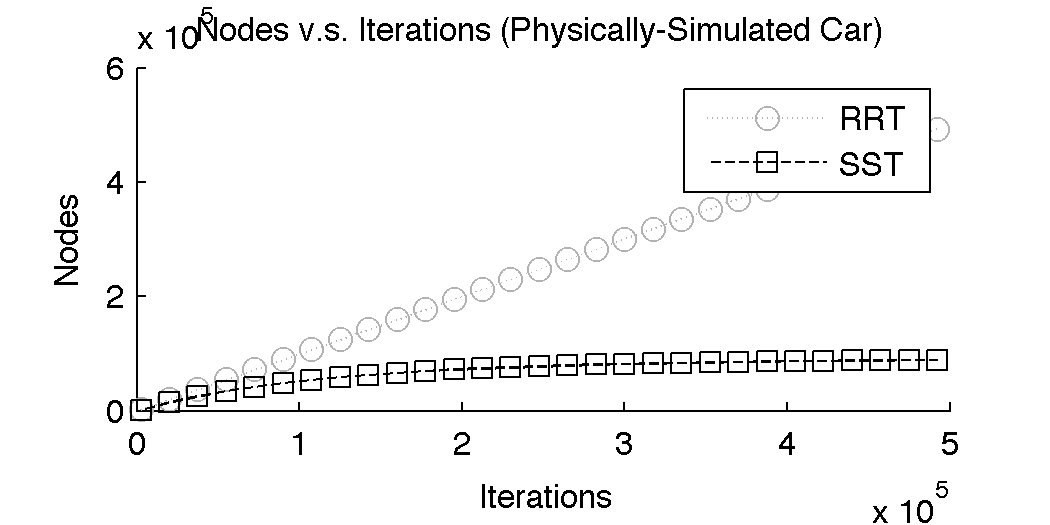}} \parbox{.33\textwidth}{\includegraphics[width=.33\textwidth]{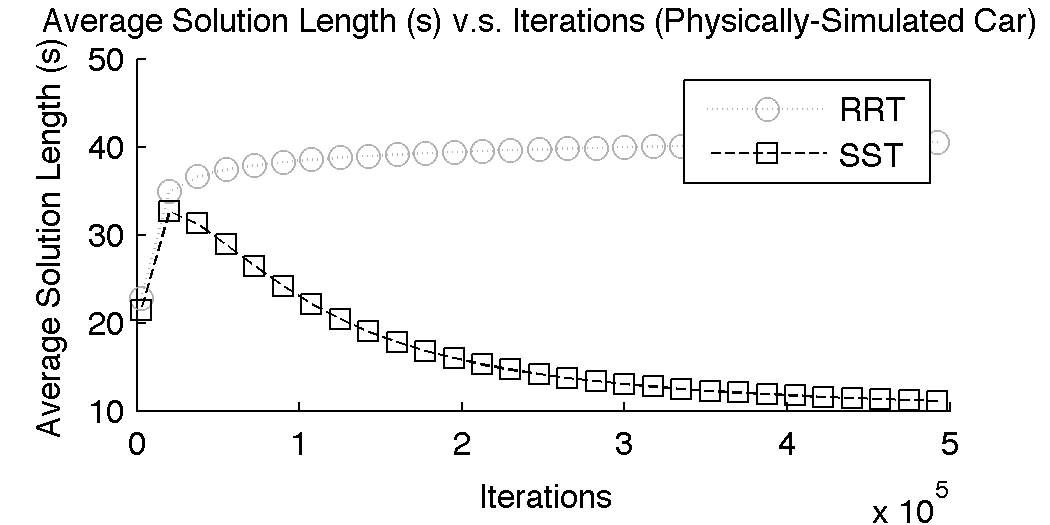}}
  \caption{Experimental results for the physically-simulated car-like
    system. The time complexity of the approach is similar to \rrt,
    but maintains a much smaller data structure.}
\label{fig:physsim}
\end{figure}

One of the more interesting applications of \sst\ is in the domain of
planning for physically-simulated systems
\citep{Advanced-Micro-Devices2012Bullet-Physics-}.  \sst\ is able to
provide improving path quality given enough time and keeps the number
of forward propagations to one per iteration as shown in
Figure \ref{fig:physsim}.  In this setup, the computational cost of
propagation overtakes the cost of nearest neighbor queries. Nearest
neighbor queries become the bottleneck in problems, such as the kinematic
point, where propagation and collision checking are cheap. In the
physically simulated case, however, these primitives are expensive,
therefore focusing the motion planner on good quality paths is
especially important. In this respect, \sst\ is suited to plan for
physically-simulated systems.

\edited{This physically-simulated car is modeled through the use of a
rectangular prism chassis, two wheel axles, and four wheels, creating
a system with 7 rigid bodies. These rigid bodies are linked together
with virtual joints in the Bullet physics
engine \citep{Advanced-Micro-Devices2012Bullet-Physics-}. The front
axle is permitted to rotate to simulate steering angle and thrust is
simulated as a force on the chassis. The data provided in
Figure \ref{fig:physsim} is generated by planning for the car in an
open environment and attempting to reach a goal state denoted by x,y
and heading.}

\edited{Using \sst\ for a physically simulated system raises the
question of whether this is a case where asymptotic optimality can be
argued formally. Note, that in this case, contacts arise between the
moving system and the plane. Such contacts typically violate the
assumptions specified in the problem setup and in this manner the
formal guarantees described in this work do not necessary
apply. Nevertheless, it is encouraging that the algorithm is still
exhibiting good performance, in terms of being able to improve the
quality of the solution computed over time. This is probably because
such real-world problems still exhibit a certain level of smoothness
that allows the algorithm to prune suboptimal solutions.  As described
in the Discussion section of this paper, future research efforts will
focus on generalizing the provided analysis and include interesting
challenges where contacts arise, including dexterous manipulation and
locomotion.}

\subsection{Graph-based Nearest Neighbor Structure}

\begin{figure}[h!]
\centering
  \parbox{.33\textwidth}{\includegraphics[width=.33\textwidth]{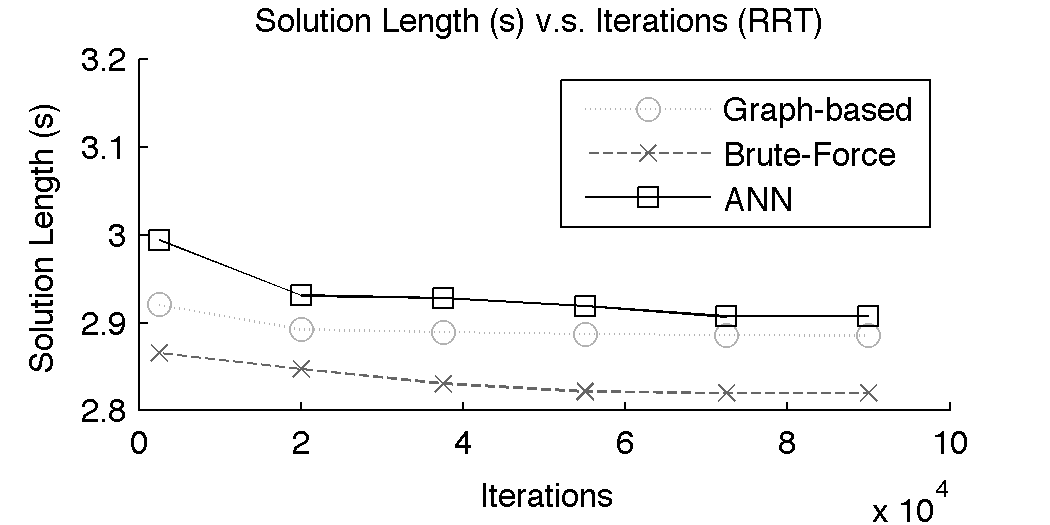}}
  \parbox{.33\textwidth}{\includegraphics[width=.33\textwidth]{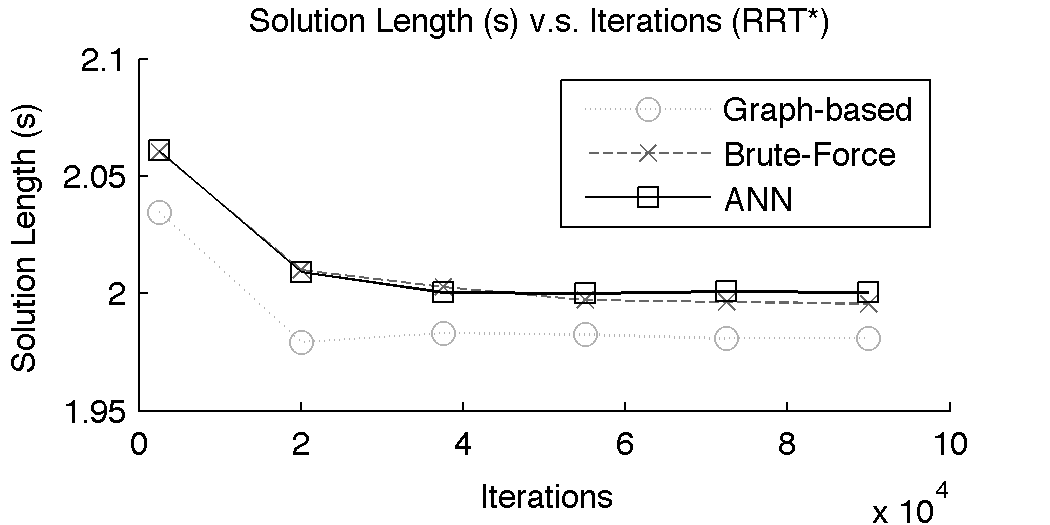}}
  \parbox{.33\textwidth}{\includegraphics[width=.33\textwidth]{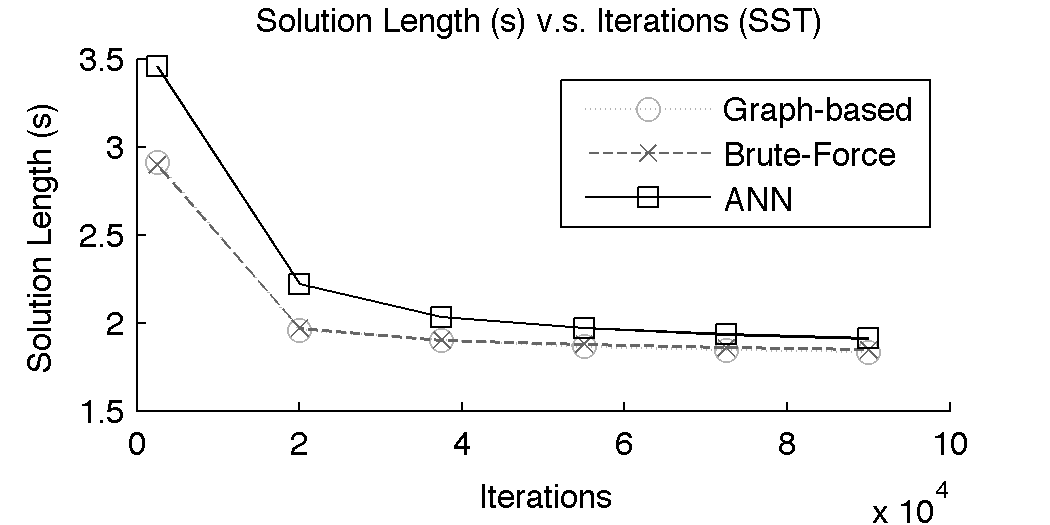}}
  \caption{A comparison of three different nearest neighbor structures
    in terms of solution quality at different iteration milestones for
    the point system. This is not considering the amount of time to
    reach these iteration milestones.}
\label{fig:metricquality}
\end{figure}

\begin{figure}[h]
\centering
  \parbox{.48\textwidth}{\includegraphics[width=.48\textwidth]{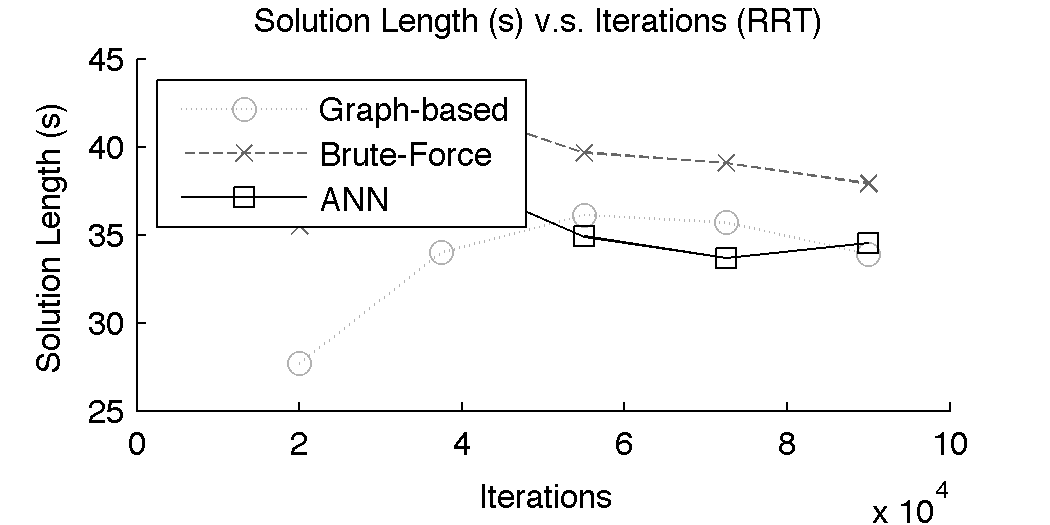}}
  \parbox{.48\textwidth}{\includegraphics[width=.48\textwidth]{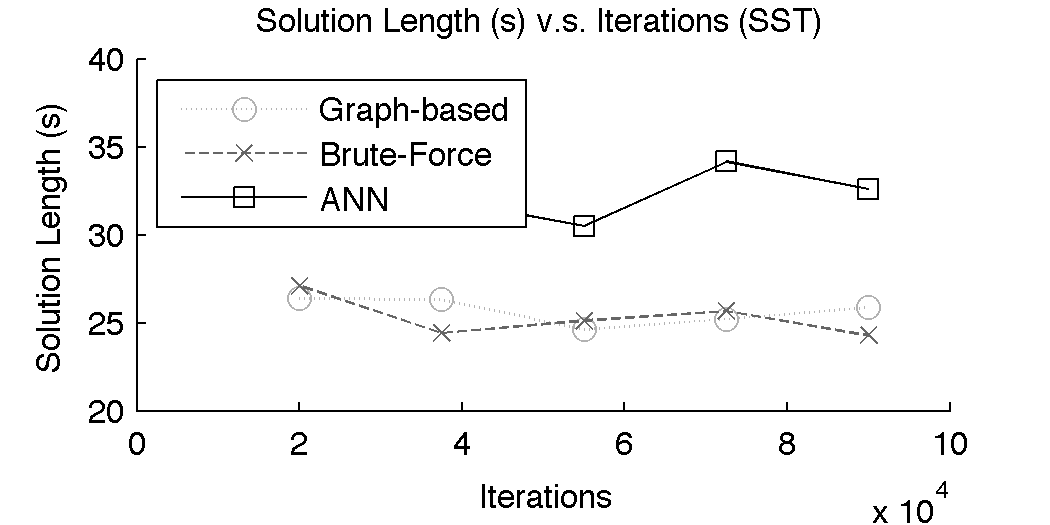}}
  \caption{A comparison of three different nearest neighbor structures
    in terms of solution quality at different iteration milestones for
    the airplane system. This is not considering the amount of time to
    reach these iteration milestones.}
\label{fig:metricquality2}
\end{figure}

In order to evaluate the graph-based nearest neighbor structure,
comparisons to two other alternatives are shown. First, a baseline
comparison with a brute force search is provided. This provides the
worst-case performance computationally that more intelligent search
methods should be able to overcome. Next, an approximate nearest
neighbor structure is used \citep{Arya1998An-Optimal-Algo}. This
approach follows the popular kd-trees approach to space decomposition
and nearest neighbor queries.  In the following experiments, the same
environments for the kinematic point and the airplane systems are
used, and comparisons are made between \rrt, \rrtstar, and \sst.

\begin{figure}[h!]
\centering
  \parbox{.33\textwidth}{\includegraphics[width=.33\textwidth]{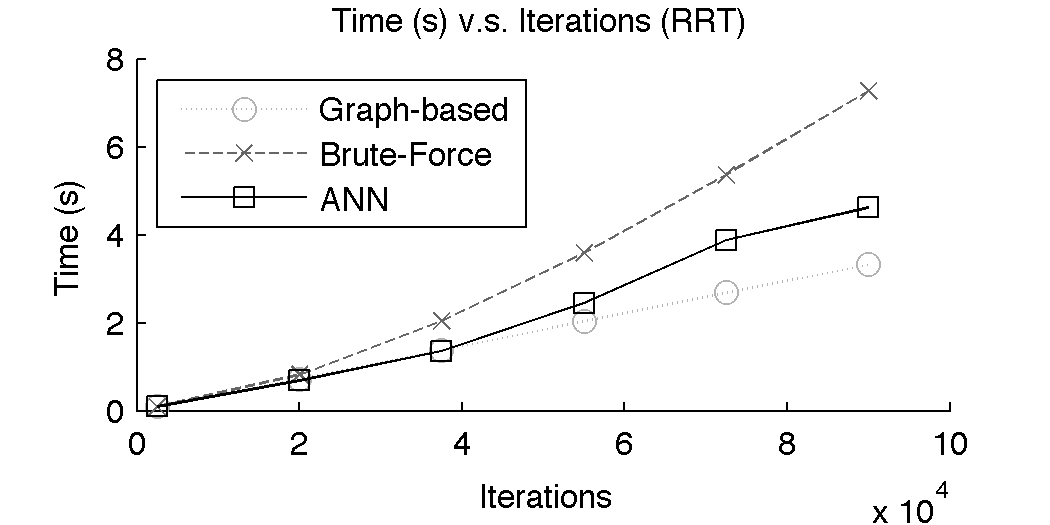}}
  \parbox{.33\textwidth}{\includegraphics[width=.33\textwidth]{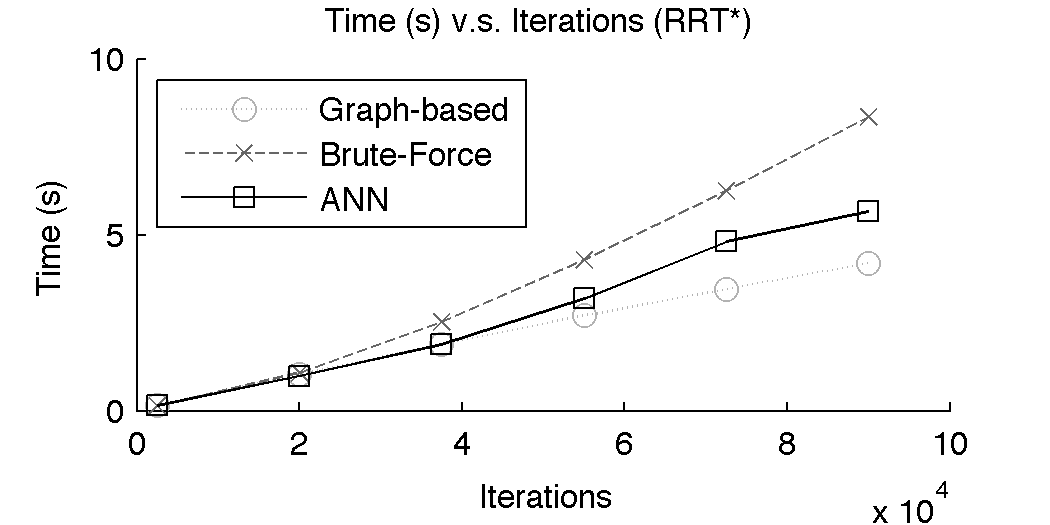}}
  \parbox{.33\textwidth}{\includegraphics[width=.33\textwidth]{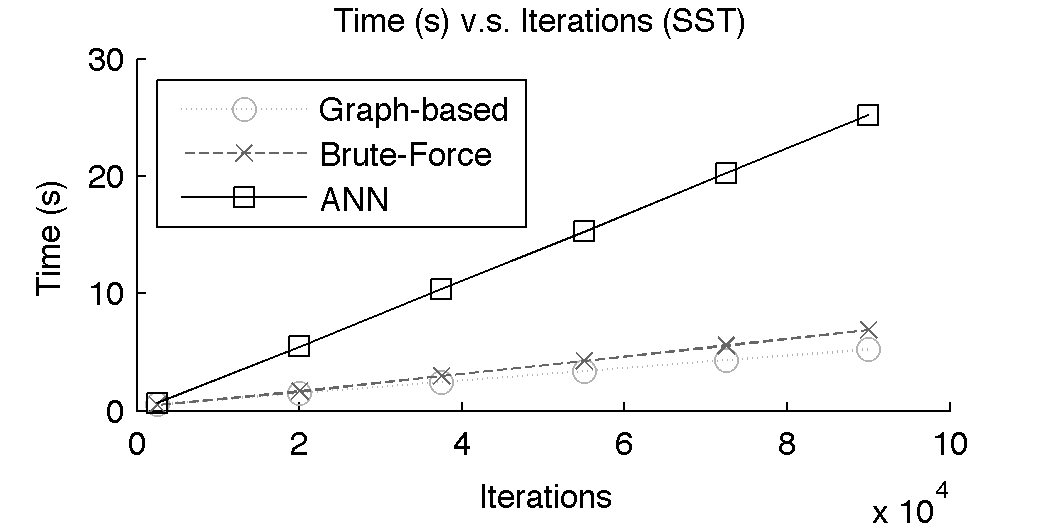}}
  \caption{A comparison of three different nearest neighbor structures
    in terms of time of execution at different iteration milestones
    for the point system.}
\label{fig:metrictime}
\end{figure}


A comparison of the resulting solution quality between planners that
use different nearest neighbor structures is shown in Figures
\ref{fig:metricquality} and \ref{fig:metricquality2}.  In the case of
\rrt, where the Voronoi bias heavily affects the expansion process,
having an exact brute force metric actually provides small benefits in
terms of quality. For \rrtstar\ and \sst, small approximation errors
when returning nearest neighbors can actually result in generating
longer edges that help in path quality. This causes a small
improvement in path quality for these algorithms.

\begin{table*}[h!]
\begin{center}
\begin{tabular}{|p{.1\textwidth}||p{.1\textwidth}|p{.1\textwidth}|p{.1\textwidth}|}
\hline
Structure & Single Query & Range Query & $k$-Query \\
\hline
Brute-Force & 100\% & 100\% & 100\% \\
\hline
ANN & 100\% & 93.48\% & 99.74\% \\
\hline
Graph-based & 100\% & 99.96\% & 100\% \\
\hline
\end{tabular}
\end{center}
\caption{The accuracy of the graph-based nearest neighbor
  structure. These results state that over 5000 queries to a data
  structure holding 50000 states, these are the percentages of queries
  that were returned with the correct result. Most errors occurred
  from not returning all relevant results \edited{(states that should
    have been returned)} or returning false positives
  \edited{(states that should not have been returned)}.}
\label{tab:accuracy}
\end{table*}


In Figure \ref{fig:metrictime}, timing data for each of the nearest
neighbor structures is shown. As expected in \rrt\ and \rrtstar, the
brute force method is worse than either of the approximate
structures. The graph-based structure slightly outperforms the
alternative method. An interesting effect \edited{occurs} in the case of
\sst\ however. Since \sst\ maintains a small number of nodes for this
problem instance, the brute force search can actually be competitive
with the graph-based nearest neighbor. The alternative method that
\edited{does not} explicitly handle removal is much slower than the graph
structure for \sst, mainly due to having to rebuild its internal
structure when too many nodes are removed.  


Table \ref{tab:accuracy} shows the accuracy of the graph-based method
compared to the other methods. While resulting in some query errors,
the number of errors is less than the comparison method.

\section{Discussion and Conclusion}
\label{sec:discussion}
Recently, the focus in sampling-based motion planning has moved to
providing optimality guarantees, while balancing the computational
efficiency of the related methods. Achieving this objective for
systems with dynamics has generally required the generation of
specialized steering functions. This work shows that a fully-random
selection/propagation procedure can achieve asymptotic optimality
under reasonable assumptions for kinodynamic systems. The same method,
however, has a very slow convergence rate to finding high-quality
solutions, which indicates that the focus should primarily be on the
convergence rate of methods that provide path improvement over
time. 

To address these issues, this work proposed a new framework for
asymptotically optimal sampling-based motion planning. The departure
from previous work is the utilization of best-first selection strategy
and a pruning process, which allow for fast convergence to
high-quality solutions and a sparse data structure. Experiments and
analytical results show the running time and space requirements of a
concrete implementation of this framework, i.e., the \sst\ approach,
are better even than that of the efficient but suboptimal \rrt,
while \sst\ can still improve path quality over time. This performance
increase is seen in many different scenarios, including in the case of
a physically-simulated system.

{\bf Parameter Selection: } The two parameters of \sst, namely
$\Ddrain$ and $\Dnear$, directly affect the performance of the
algorithm. Since the $\Ddrain$ radius controls how much pruning \sst\
will perform, it is necessary that this parameter is not set too
high because it can lead the algorithm not to discover paths through
narrow passages. Practically, $\Ddrain$ can be as large as the
clearance of paths desired from a given problem instance.  It is also
helpful to choose this value to be smaller than the radius of the goal
region, so as to allow the generation of a sample close to the goal.

The parameter $\Dnear$ should be larger than $\Ddrain$ to allow the
tree data structure to properly expand.  A value for $\Dnear$ that is
too large will result in poor exploration of the state space since
nodes closer to the root will be selected repetitively. Overall, a
balance between the state space size, $\Dnear$, and $\Ddrain$ must be
maintained to achieve good performance. The \sststar\ approach allows
to start the search using rather arbitrary large values for $\Ddrain$
and $\Dnear$, which then automatically decrease over time.

{\bf Finite-time Properties:} Since \sst\ maintains a relatively small
data structure, and in bounded spaces it results in a finite size data
structure, it is interesting to consider the finite-time properties
that can be argued \citep{Dobson2013Finite-Time-Nea}. This depends
significantly on the rate at which the witness set $S$ can cover the
free space. After this initial coverage, it may be possible to examine
the quality of the existing paths. 

{\bf Planning under Uncertainty:} By removing the requirement of the
steering function, \sst\ can be applied to other problems where
steering functions are difficult to construct.  One of these areas is
planning under uncertainty, where planning is performed in belief
space. It is difficult to compute a steering function that connects
two probability distributions in this domain, but forward propagation
can update the corresponding beliefs. Some challenges in
applying \sst\ to this domain involve computing appropriate distance
metrics for the best first and pruning operations, as well as the
increased dimensionality of the problem. \edited{Some progress has
been recently achieved in this direction, where it has been shown that
in the context of the methods described in the current paper a
suitable function based on the Earth Mover's distance can lead to
efficient solutions when planning under
uncertainty \citep{Littlefield2015The-Importance-}. This can lead
eventually to the application of such solutions to important problems
that involve significant uncertainty, such as kinodynamic and
non-prehensile manipulation (e.g., pushing, throwing, pulling, etc).}

\edited{{\bf Feedback-based Motion Planning:} Another extension relates to
feedback-based motion planning and the capability to argue that the
computed trajectories are dynamically stable. The current work follows
the majority of the literature in sampling-based kinodynamic planning
and is providing only nominal trajectories and not feedback-based
plans or policies. There has been work that takes advantage of
sampling in the context of feedback-based motion planning, such as the
work on LQR-trees \citep{Tedrake2009LQR-trees:-Feed}. Nevertheless, it
has been typically difficult to argue about the optimality of a
feedback-based solution when it comes to realistic and relatively
high-dimensional dynamical robotic systems. In this way, an
interesting research direction is to identify the conditions under
which it will be possible to provide such guarantees in the context of
feedback-based planning.}

\edited{{\bf Real-world Experiments and Applications:}} It is also
important to evaluate the effectiveness of the approach on real
systems with significant dynamics, especially aerial systems that
perform aggressive maneuvers and systems modeled through the use of
physics engines. For example, future planetary exploration missions
may involve more capable rovers. They will have the capability to move
at higher speeds in low gravity environments, potentially acquiring
ballistic trajectories for small periods of time. Thus, reasoning
about the dynamics becomes more important during the planning
process. \sst\ may be useful in this domain to optimize paths with
respect to path length, energy expenditure, or the sensitivity of the
sensor payload on-board.

\edited{{\bf Locomotion and Dexterous Manipulation:} Other potential
research domains where \sst\ may be used include locomotion and
dexterous manipulation. These challenges involve planning using models
of contact between objects and physical considerations, such as
balancing of a locomotion system or stability of a grasp for a
manipulator. The use of a physics engine to model friction and mass
effects can be useful here. As demonstrated above, \sst\ provides
control sequences that improve over time when a physics engine is
used. Nevertheless, the presence of contacts introduces important
complexities that are not currently handled by the presented
analysis.}

\edited{In particular, there are two critical assumptions which
complicate the generalization of the provided results: (a) the system
dynamics are expressed in the form of equation \ref{eq:dynamics},
which is a nonlinear ordinary differential equation, and (b) the
manifolds in which the systems live are smooth subsets of a
$d$-dimensional Euclidean space. These assumptions do not allow to
consider models of rigid body dynamics and stick-slip friction, which
are useful idealizations of locomotion and dexterous
manipulation. Such systems exhibit jump-discontinuities and in general
cannot be represented by expressions of the form in
Equation \ref{eq:dynamics}. There is also a question of whether it is
possible to address challenges in spaces, which are not locally
Euclidean.}

\edited{It would be interesting to study manifolds generated by contact
  constraints. Such manifolds can be algebraic varieties, which need
  not be smooth. Furthermore, such manifolds can be of different
  dimensions, as finger gaiting and locomotion problems really don't
  live on varieties of a single dimension, but live on stratified sets
  in a higher-dimensional ambient state space. These issues motivate
  further research in the direction of providing general
  sampling-based algorithms that exhibit asymptotic optimality
  guarantees for proper models of dexterous manipulation and
  locomotion systems.}

\vspace{-.1in}

\section*{Acknowledgments}
The authors would like to thank the anonymous reviewers of the earlier
versions of this work that appeared in IROS 2013
\citep{Littlefield2013Efficient-sampl} and WAFR 2014
\citep{Li2014Sparse-Methods-}, \edited{as well as the reviewers of the
  IJRR version} for their thoughtful comments. This work has been
supported by NSF awards (IIS-1451737, CCF-1330789) and a NASA Space \&
Technology Research Fellowship to Zakary Littlefield (NNX13AL71H).

\bibliographystyle{agsm}
\bibliography{dynamics}

\section*{Appendix A}

\noindent

This appendix provides an informal proof of Lemma
\ref{lem:closeexist}, which argues that given Assumption
\ref{ass:system}, there exist $\delta$-similar trajectories for any
specific trajectory $\pi$. The existence of $\delta$-similar
trajectories relates to notions that have been used in previous
analysis efforts, such as the \emph{linking sequence in
$\epsilon$-good free spaces} \citep{Hsu2002}, or
\emph{attraction sequences} \citep{LaValle2001}, or
being \emph{homotopic in $\delta$-interior of
$X_{free}$} \citep{Karaman2011Sampling-based-,
Karaman2013Sampling-Based-}.

{\bf Lemma \ref{lem:closeexist}.}\emph{
 Let there be a trajectory $\pi$ for a system satisfying
 Eq. \ref{eq:dynamics} and \emph{Chow's condition}. Then there exists
 a positive value $\delta_0$ called the dynamic clearance, such that: $\forall\ \delta \in
 (0,\delta_0]$, $\forall\ x_{0}^\prime\in\ball_\delta(\pi(0))$, and
 $\forall\ x_{1}^\prime\in\ball_\delta(\pi(t_{\pi}))$, there exists a
 trajectory $\pi^\prime$, so that: (i) $\pi^\prime(0)=x_{0}^\prime$
 and $\pi^\prime(t_{\pi^\prime})=x_1^\prime$; (ii) $\pi$ and $\pi^{\prime}$
 are $\delta$-similar trajectories.
}

{\bf Proof Sketch:}
Informally speaking, \emph{Chow's} condition implies that the \emph{Ball
  Box} theorem holds. It also implies that the manifold $\Xfree$ is
\emph{regular} and \emph{involutory}
\citep{Choset2005Principles-of-R}.  A real-analytic control-affine
system is small-time locally accessible (\emph{STLA}), if and only if
the \emph{distribution} satisfies \emph{Chow's} condition
\citep{Sussmann-LocalControllability1987}.  Assume every state on the
optimal trajectory is a \emph{regular point}. Then, the
\emph{sub-Riemannian ball} up to a small constant radius $t_\epsilon$
contains a weighted box of the same dimension of the state space and
it is oriented according to vector fields of the \emph{Lie
  brackets}. The bases are real analytical. Therefore there exists an
open neighborhood at each point $x$ such that the bases evaluated at a
different point $x^\prime$ converge to the bases at $x$ as $x^\prime$
approaches $x$.  Then, the weighted boxes centered by two sufficiently
close states have a non-empty intersection. It implies that a hyper
ball of some positive radius $\delta_0$ can be fitted into this
intersection region. Overall, there are two sufficiently close
hyper-ball regions on the optimal trajectory such that between any
point $x$ in one ball and any point in the other ball there exists a
horizontal curve and the length of the curve is less or equal to the
radius $t_\epsilon$ of the \emph{sub-Riemannian ball}. Then
concatenating all hyper balls along a specified trajectory, results in
the generation of $\delta$-similar trajectories. \telos

\section*{Appendix B}
\noindent
\edited{This appendix proves Theorem \ref{thm:boundedend}, which shows that sampling piece-wise constant controls can generate trajectories that are $\delta$-similar to one another. If one trajectory in question is an optimal one, then a trajectory that is $\delta$-similar to that optimal trajectory can be generated.  }

{\bf Theorem \ref{thm:boundedend}. }\emph{
 For two trajectories $\pi$ and $\pi^\prime$ such that
 $\pi(0)=\pi^\prime(0)=x_0$ and $\Delta u=\sup_t(||u(t) - u^\prime(t)||)$:
 $$||\pi^\prime(T) - \pi(T)|| < K_u\cdot T\cdot e^{K_x\cdot T}\cdot \Delta
 u,$$ for any period $T\geq 0$.}

{\bf Proof:}
Given Assumption \ref{ass:system}, for any two states $x_0,x_1$ and two controls $u_0,u_1$: 
\begin{align*}
||f(x_0, u_0) -  f(x_0, u_1) || \leq K_u || u_0 - u_1||\ \ \ \ \
  ||f(x_0, u_1) - f(x_1, u_1) || \leq K_x ||x_0 - x_1||.
\end{align*}
By summing these two inequalities:
\begin{equation}
\begin{aligned}
|| f(x_0, u_0) -  f(x_0, u_1) || + || f(x_0, u_1) - f(x_1, u_1)|| 
   \leq K_u ||u_0 - u_1 || + K_x ||x_0 -  x_1||.
\end{aligned}
\label{bound4}
\end{equation}
Given the Euclidean distance, the following inequality is true:
\begin{equation*}
\begin{aligned}
||f(x_0 u_0) - f(x_1, u_1)|| \leq || f(x_0, u_0) - f(x_0, u_1) || + ||
  f(x_0, u_1) - f(x_1, u_1) ||.\\
\end{aligned}
\end{equation*}
By joining this with (\ref{bound4}): 
\begin{equation}
\begin{aligned}
||f(x_0, u_0) - f(x_1, u_1) || \leq K_u ||u_0 - u_1|| + K_x ||x_0 - x_1||.
\end{aligned}
\label{bound6}
\end{equation}
Now, divide $[0, T]$ into $n$ segments with equal length $\Delta t$. Approximating the value of a trajectory $\pi(T)$ using Euler's Method, 
there is a sequence of states $\{x_0, x_1, ..., x_n\}$. Let $u_i$ denote $u(i\Delta t)$ corresponding to the control applied at each state.
\vspace{-.1in}
\begin{align*}
x_i = f(x_{i-1}, u_{i-1})\Delta t + x_{(i-1)}.
\end{align*}
For two trajectories $\pi$ and $\pi^\prime$ such that $\pi(0) = \pi^\prime(0) = x_0$, $u(t)$ and $u^\prime(t)$ are the corresponding control functions. Then:
\begin{align*}
x_n = x_{n-1} + f(x_{n-1}, u_{n-1})\Delta t \\
x_n^\prime = x_{n-1}^\prime + f(x_{n-1}^\prime, u_{n-1}^\prime)\Delta t.
\end{align*}
Then:
\begin{equation}
\begin{aligned}
|x_n - x_n^\prime || \leq ||x_{n-1} - x_{n-1}^\prime|| + || f(x_{n-1},
 u_{n-1}) - f(x_{n-1}^\prime, u_{n-1}^\prime) || \Delta t .
\label{bound9}
\end{aligned}
\end{equation}
Using (\ref{bound6}) and (\ref{bound9}):
\begin{align*}
||x_n - x_n^\prime|| \leq ||x_{n-1} - x_{n-1}^\prime|| + (K_u||u_{n-1} - u_{n-1}^\prime|| + K_x || x_{n-1} - x_{n-1}^\prime||)\Delta t ,
\end{align*}
\vspace{-.2in}
\begin{equation}
\begin{aligned}
||x_n - x_n^\prime|| \leq K_u\Delta t || u_{n-1} - u_{n-1}^\prime|| + (1 + K_x\Delta t)||x_{n-1} s - x_{n-1}^\prime||.
\label{bound10}
\end{aligned}
\end{equation}

Reusing (\ref{bound10}) to expand $||x_{n-1} - x_{n-1}^\prime||$:
\vspace{-.1in}
\begin{align*}
||x_n - x_n^\prime|| \leq K_u\Delta t||u_{n-1} - u_{n-1}^\prime|| + 
(1 + K_x\Delta t)(K_u\Delta t || u_{n-2} - u_{n-2}^\prime|| + (1 + K_x\Delta t)||x_{n-2} - x_{n-2}^\prime||).
\end{align*}
\vspace{-.3in}
\begin{align*}
||x_n - x_n^\prime|| \leq K_u\Delta t||u_{n-1} - u_{n-1}^\prime|| 
+ (1 + K_x\Delta t)K_u\Delta t || u_{n-2} - u_{n-2}^\prime|| + (1 + K_x\Delta t)^2||x_{n-2} - x_{n-2}^\prime||).
\end{align*}
By repeatedly expanding the right side:
\begin{align*}
||x_n - x_n^\prime|| \leq (1 + K_x\Delta t)^n||x_{0} - x_{0}^\prime|| +
K_u\Delta t || u_{n-1} - u_{n-1}^\prime|| +
(1 + K_x\Delta t)K_u\Delta t|| u_{n-2} - u_{n-2}^\prime|| +\\
\cdots +
(1 + K_x\Delta t)^{n-1}K_u\Delta t || u_{0} - u_{0}^\prime||.
\end{align*}
\vspace{-.3in}
Since $x_{0} = x_{0}^\prime = x_0$, and $\Delta u = max_{i=0}^{n-1}(|| u_{i} - u_{i}^\prime||$.:
\begin{align*}
||x_{n} - x_{n}^\prime|| \leq K_u\Delta t\sum\limits_{i=0}^{n-1}(1+K_x\Delta t)^{i}\Delta u.
\end{align*}
\vspace{-.2in}
Since $n\Delta t=T$:
\begin{align*}
||x_{n} - x_{n}^\prime|| \leq K_uT\frac{1}{n}\sum\limits_{i=0}^{n-1}(1+\frac{K_xT}{n})^{i}\Delta u
\end{align*}
Due to the fact that $1<(1+\frac{\alpha}{n})^i<e^\alpha$, where $1\leq i\leq n$ and $\alpha>0$:
\begin{align*}
||x_{n} - x_{n}^\prime||<K_uTn\frac{1}{n}e^{K_xT}\Delta u \Rightarrow ||x_{n} - x_{n}^\prime||<K_uTe^{K_xT}\Delta u.
\end{align*}
Given Assumption \ref{ass:system}, Euler's method converges to the solution of the {\tt Initial Value Problem}. Then:
\begin{align*}
||\pi(T) - \pi^\prime(T)|| = \lim_{n\to\infty}||x_{n} - x_{n}^\prime||\text{, where } n\Delta t=T.
\end{align*}
Therefore:$||\pi(T) - \pi^\prime(T)|| <K_uTe^{K_xT}\Delta u$\telos

\section*{Appendix C}
\noindent
\edited{This appendix proves Theorem \ref{thm:drtsimilar}. The proof shows that given an optimal trajectory, it is possible for \rt\ to generate a trajectory that is close to the optimal. In effect, this shows that \rt\ is probabilistically complete. Appendix D will examine the path cost of this generated trajectory. }

{\bf Theorem \ref{thm:drtsimilar}.}\emph{
\rt\ will eventually generate a $\delta$-similar trajectory to any optimal trajectory for any robust clearance $\delta > 0$.
}

{\bf Proof:} Let $\pi^\ast$ denote an optimal trajectory of cost
$C^\ast$ for a $\delta$-robustly feasible motion planning problem
($\Xfree$, $\xinit$, $\Xgoal$, $\delta$) and consider the covering
ball sequence $\balls$($\pi^\ast(t)$, $\delta$, $C_\Delta$) over the optimal trajectory. 
Recall that from Def. \ref{def:balls} and Assumption \ref{ass:cost}, for a given value $T_{prop}$, 
which is one parameter of \mcprop, it is always possible to find a value $C_\Delta$ such that 
a ball sequence can be defined. 
Consider the event
$\neg E_k^{(n)}$ that the algorithm fails to generate any
near-optimal trajectory inside a $\delta$-ball centered at the $k^{th}$ segment of $\pi^\ast$, $\ball_{\delta}(x_k^\ast)$,
after $n$ iterations, which only happens when all the $n$ consecutive iterations fail,
i.e.:
\vspace{-.1in}
\begin{align}
\neg E_k^{(n)} &= \neg A_k^{(1)} \cap \neg A_k^{(2)} \cap \text{ ... } \cap \neg A_k^{(n)}\notag\\
\Prob(\neg E_k^{(n)}) &= \Prob(\neg A_k^{(1)})\cdot \Prob(\neg A_k^{(2)}|\neg A_k^{(1)})
\cdot \text{...} \cdot \Prob(\neg A_k^{(n)}|\bigcap_{j=1}^{n-1}\neg A_k^{(j)}).
\label{eq:Pconcat}
\end{align}

\vspace{-.in}
The probability that $\neg A_k^{(n)}$ happens given
$\bigcap_{j=1}^{n-1} \neg A_k^{(j)}$ is equivalent to the probability
of failing to generate a trajectory to
$\ball_{\delta}(x_{k-1}^\ast)$ plus the probability that a
trajectory has been generated to $\ball_{\delta}(x_{k-1}^\ast)$, but
the algorithm has to generate a new trajectory to $\ball_{\delta}(x_{k}^\ast)$,
i.e.:

\vspace{-.25in}
\begin{align}
\Prob(\neg A_k^{(n)}|\bigcap_{j=1}^{n-1}\neg A_k^{(j)})
&= \Prob(\neg E_{k-1}^{(n)}) + \Prob(E_{k-1}^{(n)})\cdot \Prob(\{\text{fail stepping to } \ball_{\delta}(x_k^\ast) \}) \notag\\
&\leq \Prob(\neg E_{k-1}^{(n)}) + \Prob(E_{k-1}^{(n)})(1-\frac{\rho_{\delta\to\delta}}{n})\notag\\
&= 1 - \Prob(E_{k-1}^{(n)})\cdot \frac{\rho_{\delta\to\delta}}{n}.
\label{eq:PAi}
\end{align}

\noindent
Therefore, using \edited{Equation} \ref{eq:Pconcat} and \edited{Equation} \ref{eq:PAi}:
\vspace{-.1in}
\begin{align}
\Prob(E_k^{(n)}) &\geq 1 - \prod_{j=1}^n (1 - \Prob(E_{k-1}^{(j)})\cdot \frac{\rho_{\delta\to\delta}}{j}).
	\label{eq:Epropagate}
\end{align}
\vspace{-.1in}

\noindent
{\bf For the base case}, $\Prob(E_0^{(j)}) = 1$ because $\xinit$ is always in $\ball_{\delta}(x_0)$. Then,
consider event $E_{1}$ from iteration 1 to $n$ using \edited{Equation} (\ref{eq:Epropagate}), and set $y^{(n)}_1=\prod_{j=1}^n(1-\frac{\rho_{\delta\to\delta}}{j})$:
\vspace{-.06in}
\begin{align*}
\Prob(E_1^{(n)}) \geq 1-\prod_{j=1}^n(1-\frac{\rho_{\delta\to\delta}}{j}) = 1-y^{(n)}_1 .
\end{align*}

\noindent
The logarithm of $y^{(n)}_1$ behaves as follows:
\begin{align}
\log y^{(n)}_1 = \log \prod_{j=1}^n(1-\frac{\rho_{\delta\to\delta}}{j}) 
	= \sum_{j=1}^n \log (1-\frac{\rho_{\delta\to\delta}}{j})
	< \sum_{j=1}^n -\frac{\rho_{\delta\to\delta}}{j} = -\rho_{\delta\to\delta}\cdot\sum_{j=1}^n \frac{1}{j}.\label{eq:baseCaseRT}
\end{align}

\noindent
Clearly, \edited{Equation} \ref{eq:baseCaseRT} diverges as $n\to\infty$:
\vspace{-0.1in}
\begin{align}
\lim_{n\to\infty}\log y^{(n)}_1 &< \lim_{n\to\infty} -\rho_{\delta\to\delta}\cdot\sum_{j=1}^n \frac{1}{j} = -\infty
\iff \lim_{n\to\infty} y^{(n)}_1 = 0. \notag \\
&\lim_{n\to\infty} \Prob(E_1^{(n)}) \geq = 1 - 0 = 1.\label{eq:baseCaseHold}
\end{align}
\vspace{-0.1in}

\noindent
{\bf Now consider the induction step}, if $\lim_{n\to\infty} \Prob(E_{k-1}^{(n)}) =1$, we need to show that the same will be true for $E_{k}^{(n)}$. Similarly, set $y_k^{(n)} = \prod_{j=1}^n (1 - \Prob(E_{k-1}^{(j)})\cdot \frac{\rho_{\delta\to\delta}}{j})$. The logarithm of $y_k^{(n)}$ behaves as follows:
\vspace{-0.05in}
\begin{align}
\log y_k^{(n)}	= \log \prod_{j=1}^n (1 - \Prob(E_{k-1}^{(j)})\cdot \frac{\rho_{\delta\to\delta}}{j})
= \sum_{j=1}^n \log (1 - \Prob(E_{k-1}^{(j)})\cdot \frac{\rho_{\delta\to\delta}}{j})
				< -\rho_{\delta\to\delta} \cdot \sum_{j=1}^n \frac{\Prob(E_{k-1}^{(j)})}{j}. \label{eq:induction}
\vspace{-0.05in}
\end{align}

Next, we want to show that for any constant $c_1\in (0, 1)$:
\vspace{-0.1in}
\begin{align*}
\sum_{j=1}^\infty\frac{\Prob(E^{(j)}_{k-1})}{j}>\sum_{j=1}^\infty\frac{c_1}{j}.
\end{align*}

\noindent
To show the above expression holds, let $c_2$ be another constant such that $c_1<c_2<1$. Clearly, \edited{Equation} \ref{eq:Epropagate} indicates that $E^{(n)}_k$ monotonically increases when $\rho_{\delta\to\delta}>0$ and $\Prob(E^{(j)}_{k-1})>0$. From the induction assumption, $\lim_{n\to\infty} \Prob(E_{k-1}^{(n)}) =1$, then there exist corresponding numbers $j_1<j_2$ such that $c_1\geq \Prob(E_{k-1}^{(j_1-1)}), c_1 < \Prob(E_{k-1}^{(j_1)})$ and $c_2\geq \Prob(E_{k-1}^{(j_2-1)}), c_2 < \Prob(E_{k-1}^{(j_2)})$.

\noindent
Now examine the following summation from $j_1$ to $\infty$, according to the definition of summation:
\begin{align}
 &\sum_{j=j_1}^\infty\frac{\Prob(E^{(j)}_{k-1})-c_1}{j}\label{eq:PEminus}\\
=& \sum_{j=j_1}^{j_2-1}\frac{\Prob(E^{(j)}_{k-1})-c_1}{j} + \sum_{j=j_2}^{\infty}\frac{\Prob(E^{(j)}_{k-1})-c_1}{j}\notag\\
>& \sum_{j=j_1}^{j_2-1}\frac{\Prob(E^{(j)}_{k-1})-c_1}{j} + \sum_{j=j_2}^{\infty}\frac{c_2-c_1}{j} = \infty.\label{eq:PEtwopart}
\end{align}

\noindent
Clearly, the first term in \edited{Equation} \ref{eq:PEtwopart} is positive. The second term in \ref{eq:PEtwopart} diverges to infinity. Then \ref{eq:PEminus} is positive and unbounded.
Consider the following summation from $1$ to $j_1-1$:
\begin{align}
-\infty < \sum_{j=1}^{j_1-1}\frac{\Prob(E^{(j)}_{k-1})-c_1}{j} < 0.\label{eq:PEj1}
\end{align}
Clearly, \edited{Equation} \ref{eq:PEj1} is negative but bounded, since there are only finite terms.

\noindent
Then combining \edited{Equation} \ref{eq:PEminus} and \edited{Equation} \ref{eq:PEj1},
\begin{align}
\sum_{j=1}^\infty\frac{\Prob(E^{(j)}_{k-1})-c_1}{j}>0
\iff \sum_{j=1}^\infty\frac{\Prob(E^{(j)}_{k-1})}{j}>\sum_{j=1}^\infty\frac{c_1}{j}.\label{eq:PEj}
\end{align}

\noindent
Combining \edited{Equation} \ref{eq:induction} and \edited{Equation} \ref{eq:PEj},
\vspace{-0.1in}
\begin{align}
\lim_{n\to\infty}\log y_k^{(n)} < -\rho_{\delta\to\delta}\cdot \lim_{n\to\infty}\sum_{j=1}^n \frac{ \Prob(E^{(j)}_{k-1}) }{j}
				< -\rho_{\delta\to\delta}\cdot \lim_{n\to\infty}\sum_{j=1}^n \frac{ c_1 }{j}=-\infty
\iff \lim_{n\to\infty} y_k^{(n)} = 0.\notag
\vspace{-0.05in}
\end{align}

\noindent
Then, the induction step holds, i.e.:
\vspace{-0.1in}
\begin{align}
\lim_{n\to\infty} \Prob(E_{k}^{(n)}) = 1 - 0 = 1\text{, if }\lim_{n\to\infty} \Prob(E_{k-1}^{(n)}) = 1.\label{eq:inductionhold}
\vspace{-0.05in}
\end{align}

\noindent
Both of the base case \ref{eq:baseCaseHold} and induction step \ref{eq:inductionhold} hold. Therefore, it is true that:
\begin{align}
\lim_{n\to\infty} \Prob(E_{k}^{(n)}) = 1
\implies\liminf_{n\to\infty} \Prob(E_k^{(n)})=1.\label{eq:complete}
\end{align}

\noindent
Therefore, \rt\ will eventually generate a $\delta$-similar trajectory to any optimal trajectory for any robust clearance $0<\delta^\prime\leq \delta$.

\telos

\section*{Appendix D}
{\bf Theorem \ref{thm:drtoptimal}.}\emph{
\rt\ is {\it asymptotically optimal}.
}

{\bf Proof:} \edited{Theorem} \ref{thm:drtsimilar} indicates that a
$\delta$-similar trajectory to an optimal trajectory $\pi^\ast$ with
cost of $C^\ast$ \emph{almost surely} exists and is discovered by
\rt. According to the definition of $\delta$\emph{-similar}
trajectories and assumption \ref{ass:cost}: $|{\tt cost}(\pi) - {\tt
  cost}(\pi^\ast)|\leq K_c\cdot\delta$.  Then: ${\tt
  cost}(\pi)\leq{\tt cost}(\pi^\ast)+K_c\cdot\delta$. Therefore, event
$E_k^{(\infty)}$ implies event $\{Y^{RT}_\infty\leq {\tt
  cost}(\pi^\ast)+k\cdot K_c\cdot\delta\}$, where
$k=\frac{C^\ast}{C_\Delta}$. In other words, \edited{Theorem}
\ref{thm:drtsimilar} implies that:
\begin{equation}
\Prob(\{\limsup_{n\to\infty}Y^{RT}_n\leq (1+\frac{K_c\cdot\delta}{C_\Delta})\cdot {\tt cost}(\pi^\ast)\})=1.
\end{equation}

Therefore, the \rt\ is \emph{asymptotically $\delta$-robust
  near-optimal} for the given $\delta$. In fact, however, due to \edited{Theorems}
\ref{thm:mccomplete} and \ref{thm:drtsimilar} the above holds true for
any $\delta>0$. Note that $C_\Delta$ is the step cost of the optimal
trajectory segment between $x_i^\ast$ and $x_{i+1}^\ast$. And, most
importantly, $C_\Delta$ is determined by the ball sequence, which
means $C_\Delta$ does not shrink when $\delta$ decreases.  Then, as
$\delta \rightarrow 0$:
$\Prob(\{\limsup_{n\to\infty}Y^{RT}_n\leq {\tt
  cost}(\pi^\ast(t))\})=1$\telos

\section*{Appendix E}
{\bf Theorem \ref{thm:convergenRateRTlog} }\emph{
In the worst case, the $k^{th}$ segment of the trajectory returned by \rt\ converges logarithmically to the near optimal solution, i.e.:
$\lim_{n\to\infty}\frac{|\Prob(E_{k}^{(n+2)}) - \Prob(E_{k}^{(n+1)})|}{|\Prob(E_k^{(n+1)}) - \Prob(E_k^{(n)})|}=1$
}

{\bf Proof:}
Considering the worst case of \edited{Equation} \ref{eq:Epropagate}:
\begin{align}
\frac{|\Prob(E_{k}^{(n+2)}) - \Prob(E_{k}^{(n+1)})|}{|\Prob(E_k^{(n+1)}) - \Prob(E_k^{(n)})|}
&=\frac{\prod_{j=1}^{n+1}(1-\frac{\Prob(E_{k-1}^{(j)})\cdot\rho}{j}) - \prod_{j=1}^{n+2}(1-\frac{\Prob(E_{k-1}^{(j)})\cdot\rho}{j})}{\prod_{j=1}^{n}(1-\frac{\Prob(E_{k-1}^{(j)})\cdot\rho}{j}) - \prod_{j=1}^{n+1}(1-\frac{\Prob(E_{k-1}^{(j)})\cdot\rho}{j})}\notag\\
&=\frac{(1-\frac{\Prob(E_{k-1}^{(n+1)})\cdot\rho}{n+1})- (1-\frac{\Prob(E_{k-1}^{(n+1)})\cdot\rho}{n+1})(1-\frac{\Prob(E_{k-1}^{(n+2)})\cdot\rho}{n+2})}{1-(1-\frac{\Prob(E_{k-1}^{(n+1)})\cdot\rho}{n+1})}\notag\\
&=(1-\frac{\Prob(E_{k-1}^{(n+1)})\cdot\rho}{n+1})\cdot\frac{\frac{\Prob(E_{k-1}^{(n+2)})\cdot\rho}{n+2}}{\frac{\Prob(E_{k-1}^{(n+1)})\cdot\rho}{n+1}}\notag\\
&=(1-\frac{\Prob(E_{k-1}^{(n+1)})\cdot\rho}{n+1})           \cdot \frac{n+1}{n+2}             \cdot \frac{\Prob(E_{k-1}^{(n+2)})}{\Prob(E_{k-1}^{(n+1)})}\label{eq:tripleConverge}
\end{align}
\noindent Clearly for \edited{Equation} \ref{eq:tripleConverge}, as $n\to\infty$, $(1-\frac{\Prob(E_{k-1}^{(n+1)})\cdot\rho}{n+1})$ converges to 1, as well as $\frac{n+1}{n+2}$ converges to 1. Given Thm. \ref{thm:drtsimilar}, the limit for $\Prob(E_{k-1}^{(n)})$ exists and is non-zero. Furthermore, the monotonicity of $\Prob(E_{k-1}^{(n)})$ implies that $\frac{\Prob(E_{k-1}^{(n+2)})}{\Prob(E_{k-1}^{(n+1)})}$ converges to 1. Therefore:
\vspace{-.1in}
\begin{align}
\lim_{n\to\infty}\frac{|\Prob(E_{k}^{(n+2)}) - \Prob(E_{k}^{(n+1)})|}{|\Prob(E_k^{(n+1)}) - \Prob(E_k^{(n)})|}=1\cdot 1\cdot 1=1.\notag
\end{align}\telos

\section*{Appendix F}
{\bf Theorem \ref{thm:drtinfinite}. }\emph{
 For any state $x_i\in V$ such that $x_i$ is added into the set of vertices $V$ at iteration $i$, then \rt\ will select $x_i$ for \mcprop\ {\it infinitely often} as the number of iterations reach infinity, i.e.: \[\Prob(\limsup_{n\to\infty}\{x_i\text{ is selected}\}) = 1.\]
}
\vspace{-.05in}
{\bf Proof:}
Let $x_i$ denote a state which is added to $V$ at iteration $i$ and let $S_i^{(n)}$ denote the event such that $x_i$ is selected for \mcprop\ at iteration $n$ (clearly $i<n$). During each iteration, the algorithm uniformly at random selects a state for \mcprop. The probability of such an event can be written as
$\Prob(S_i^{(n)})= \frac{1}{n}$.
The summation of the first $n-i$ terms of the sequence is:
\begin{align}
\sum_{j=1}^n \Prob(S_i^{(j)})=\sum_{j=1}^n \frac{1}{j} - \sum_{j=1}^i \frac{1}{j}.
\label{inf1}
\end{align}

\noindent
The first term on the right side is {\it harmonic series}, and the second term is the $i$-th {\it harmonic number}.
A property of {\it harmonic series} is:
\vspace{-.2in}
\begin{align*}
&\sum_{j=1}^n \frac{1}{j} = ln(n) + c_\gamma + \epsilon_n\\
c_\gamma=0.577...&\text{ (\emph{Euler-Mascheroni} constant)}, \epsilon_n\sim\frac{1}{2n}\text{ such that }\lim_{n\to\infty}\epsilon_n=0.
\end{align*}

\noindent
Therefore, Eq. \ref{inf1} diverges as $n\to\infty$.
\begin{align*}
\sum_{j=1}^\infty \Prob(S_i^{(j)}) = \lim_{n\to\infty} [ln(n) + c_\gamma + \epsilon_n - \sum_{j=1}^i \frac{1}{j}] \geq +\infty.
\end{align*}

\noindent Selecting $x_i$ is independent at any two different iterations (and
combinations) after $x_i$ has been extended. This is because the algorithm uniformly at
random picks one vertex among existing ones at each iteration. Then,
according to the \emph{second Borel$-$Cantelli lemma},

\vspace{-.4in}
\begin{align*}
\Prob(\limsup_{n\to\infty} S_i^{(n)}) = 1
\end{align*}
\vspace{-.05in}
Therefore, $x_i$ shall be selected for \mcprop\ \emph{infinitely often} as the number of execution times $n\to\infty$. 
\telos

\edited{\section*{Appendix G}}
{\bf Theorem \ref{thm:deltacompleteBN}.}\emph{
\rrtbestnear\ will eventually generate a $\delta$-similar trajectory to any optimal trajectory.}

\noindent {\bf Proof:} The probability of $\neg E_k^{(n)}$ occurring
depends on a sequence of $A_k$ events failing:


\vspace{-0.2in}
\begin{align}
\Prob(\neg E_k^{(n)}) = \Prob(\neg A_k^{(1)})\cdot \Prob(\neg
A_k^{(2)}|\neg A_k^{(1)})\cdot \text{ ... }\cdot \Prob(\neg
A_k^{(n)}|\bigcap_{j=1}^{n-1}\neg A_k^{(j)})
	\label{eq:PconcatBN}
\end{align}
\vspace*{-.15in}

The probability that $\neg A_k^{(n)}$ happens given
$\bigcap_{j=1}^{n-1} \neg A_k^{(j)}$ is equivalent to the probability
of failing to generate a trajectory reaching
$\ball_{\delta_{BN}}(x_{k-1}^\ast)$ plus the probability that a
trajectory has been generated to $\ball_{\delta_{BN}}(x_{k-1}^\ast)$,
but fails to generate a new trajectory segment to
$\ball_{\delta_{BN}}(x_{k}^\ast)$, i.e.:

\vspace{-0.2in}
\begin{align}
\Prob(\neg A_k^{(n)}|\bigcap_{j=1}^{n-1}\neg A_k^{(j)})
&=\Prob(\neg E_{k-1}^{(n)}) + \Prob(E_{k-1}^{(n)})\cdot \Prob(\{\text{step fail to } \ball_{\delta_{BN}}(x_k^\ast) \})\notag\\
&\leq \Prob(\neg E_{k-1}^{(n)}) + \Prob(E_{k-1}^{(n)})(1-\gamma\rho_{\delta\to\delta_{BN}})\notag\\
&= 1 - \Prob(E_{k-1}^{(n)})\cdot \gamma\rho_{\delta\to\delta_{BN}}.
	\label{eq:PAiBN}
\end{align}
\vspace{-0.2in}

Therefore, using \edited{Equation} \ref{eq:PconcatBN} and \edited{Equation} \ref{eq:PAiBN}:
\vspace{-0.15in}
\begin{equation}
\Prob(E_k^{(n)}) \geq 1 - \prod_{j=1}^n (1 -
\Prob(E_{k-1}^{(j)})\cdot \gamma\rho_{\delta\to\delta_{BN}}).
\label{eq:EpropagateBN}
\end{equation}
\vspace{-0.15in}

\noindent {\bf For the base case}, $\Prob(E_0^{(j)}) = 1$ because
$\xinit$ is always in $\ball_{\delta_{BN}}(x_0)$. Then, consider event
$E_{1}$ from iteration 1 to $n$ using the last equation above. The
probability of $E_1$ is:

\vspace{-.1in}
$$\Prob(E_1^{(n)}) \geq
1-\prod_{j=1}^n(1-\gamma\rho_{\delta\to\delta_{BN}}) =
1-(1-\gamma\rho_{\delta\to\delta_{BN}})^n \Rightarrow$$

\vspace{-.1in}
$$\lim_{n\to\infty} \Prob(E_1^{(n)}) \geq
1-\lim_{n\to\infty}(1-\gamma\rho_{\delta\to\delta_{BN}})^n = 1-0 = 1.$$
\vspace{-.1in}

\noindent
{\bf For the induction step}, if $\lim_{n\to\infty} \Prob(E_{k}^{(j)})
=1$, we need to show that the same will be true for
$E_{k+1}^{(n)}$. Set $y_k^{(n)} = \prod_{j=1}^n (1 -
\Prob(E_{k-1}^{(j)})\cdot \gamma\rho_{\delta\to\delta_{BN}})$. The
logarithm of $y_k^{(n)}$ behaves as follows:

\vspace{-0.1in}

$$\log y_k^{(n)} = \log \prod_{j=1}^n (1 - \Prob(E_{k-1}^{(j)})\cdot \gamma\rho_{\delta\to\delta_{BN}})
= \sum_{j=1}^n \log (1 - \Prob(E_{k-1}^{(j)})\cdot
\gamma\rho_{\delta\to\delta_{BN}}) \Rightarrow $$
\vspace{-0.25in}

\begin{equation}
\log y_k^{(n)} < \sum_{j=1}^n - \Prob(E_{k-1}^{(j)})\cdot
\gamma\rho_{\delta\to\delta_{BN}} =
-\gamma\rho_{\delta\to\delta_{BN}}\cdot\sum_{j=1}^n
\Prob(E_{k-1}^{(j)}).
\label{eq:logyi}
\end{equation}
\vspace{-0.1in}

\noindent
From the inductive assumption that, $\Prob(E_k^{(j)})$ converges to 1 as
$j\to\infty$, then $\lim_{n\to\infty}\sum_{j=1}^n
\Prob(E_k^{(j)})=\infty$. Then:
\vspace{-.15in}
\begin{align}
\lim_{n\to\infty} \log y_{k+1}^{(n)} < -\gamma\rho_{\delta\to\delta_{BN}}\cdot\lim_{n\to\infty}\sum_{j=1}^n \Prob(E_{k}^{(j)})=-\infty
 \iff \lim_{n\to\infty} y_{k+1}^{(n)} = 0.\notag
\end{align}
\vspace{-.15in}

\noindent
Using \edited{Equation} (\ref{eq:EpropagateBN}), with $\lim_{n\to\infty} y_{k+1}^{(n)}
= 0$, it can be shown that:
$
\lim_{n\to\infty} \Prob(E_{k+1}^{(n)}) = 1-\lim_{n\to\infty}y_{k+1}^{(n)}
= 1-0 = 1.
$\telos

\vspace{-.05in}
\begin{coro}
\rrtbestnear\ is probabilistically $\delta$-robustly complete.
\label{coro:deltacompleteBN}
\end{coro}
\vspace{-.1in}

\edited{\section*{Appendix H}}
{\bf Theorem \ref{thm:bestnearoptimal}.}\emph{
\rrtbestnear\ is asymptotically $\delta$-robustly near-optimal.}

\noindent {\bf Proof:} Let $\overline{x^\prime_{i-1}\to x_i}$ denote
the $\delta$-similar trajectory segment generated by \rrtbestnear\
where $x^\prime_{i-1} \in \ball_{\delta}(x^\ast_{i-1})$ of the optimal
path and
$x_i \in \ball_{\delta_{BN}}(x^\ast_i)$. \edited{Theorem} \ref{thm:mccomplete}
guarantees the probability of generating it by \mcprop\ can be lower
bounded as $\rho_{\delta\to\delta_{BN}}$. Then from the definition of
$\delta$-similar trajectories and \emph{Lipschitz continuity} for
cost:
\vspace{-.1in}
\begin{align}
\text{{\tt cost}}(\overline{x^\prime_{i-1}\to x_i})\leq\text{{\tt cost}}(\overline{x_{i-1}^\ast\to x_i^\ast})+K_c\cdot\delta.
\label{eq3}
\end{align}
\vspace{-.25in}

\edited{Lemma} \ref{lem:gammaBN} guarantees that when $x_i$ exists in
$\ball_{\delta_{BN}}(x^\ast_i)$, then $x^\prime_i$, returned by the
{\tt BestNear} function with probability $\gamma$, must have equal or
less cost, i.e., $x^\prime_i$ can be the same state as $x_i$ or a
different state with smaller or equal cost:

\begin{align}
\text{{\tt cost}}(x^\prime_i)\leq\text{{\tt cost}}(x_{i}).
\label{eq4}
\end{align}
\vspace{-.3in}

\noindent Consider $\ball_\delta(x_1^\ast)$, as illustrated in
\edited{Figure} \ref{fig:sparserrt_reasoning}, according to \edited{Equation} \ref{eq3} and
\edited{Equation} \ref{eq4}: $ \text{{\tt cost}}(\overline{x_{0}\to
x^\prime_1}) \leq\text{{\tt cost}}(\overline{x_{0}\to
x_1}) \leq\text{{\tt cost}}(\overline{x_{0}\to
x_1^\ast})+K_c\cdot\delta\notag $.  Assume this is true for $k$
segments, $$
\text{{\tt cost}}(\overline{x_{0}\to x^\prime_k})\leq\text{{\tt cost}}(\overline{x_{0}\to x_k^\ast})+k\cdot K_c\cdot\delta
.$$ Then, the cost of the trajectory with $k+1$ segments is:

\vspace{-.3in}
\begin{align*}
\text{{\tt cost}}(\overline{x_{0}\to x^\prime_{k+1}})\leq	\text{{\tt cost}}(\overline{x_{0}\to x_{k+1}})
=&\ \text{{\tt cost}}(\overline{x_{0}\to x^\prime_k}) + \text{{\tt
cost}}(\overline{x^\prime_{k}\to x_{k+1}})\\
\leq&\ \text{{\tt cost}}(\overline{x_{0}\to x_k^\ast})+k
K_c\delta+\text{{\tt cost}}(\overline{x^\prime_{k}\to x_{k+1}})\\
\leq&\ \text{{\tt cost}}(\overline{x_{0}\to x_k^\ast})+k
K_c\delta+\text{{\tt cost}}(\overline{x_{k}^\ast\to
x_{k+1}^\ast})+K_c\delta\\
=&\ \text{{\tt cost}}(\overline{x_{0}\to x_{k+1}^\ast})+(k+1)
K_c\delta.
\end{align*}
\vspace{-.25in}

\noindent By induction, this holds for all $k$.  Since the largest $k$
is $\frac{C^\ast}{C_\Delta}$:

\vspace{-.3in}
\begin{align*}
\text{{\tt cost}}(\overline{x_{0}\to x^\prime_k})
\leq\text{{\tt cost}}(\overline{x_{0}\to x_k^\ast})+k\cdot K_c\cdot\delta
=(1+\frac{K_c\cdot\delta}{C_\Delta})\cdot C^\ast.
\end{align*}
\vspace{-.25in}

\noindent Recall from Theorem \ref{thm:deltacompleteBN}, event $E_k$
of generating a $\delta$-similar trajectory to the $k$-th segment of the optimal trajectory $\pi^{\ast}$. Then:
\vspace{-.1in}
$$\mathbb{P}\Big(E_k^{(n)}\Big)=\mathbb{P}\Big(\big\{Y_n^{SST}\leq(1+\frac{K_c\cdot\delta}{C_\Delta})\cdot
C^\ast\big\}\Big)$$
\vspace{-.15in}

\noindent As $n\to\infty$, since $\rho_{\delta\to\delta_{BN}} > 0$,
$E_k^{(\infty)}$ almost surely happens:

\vspace{-.1in}
$$\mathbb{P}\Big(\big\{\limsup_{n\to\infty}Y_n^{SST}\leq(1+\frac{K_c\cdot\delta}{C_\Delta})\cdot
C^\ast\big\}\Big)=\lim_{n\to\infty}\mathbb{P}\Big(E_k^{(n)}\Big)=1. \ \ \ \ \ \ \telos$$
\vspace{-.1in}

\edited{\section*{Appendix I}}


In Lemma \ref{lem:rhoRatio}, the goal is to show that given different radii for covering balls, the probability for extending new trajectories to the small ball is proportional to the original probability.

{\bf Lemma \ref{lem:rhoRatio}. }\emph{
For a $\ball_i$ of radius $\delta$ and a ball $\ball_i^\prime$ with
radius $\delta^\prime$, such that $\delta^\prime / \delta=\alpha$, where
$\alpha\in (0,1)$, there is
$$\frac{{\rho}_{\delta^\prime}}{{\rho}_{\delta}}=\alpha^{w+1}$$
}

{\bf Proof:} Recall the system equation \ref{eq:dynamics}. If $x_1$ and $x_2$ are on
the same trajectory $x(t)$ such that $x_1=x(t_1)$ and
$x_2=x(t_1+\Delta t)$, then: 
\[x_2=x_1+\int_{t1}^{(t_1+\Delta t)}
f(x(t),u(t))\cdot dt\] Then \[||x_2-x_1||=||\int_{t_1}^{(t_1+\Delta
t)} f(x(t),u(t))\cdot dt||\] From Assumption \ref{ass:system},
$f(x(t),u(t))$ is bounded as well, e.g., $f(x(t),u(t))\leq M_f\in
R^+$, meaning that:
\begin{align*}
||x_2-x_1||	\leq|| \int_{t1}^{(t_1+\Delta t)} M_f\cdot dt|| =M_f\cdot \Delta t
\end{align*}

In other words, for two states that are on a same trajectory, the
duration of the trajectory connecting them and their \emph{Euclidean} distance in the state space satisfy the
following property:
\[\Delta t\geq\frac{||x_2-x_1||}{M_f}\]

Recall that $\rho$ is lower bounded by Theorem \ref{thm:mccomplete}. It can be
further reduced to
\[  {\rho}_{\delta}=\frac{2(1-\lambda)\delta}{M_f\cdot T_{prop}}\cdot \frac{\zeta\cdot(\frac{\lambda\delta}{K_u\cdot T_{prop}\cdot e^{K_x\cdot T_{prop}}})^w}{\mu(U_m)}>0  \]

Consider two sets of $\balls$($x(t),\delta,T$) and
$\balls$($x(t),\delta^\prime,T$). Note that the only
difference between $\delta$ and $\delta^\prime$ is that
$\delta^\prime/\delta = \alpha\in (0, 1]$. 
Given the above expression, if the ratio $\frac{\rho_{delta^\prime}}{\rho_{delta}}$ is evaluated, then it will always be possible to express this ratio in terms of $\alpha$, so that 
$\frac{{\rho}_{\delta^\prime}}{{\rho}_{\delta}}=\alpha^{w+1}$
\telos

\end{document}